\documentclass{article} 

\usepackage[final]{neurips_2024}
\usepackage{microtype}
\usepackage{booktabs} 
\usepackage{float}
\usepackage{amsfonts}
\usepackage{bbm}
\usepackage{hyperref}
\usepackage{color}
\usepackage{url}
\usepackage{multirow}
\usepackage{enumitem}
\usepackage{xcolor}        
\usepackage{pifont}

\usepackage{mathtools}
\usepackage[utf8]{inputenc} 
\usepackage{xr-hyper}
\usepackage[T1]{fontenc}    
\usepackage{url}            
\usepackage{nicefrac}       
\usepackage{wrapfig}

\usepackage{amsmath}
\usepackage{amssymb}
\usepackage{amsthm}
\usepackage{wrapfig}
\usepackage{subfig}
\usepackage{comment}
\usepackage{xr}
\usepackage{listings}
\usepackage[countmax]{subfloat}
\usepackage{cancel}
\usepackage[capitalize,noabbrev]{cleveref}

\usepackage{subfiles}
\definecolor{mypink1}{rgb}{0.858, 0.188, 0.478}

\makeatletter
\newtheorem*{rep@theorem}{\rep@title}
\newcommand{\newreptheorem}[2]{%
\newenvironment{rep#1}[1]{%
 \def\rep@title{#2 \ref{##1}}%
 \begin{rep@theorem}}%
 {\end{rep@theorem}}}
\makeatother

\newtheorem{proposition}{Proposition}
\newtheorem{theorem}{Theorem}
\newreptheorem{corollary}{Corollary}
\newtheorem{definition}{Definition}
\newtheorem{claim}{Claim}

\newtheorem{assumption}{Assumption}

\newcommand{\W}[1]{ W^{(#1)}}

\newcommand{\I}{\mathbf{I}}

\newcommand{\U}{\mathbf{U}}

\newcommand{\y}{\mathbf{y}}

\newcommand{\Z}{\mathbf{Z}}

\newcommand{\bS}{\mathbf{S}}

\newcommand{\diag}{\operatorname{diag}}

\def\W{{\cal{W}}}

\newcommand{\Or}{or}

\newcommand{\x}{{\bf x}}

\newcommand{\C}{{\bf C}}
\newcommand{\K}{{\bf K}}
\newcommand{\PP}{{\bf P}}
\newcommand{\uu}{{\bf u}}
\newcommand{\vv}{{\bf v}}
\newcommand{\z}{{\bf z}}
\newtheorem{lemma}{Lemma}
\newtheorem{corollary}{Corollary}
\newcommand{\ours}{GCA}

\DeclareCaptionLabelFormat{andtable}{#1~#2  \&  \tablename~\thetable}

\usepackage{algorithm}
\usepackage{algpseudocode}
\newcounter{appendixalgorithm}

\usepackage[toc,page,header]{appendix}
\usepackage{minitoc}

\makeatletter
\@ifundefined{c@lofdepth}{}{%
  \let\c@lofdepth\relax
  
}
\@ifundefined{c@lotdepth}{}{%
  \let\c@lotdepth\relax
  
}
\makeatother

\usepackage{tocloft}

\title{Your contrastive learning problem is secretly a distribution alignment problem}

\author{Zihao Chen\thanks{Contact: \{zchen959, evadyer\}@gatech.edu}\ , 
Chi-Heng Lin, 
Ran Liu, 
Jingyun Xiao, 
Eva L.~Dyer$^*$\\
School of Electrical \& Computer Engineering\\
Georgia Tech, Atlanta, GA}

\begin{document}

\newpage
\maketitle

\begin{abstract}
Despite the success of contrastive learning (CL) in vision and language, its theoretical foundations and mechanisms for building representations remain poorly understood. In this work, we build connections between noise contrastive estimation losses widely used in CL and distribution alignment with entropic optimal transport (OT). This connection allows us to develop a family of different losses and multistep iterative variants for existing CL methods. Intuitively, by using more information from the distribution of latents, our approach allows a more distribution-aware  manipulation of the relationships within augmented sample sets.
We provide theoretical insights and experimental evidence demonstrating the benefits of our approach for {\em generalized contrastive alignment}. Through this framework, it is possible to leverage tools in OT to build unbalanced losses to handle noisy views and customize the representation space by changing the constraints on alignment.
By reframing contrastive learning as an alignment problem and leveraging existing optimization tools for OT, our work provides new insights and connections between different self-supervised learning models in addition to new tools that can be more easily adapted to incorporate domain knowledge into learning.

\end{abstract}
\section{Introduction}

In machine learning, the availability of vast amounts of unlabeled data has created an opportunity to learn meaningful representations without relying on costly labeled datasets \cite{jaiswal2020survey,shurrab2022self,jing2020self}. Self-supervised learning has emerged as a powerful solution to this problem, allowing models to leverage the inherent structure in data to build useful representations. Among self-supervised methods, contrastive learning (CL) is widely adopted for its ability to create robust representations by distinguishing between similar (positive) and dissimilar (negative) data pairs. With success in fields like image and language processing \cite{chen2020simple,radford2021learning}, contrastive learning now also shows promise in domains where cross-modal, noisy, or structurally complex data make labeling especially challenging \cite{liu2021drop,vishnubhotla2024towards,chen2023instance}.

Traditional contrastive learning methods primarily aim to bring positive pairs---often augmentations of the same sample---closer together in representation space. While effective, this approach often struggles with real-world challenges such as noise in views, variations in data quality, or shifts introduced by complex transformations, where positive pairs may not perfectly align. Additionally, in tasks requiring domain generalization, aligning representations across diverse domains (e.g., variations in style or sensor type) is critical but difficult to achieve with standard contrastive learning, which typically lacks mechanisms for incorporating domain-specific relationships. These limitations highlight the need for a more flexible approach that can adapt alignment strategies based on the data structure, allowing for finer control over similarity and dissimilarity among samples.

To address this challenge, we introduce a novel \emph{generalized contrastive alignment} (GCA) framework, which reinterprets contrastive learning as a distributional alignment problem. Our method allows flexible control over the alignment of samples by defining a target transport plan, \(\mathbf{P}_{tgt}\), that serves as a customizable alignment guide. For example, setting \(\mathbf{P}_{tgt}\) to resemble a diagonal matrix encourages each positive to align primarily with itself or its augmentations, thereby reducing the effect of noise between views. Alternatively, we can incorporate more complex constraints, such as weighting alignments based on view quality or enforcing partial alignment structures where noise or data heterogeneity is prevalent. This flexibility enables GCA to adapt effectively to a wide range of tasks, from simple twin view alignments to scenarios with noisy or variably aligned data.

Our approach also bridges connections between GCA and established methods, such as InfoNCE (INCE) \cite{oord2018representation}, Robust InfoNCE (RINCE) \cite{chuang2022robust}, and BYOL \cite{grill2020bootstrap}, demonstrating that these can be viewed as iterative alignment objectives with Bregman projections \cite{cai2022developments,grathwohl2019your}. This perspective allows us to systematically analyze and improve uniformity within the latent space, a property that enhances representation quality and ultimately boosts downstream classification performance.

We validate our method through extensive experiments on both image classification and noisy data tasks, demonstrating that GCA’s unbalanced OT (UOT) formulations improve classification performance by relaxing our constraints on alignment. Our results show that \ours~offers a robust and versatile framework for contrastive learning, providing flexibility and performance gains over existing methods and presenting a promising approach to addressing different sources of variability in self-supervised learning.

The contributions of this work include:
\begin{itemize}
    \item A new framework called \emph{generalized contrastive alignment} (GCA), which reinterprets standard contrastive learning as a distributional alignment problem, using optimal transport to provide flexible control over alignment objectives. This approach allows us to derive a novel class of contrastive losses and algorithms that adapt effectively to varied data structures and build  customizable transport plans.

    \item We present GCA-UOT, a contrastive learning method that achieves strong performance on standard augmentation regimes and excels in scenarios with more extreme augmentations or data corrupted by transformations. GCA-UOT leverages unbalanced transport to adaptively weight positive alignments, enhancing robustness against view noise and cross-domain variations.

    \item We provide theoretical guarantees for the convergence of our GCA-based methods and show that our alignment objectives improve representation quality by enhancing the uniformity of negatives and strengthening alignment within positive pairs. This leads to more discriminative and resilient representations, even in challenging data conditions.

    \item Empirically, we demonstrate the effectiveness of \ours~in both image classification and domain generalization tasks. Through flexible, unbalanced OT-based losses, \ours~achieves superior classification performance and adapts alignment to include domain-specific information where relevant, without compromising classification accuracy in domain generalization.
\end{itemize}

\section{Background}

\subsection{Contrastive learning}
\vspace{-1mm}
Contrastive learning (CL) is a representation learning methodology that uses positive and negative pairs to define similarity in the latent space. Let $\mathcal{D} = \{ \x_i \}_{i=1}^N$ denote our dataset. For each sample $\x_i$ in a batch of training data with size \(B\), 
we create two augmented copies $\x'_i$ and $\x''_i$ independently, i.e., $\x_i'=\psi(\x_i)$ where $\psi$ is a randomly drawn augmentation function from some augmentation class $\mathcal{A}$ and likewise for $\x''_i$. The $(\x'_i,\x''_i)$ is called a positive pair of $\x_i$ while $(\x_i',\x''_j)$ is treated as a negative pair for any $j \neq i$. One of the most widely used formulations of the CL problem, InfoNCE (INCE)~\cite{chen2020simple}, seeks to maximize the negative log probability that a sample is correctly classified as
\begin{equation}
\label{eq:INCE}
    \mathcal{L}_\text{INCE}=  -  \log \bigg( \frac{e^{s_{ii}}}{e^{s_{ii}}+\sum_{i\neq j} e^{s_{ij}} } \bigg),
\end{equation}
where $s_{ij} = \varepsilon^{-1}f_{\theta}({\x_i'})^\top f_{\theta}(\x_j'')/\|f_{\theta}({\x_i'})\|\|f_{\theta}({\x_j''})\|$ is the score between augmented samples.

Building upon the principles of INCE, SimCLR \cite{chen2020simple} and MoCo \cite{he2020momentum} are two representative works that form the foundation of contrastive learning methods for visual representation tasks. Alternatively, BYOL \cite{grill2020bootstrap} and SimSiam \cite{chen2021exploring} discard the use of negative samples to avoid large batch size and instead use exponential moving average-based updates to avoid representational collapse. 
Recent contrastive methods have focused on improving the tolerance to noise in samples to enhance robustness in diverse scenarios \cite{chuang2020debiased}. 
Among them, Robust INCE (RINCE) is a robust contrastive loss function characterized by its symmetric properties and theoretical resistance to noisy labels \cite{robinson2020contrastive,chuang2022robust}.
Specifically, RINCE provides robustness to noisy views by introducing adjustable parameters $\lambda$ and $q$ ~\cite{chuang2022robust} which rebalance the cost of positive and negative views, resulting in the following loss: 
\begin{equation}\label{eq:RINCE}
    \mathcal{L}^{\lambda,q}_\text{RINCE}=\frac{1}{q}\big(-e^{ q s_{ii}} + \lambda^q(e^{ s_{ii}}+\sideset{}{_{i\neq j}}\sum e^{s_{ij}})^q\big)
\end{equation}
By optimizing the above loss functions, the encoder $f$ is trained to construct a semantically coherent representation space where positive pairs of samples are positioned nearby, while those negative pairs with divergent semantic attributes are separated~\cite{wang2020understanding}.

\vspace{-2mm}
\subsection{Proximal Operators and Projections}

To make the connections between different CL losses clearer later, we use the notion of proximal operators. In words, the proximal operator will provide a way to find the closest point in some closed convex set. Formally, we can define the proximal operator as follows.
\begin{definition} [Proximal Operator]\label{def:prox} Let $d_\Gamma(\x, {\bf v}) = \Gamma(\x) - \Gamma({\bf v}) - \langle \nabla \Gamma({\bf v}), \x - {\bf v} \rangle$ be a Bregman divergence with a convex function \(\Gamma\). The proximal operator of $h: \mathcal{X} \to \mathbb{R} \cup \{+\infty\}$ is defined for a point ${\bf v} \in \mathcal{X}$ with a closed convex set $\mathcal{B} \subseteq \mathcal{X}$ :
\begin{equation*}~\label{eq:prox}
\text{Prox}_{h,\mathcal{B}}^{d_\Gamma}({\bf v}) = \arg\min_{{\bf x} \in \mathcal{B}} \left\{ h({\bf x}) +  d_\Gamma({\bf x}, {\bf v}) \right\}. 
\end{equation*}
\vspace{-2mm}
\end{definition}
\vspace{-3mm}

Moreover, we can define the concept of a projection as a special case of the proximal operator when we let \(h(\x)\) be an indicator function $
h_\mathcal{B}(x) = \{ 0, \text{if } x \in \mathcal{B}; 
\infty, \text{if } x \notin \mathcal{B}\}$ on constraint set \(\mathcal{B}\). See Appendix~\ref{app:prox} for more details.

\vspace{-2mm}
\subsection{Solving Optimal Transport Through Proximal Point Methods}\label{sec:otppm}
Optimal transport (OT) is widely used in characterizing the distance between two collections of samples $\{ {\bf x}_i \}_{i=1}^B $ and $\{ {\bf y}_j \}_{j=1}^B$ with associated measures \( \mu = \sum_{i=1}^{B} \delta_{{\bf x}_i} p_i~\text{and}~\nu = \sum_{j=1}^{B} \delta_{{\bf y}_j} q_j\) with Dirac delta function \(\delta_{{\bf x}}\) and \(\delta_\y\) on finite support~\cite{peyre2019computational}. Here, \( p \) and \( q \) are vertices of the \( \mathbb{R}^{B} \) simplex defined as \( \Delta_B := \{ v \in \mathbb{R}^B : v_i \geq 0, \sum_{i=1}^B v_i = 1 \}\). OT aims to learn a joint coupling matrix, or transport plan \(\PP \in \mathbbm R^{B\times B}_{+}\) that minimizes the cost of transporting mass encoded by cost matrix \(\C\in \mathbbm R^{B\times B}_{+}\), from one distribution to another.
In practice, entropy regularization is used to solve the OT objective, resulting in the following entropy-regularized OT (EOT) objective: 
\begin{equation}\label{eq:eot}
\min_{\textbf{P} \in \mathcal{B}} ~ \langle\textbf{P},\textbf{C} \rangle -\varepsilon H(\textbf{P}), \quad \text{where}~H(\PP)= -\sum_{ij} \PP_{ij} \log({ \PP}_{ij}),
\end{equation} 
where $\varepsilon$ is a user specified parameter that controls the amount of smoothing in the transport plan, and ${\bf C }({\bf x}, {\bf y}) = 1- \langle {\bf x} , {\bf y}\rangle / \| {\bf x} \| \| {\bf y} \|$ is often set to encode the cosine similarity between pairs of samples.

\vspace{-2mm}
\paragraph{The Sinkhorn Algorithm and its Interpretation as a Bregman Projection.} 
Solving Equation~\eqref{eq:eot} could be interpreted as iterative alignment problem on a Hilbert space generated from the kernel \(\K_{ij} =\exp(-\C_{i,j} / \varepsilon) \). 
This alignment problem can be solved through iterative Bregman projections onto the two constraints sets that encode the marginals along the rows and columns ~\cite{benamou2015iterative, bregman1967relaxation, peyre2019computational}:
 \begin{equation}\label{eq:Birkhoff}
    C_1^\mu \coloneqq \{\PP : \PP\mathbbm{1}_B = \mu\}, C_2^\nu \coloneqq \{\PP : \PP^\top \mathbbm{1}_B = \nu\}
\end{equation}
The first step of Bregman projection is to find the minimizer \(\PP^{(1)}=\arg\min \{\varepsilon\text{KL}(\PP\|\K): \PP\mathbbm{1}_B =\mu \}\) by the proximal operator \(\text{Prox}_{C_1^\mu}^{\text{KL}}(\K)\) with Lagrange multiplier \(f\) on the row constraint set $\mathcal{C}_1^{\mu}$, and compute its derivatives with respect to \(\PP\) with \(\uu = e^{f/\varepsilon} > 0\):
\begin{equation}
\varepsilon\log(\PP^{(1)}/ \K) - f\mathbbm{1} = 0 \Rightarrow \PP^{(1)} = \uu\K, \quad \langle \PP^{(1)}, \mathbbm{1}\rangle = \mu \Rightarrow \langle \uu\K, \mathbbm{1}\rangle = \mu, \uu = \frac{\mu}{\K \mathbbm{1}}
\end{equation}
Next, we project \(\PP^{(1)}\) onto the column constraint set \(C_2^\nu\), resulting in 
$
\PP^{(2)} \coloneqq \text{Prox}^\text{KL}_{C_2^\nu}(\PP^{(1)}) = \PP^{(1)} \text{diag}(\frac{\nu}{\PP^{(1)\top}\mathbbm{1}_B}).$ The iterative updates can be succinctly expressed as the Sinkhorn iterations:
\begin{equation}\label{eq:sinkp}
    {\bf P}^{(2t+1)}=\operatorname{diag}({\bf u}^{(t+1)})\mathbf{K}\operatorname{diag}({\bf v}^{(t)}), \quad {\bf P}^{(2t+2)}=\operatorname{diag}({\bf u}^{(t+1)})\mathbf{K}\operatorname{diag}({\bf v}^{(t+1)}),
\end{equation}
with the scaling vectors \(\uu^{(t)}\) and \(\vv^{(t)}\) updated according to:
\begin{equation}\label{eq:uvupdates}  
{\bf u}^{(t+1)} \stackrel{\text { def }}{=} \frac{{\mu}}{\mathbf{K} \vv^{(t)}}, \quad {\bf v}^{(t+1)} \stackrel{\text { def }}{=} \frac{{\nu}}{\mathbf{K}^{\mathrm{T}} {\bf u}^{(t)}}.
\end{equation}
Here, iterations converge to a stable transport plan \(\PP^{(\infty)}\)as the optimal solution of Equation~\eqref{eq:eot}, which provides the minimum cost matching between two distributions.
The convergence and dynamics of OT and its dual formulation have been studied extensively in~\cite{berman2020sinkhorn, peyre2019computational, ghosal2022convergence, an2022efficient}. Thus, these results guarantee that the iterates will converge to the optimal solution of the EOT objective, or that \(\PP^{(t)} \rightarrow \PP^{(\infty)}\) with \(t \rightarrow \infty\). 
See Appendix~\ref{app:ot} for more details on both the continuous and discrete formulations of OT.  
\vspace{-2mm}
\subsection{Wasserstein Dependency Measure}
The Wasserstein Dependency Measure (WDM) is a measure of deviation between two probability measures. We will use this later and thus provide the formal definition here~\cite{ozair2019wasserstein}.
\begin{definition}[Wasserstein Dependency Measure] \label{def:wdm} Define the WDM as the Wasserstein distance (\(W_1\)) between the joint distribution $\pi(x, y)$ and the product of marginal distributions $\mu \otimes \nu(x, y)$ of two random variables $x$ and $y$. \( W_1(\pi, \mu \otimes \nu) = \sup_{f \in \mathcal{C}(\mathcal{X} \times \mathcal{Y})} \left(\mathbb{E}_{\pi(x, y)}[f(x, y)] - \mathbb{E}_{\mu \otimes \nu(x, y)}[f(x, y)]\right)\), where \(\mathcal{C}(\mathcal{X} \times \mathcal{Y})\) denotes the set of all 1-Lipschitz functions from \(\mathcal{X} \times \mathcal{Y}\) to \(\mathbb{R}\). 
\end{definition}

\subsection{Optimal Transport and Alignment in Representation Learning}

Distribution alignment and OT have been widely used for domain adaptation \cite{lin2021making, courty2014domain, lee2019hierarchical, wang2024extraction}, and in generative modeling \cite{arjovsky2017wasserstein,tolstikhin2017wasserstein, rout2021generative, wang2024exploring}.
The connections between distribution alignment and contrastive learning, however, are still nascent.  In \cite{shi2023understanding}, the authors explore the connection between inverse OT (IOT) \cite{li2019learning, stuart2020inverse, fang2023s} and INCE. Our work builds on this connection to OT to build robust divergences (RINCE) and to build a novel unbalanced optimal transport (UOT) method (Section~\ref{subsec:relax}). Additionally, we show how our framework can be used to build flexible methods for encouraging contrast at multiple levels. We use this concept of hierarchical contrast and show that it can be used in  domain generalization settings (Section~\ref{sec:domaingen}). 
It is of note that GCA-UOT focuses on relaxing the hard constraints on the row and columns into the soft penalties,  which is different with the idea of ``unbalanced matching''  in \cite{shi2023understanding} which considers the case where the encoders may not have the same weights. 

\section{Generalized Contrastive Alignment (GCA)}
\label{sec:gca}

In this section, we will introduce a new framework for {\em generalized contrastive alignment} and demonstrate the connections between contrastive learning and optimal transport.

\vspace{-2mm}
\subsection{Problem Formulation}
\vspace{-1mm}

Traditional contrastive learning methods focus on bringing positive examples, such as augmentations of the same sample, closer together in representation space. In contrast, our approach reframes contrastive learning as a distributional alignment problem, allowing flexible control over how pairs are matched by imposing specific constraints on the target transport plan, \(\mathbf{P}_{tgt}\).

Our objective is to learn an encoder \(f_\theta\) that minimizes the {\em transport cost} between positive samples. By defining \(\mathbf{P}_{tgt}\) with specific alignment rules, such as domain-specific or hierarchical constraints, we can influence how samples are organized in the latent space. For instance, setting \(\mathbf{P}_{tgt}\) to resemble a diagonal matrix encourages each positive to align primarily with itself or its augmentations, minimizing \(\text{div}({\bf I} || {\bf P}) \approx 0\), where \(\text{div}\) measures the deviation from an identity matrix (e.g., KL-divergence).

This flexibility allows us to encode more nuanced forms of similarity, adapting to tasks where alignment structure varies based on domain, class, or other high-level constraints. By expanding contrastive learning in this way, our method enhances separation of negatives while addressing complex relational patterns, making it suitable for a wider range of learning tasks.

\vspace{-2mm}
\paragraph{Defining the Kernel Space.}
Before formally stating our objective, we first need to define the concept of an augmentation kernel for our positive and negative examples.

\begin{definition}[Augmentation Kernel]
\label{def:gibbs}
Let $f_\theta$ denote an encoder with parameters $\theta$ and let $({\bf x}'_i,{\bf x}''_j) \sim \mathcal{A}$ be two views drawn from the family of augmentations $\mathcal{A}$. The augmentation kernel for the encoder $\theta$ is defined as ${\bf K}_\theta ( {\bf x}'_i , {\bf x}''_j ) = \exp(- \text{dist} (\widetilde{f}_\theta ({\bf x}'_i), \widetilde{f}_\theta ({\bf x}''_j))/\varepsilon)$, where $\text{dist}(\cdot)$ can be an arbitrary distance metric, and \(\widetilde{f}_\theta ({\bf x}'_i)\) is the normalized output of \(f_\theta\), and \(\varepsilon\) is the regularization parameter.
\end{definition}

\vspace{-2mm}
\paragraph{Main Objective.}

With this definition in hand, we can now formalize our objective as follows:
\begin{equation}\label{eq:mainobj}
    \min_{\theta} ~~ d_M \big( {\PP}_{\text{tgt}} || {\bf P}_\theta),~\text{with} \quad \PP_\theta= 
  \arg\min_{\PP \in \mathcal{B}}\{ h(\PP)+ d_\Gamma (\PP || \K_\theta)\},
\end{equation}

where ${\bf K}_\theta$ is the augmentation kernel defined in Definition~\eqref{def:gibbs}, \(h(x)\) is a convex function (typically an indicator function), $\mathcal{B}$ is a closed convex constraint set (i.e. Birkhoff polytope) that defines the constraints of proximal operators,  $d_{\Gamma}$ is a Bregman divergence that is used to find the nearest points \(\PP_\theta\) on the constraint set \(\mathcal{B}\) of \(\K_\theta\), \(d_M\) is a convex function (e.g., KL-divergence) that measures divergence between ${\bf P}_{\theta}$ and the target coupling plan \(\PP_{\text{tgt}}\). 

Our objective is a bi-level optimization problem which aims to learn a representation that minimizes the divergence between the transport plan \(\PP_\theta\) with the target alignment plan \({\PP}_{\text{tgt}}\) that encodes the matching constraints. When we consider a standard contrastive learning setup where we have pairs of positive examples the source and target distribution, then the target \({\PP}_{\text{tgt}}\) is the identity matrix \(\I\). However, we will show later that other alignment constraints can be considered. Moreover, when \(\mathcal{B}\) is the intersection of more constraint sets like \(C_1^\mu \cap C_2^\nu\) in Equation~\eqref{eq:Birkhoff}, a nature way to get the approximation of the nearest points \(\PP_\theta\) of \(\K_\theta\) is to run iterative projections algorithm~\cite{benamou2015iterative}, which could be extended into the intersection of several constraint sets like \(\{\cap_{i=1}^n C_i\}\), resulting in a multi-marginal problem~\cite{pass2015multi}.

\vspace{-2mm}
\subsection{A Proximal Point Algorithm for GCA }
\label{sec:multistep} 
\vspace{-1mm}
In practice, we can solve the alignment problem above by iteratively updating the two main components in our bi-level objective.  First, for a fixed encoder parameters $\theta$, we obtain the transport coupling $\PP_\theta$ through our corresponding proximal operator. 
Second, we measure the deviation between the transport plan $\PP_{\theta}$ with the target ${\PP}_{\text{tgt}}$ that encodes our matching constraints, which denotes the ideal alignment plan on the intersection of the constraint sets. 
We provide pseudocode for this iterative approach in Algorithm~\ref{alg:gca}, which we refer to as generalized contrastive alignment or GCA. The implementation of our methods is in 
\href{https://github.com/nerdslab/gca}{https://github.com/nerdslab/gca}.

\begin{algorithm}
\caption{~Proximal-Point Algorithm for Generalized Contrastive Alignment (GCA)\label{alg:gca}}
\begin{algorithmic}[1]
\State \textbf{Initialization:} Initial encoder parameters $\theta$, target transport plan ${\PP}_{\text{tgt}}$, kernel function \(\K_\theta\), the function \(h(x)\), divergences \(d_\Gamma\) and \(d_M\) (KL or \(W_1\)). Initialize transport plan ${\PP}_{\theta}$ based on $\theta$. 

\State \textbf{Compute the transport coupling ${\bf P}_\theta$:} Update ${\PP}_{\theta}$ using the proximal operator scaling for fixed $\theta$ as described in Eq.~\eqref{eq:mainobj}:
$$\PP_\theta= 
  \arg\min_{\PP \in \mathcal{B}}\{ h(\PP)+ d_\Gamma (\PP || \K_\theta)\}.$$ 
\State \textbf{Calculate the loss:} Calculate deviation between the target and current transport plans $$\mathcal{L}_{GCA}=d_M({\PP}_{\theta}, {\PP}_{\text{tgt}}).$$ 
Update networks $f_\theta$ (encoder) and $g_\theta$ (projector) to minimize $\mathcal{L}_{GCA}$. 
\State \textbf{Repeat until convergence:} Repeat steps 2 and 3 until convergence.

\end{algorithmic}
\end{algorithm}

Computing the transport coupling \(\PP_\theta\)~\footnote{With a single constraint set like \(C_1^\mu\) in Equation~\eqref{eq:Birkhoff}, computing the proximal point only involves a single projection. However, if there are intersecting constraint sets like \(C_1^\mu\cap C_2^\nu\), solving for the  proximal point requires multiple projections before we approach the nearest point on their intersection.} (forward-pass) in GCA algorithms could be treated as a specific type of Dykstra's projection algorithms~\cite{bregman1967relaxation}, which computes the \textbf{iterative projection} on the intersection of affine convex sets~\cite{benamou2015iterative,peyre2015entropic}. The proofs of convergence are provided in  Appendix~\ref{app:gca_converged}.

\subsection{GCA-UOT Method}
\label{subsec:relax}

\vspace{-1mm}

We can also benefit from the rich literature on optimal transport to build different relaxations of our objective~\cite{peyre2019computational, chen2021wasserstein, taherkhani2021self, lin2021making, montesuma2023recent}. 
In particular, we choose to leverage a formulation of {\em unbalanced optimal transport} (UOT) to further relax the marginal constraints~\cite{chizat2018scaling} in our objective.

In this case, we can add the dual form of \(d_\Gamma\) to the Equation~\eqref{eq:mainobj} and reformulate our objective as:
\begin{equation}
\min_{\theta}~ d_M ( \bf P_{\text{tgt}} \| \bf P_\theta)  + \lambda_1 h_{\mathcal{F}}(\bf P_\theta\mathbbm{1} || \mu) + \lambda_2 h_{\mathcal{G}}(\bf P_\theta^\top\mathbbm{1} || \nu) +\varepsilon H(\bf P_\theta).
\label{eq:uot}
\end{equation} 
Here \( h_{\mathcal{F}} \) and \( h_{\mathcal{G}} \) can be different divergence measures (e.g., KL divergence) that penalize deviations from the desired marginals \( \mu \) and \( \nu \), and \( \lambda_1 \) and \( \lambda_2 \) are regularization parameters that control the trade-off between the transport cost and the divergence penalties. This relaxation leads to different types of proximal operators which we outline in Appendix~\ref{app:gcauot}. The impact of the entropy regularization parameter \(\varepsilon\) on the coupling matrix is studied in Figure~\ref{fig:epsilon_study}, along with the number of iterations and corresponding sensitivity is provided in Figure~\ref{fig:sensitivity_study}.

\subsection{Modifying the Target Transport Plan to Encode Matching Constraints}\label{sec:ptgt}
Contrastive learning objectives can be cast as a minimization of the  deviations between the transport plan ${\bf P}_{\theta}$ and the identity matrix, i.e., ${\bf P}_{tgt} = {\bf I}$.
However, our \ours~formulation enables learning representations that extend beyond this one-to-one matching constraint. This flexibility allows us to incorporate additional matching constraints informed by domain-specific knowledge. For example, in domain generalization scenarios \cite{gulrajani2020search,kim2021selfreg}, where each batch contains samples from multiple domains, the target alignment plan can be structured as:
\[
{\bf P}_{\text{tgt}}[i,j] = {\bf I}[i,j] + \alpha \cdot \mathbb{I}(D_i = D_j, \, i \neq j) + \beta \cdot \mathbb{I}(D_i \neq D_j, \, i \neq j),
\]
Where \(\mathbb{I}(\cdot)\) is the indicator function, which equals 1 if the condition inside is true and 0 otherwise. \(D_i\) represents the domain of sample \(i\), where $\alpha \geq 0$ and $\beta \geq 0$. In this case, we can improve the representation by building the block constraints which encode either class information (in supervised setting) or domain information (in across domain generalization, visualized in Figure~\ref{fig:Ptgt}).

\subsection{Computational Complexity}
The forward-pass only involves the scaling operations in Equation~\eqref{eq:uvupdates} and doesn't affect the complexity of the backward-pass. Therefore, GCA methods can be thought of as a form of batch normalization operations with adaptive scaling. An analysis of the complexity is provided along with experiments in  Appendix~\ref{app:gca_converged}. Our results show that GCA iterations only slightly increase the computational complexity when compared with their single step equivalent (GCA-INCE vs. INCE). However, we found that GCA-UOT is faster than INCE due to the improved symmetry and smoothness of the loss. Moreover, we record the floating point operations per second (Flops) of running GCA methods. We find that GCA-INCE (6.65 MFlops) has $5\%$ more Flops than INCE (6.31 MFlops), while GCA-UOT saves $30\%$ Flops (4.54 MFlops).  These results show that our GCA-UOT method is not only superior in terms of accuracy but also in speed.

\section{Building Connections to Different CL Objectives}
\label{sec:connectionsCL}
\begin{wraptable}{ht}{0.49\textwidth}
  \begin{center}
    \vspace{-7mm}
    \caption{\footnotesize {\em Comparison of different contrastive alignment objectives.} Here we have $C_1^\mu$ and $C_2^\nu$ as constraint sets (denoted as \(\mathcal{B}\)) defined in Equation~\eqref{eq:Birkhoff} with their corresponding indicator function. "Iter" refers to iterative methods.}
    \vspace{-2mm}
    \resizebox{0.49\textwidth}{!}{
      \begin{tabular}{l|c|c|c|c} 
        \footnotesize{$\textbf{Methods}$} & \footnotesize{$d_M$} &  \footnotesize{$d_\Gamma$}    & \(\mathcal{B}\) & Iter\\
        \hline
        \footnotesize{INCE} & KL & KL  &\(C_1^\mu\) &    \\
        \footnotesize{GCA-INCE} & KL & KL & \(C_1^\mu\cap C_2^\nu\) & $\checkmark$ \\
        \footnotesize{RINCE (q=1)} & W1 & KL &\(C_1^\mu\) &   \\
        \footnotesize{GCA-RINCE (q=1)} & W1 & KL & \(C_1^\mu\cap C_2^\nu\) & $\checkmark$  \\
        \footnotesize{BYOL} & KL & L2 & \(R^{B\times B}\) &   \\
      \end{tabular}
    }
    \label{tab:diff}
    \vspace{-7mm}
  \end{center}
\end{wraptable}

In this section, we show how the modification of the different parts of our main objective (\(d_\Gamma, d_M, \mathcal{B}, \K_\theta\)) in Equation~\eqref{eq:mainobj} can be connected to different contrastive losses. See  Table~\ref{tab:diff} for a summary of how different losses can be mapped back to our formulation. 

\vspace{-2mm}
\subsection{Connection to INCE}
\vspace{-1mm}

An interesting connection that we can make between GCA main objective and contrastive learning is that we can interpret INCE as a \textbf{single step} in a iterative GCA objective~\cite{shi2023understanding}. 
This connection can be further summarized through the following theorem.
\begin{theorem}[INCE Equivalence]
Let ${\bf K}_\theta$ denote the augmentation kernel as in Definition~\eqref{def:gibbs} with cosine similarity, $d_\Gamma$ and $d_M$ equal to KL-divergence, and constraint set as \(C_1^\mu\) in Equation~\eqref{eq:Birkhoff}.
The INCE objective in Equation~\eqref{eq:INCE} can be re-expressed as a GCA problem in Equation~\eqref{eq:mainobj} as follows:
\begin{equation}\label{eq:kleqince}
\min_{\theta}\text{KL} \big( {\bf I}  ||\text{Prox}_{C_1^\mu}^{KL}(\K_\theta)).\end{equation}
\label{thm:cl_eq_ot}
\end{theorem}
The proof is contained in Appendix \ref{app:proof_cleqot}. Theorem~\eqref{thm:cl_eq_ot} shows that the INCE loss can be viewed as solving the matching problems in Equation~\eqref{eq:eot} with  row normalization constraints \(C_1^\mu\). 
This connection between GCA and INCE allows us to derive the iterative algorithm for GCA-INCE by running Bregman projection iteratively on both row and column normalization sets.

\vspace{-2mm}
\subsection{Connection to RINCE}
We introduce the following result to build the connection between our framework and RINCE~\cite{chuang2022robust}.
\begin{theorem}[RINCE Equivalence]\label{thm:rince}
Let ${\bf K}_\theta$ denote the augmentation kernel as in Definition~\eqref{def:gibbs}. Set target plan \(\PP_{\text tgt}=\bf I\), $d_\Gamma$ equal to the KL-divergence, \(d_M(\I\|\PP)=-\frac{1}{q}(\frac{\operatorname{diag}(\PP_\theta)}{\uu})^q+\left(\frac{\lambda \I}{\uu}\right)^q\) with \(\lambda\), \(q\), and \(\uu=\operatorname{diag}\left(\frac{\mu}{\PP^{(0)}\mathbbm{1}}\right)\), and constraint set \(C_1^\mu\) defined in Equation~\eqref{eq:Birkhoff}. The RINCE objective in Equation~\eqref{eq:RINCE} can be re-expressed as a GCA problem as follows:
\begin{equation}\label{eq:hsrince}
     \min_\theta d_M(\I\|\PP_\theta),~~\text{with}~~\PP_\theta = \text{Prox}_{C_1^\mu}^{\text{KL}}(\K_\theta),
\end{equation} 
\end{theorem}
 
\vspace{-4mm}
The proof is provided in Appendix~\ref{app:rinceeqprox}. As we can see, RINCE introduces adjustable parameters \( q \) and \( \lambda \), with \( \lambda \) controlling the weight of negative samples, while \( q \in (0,1]\) serves to switch between KL divergence and Wasserstein discrepancy. When $q=1$, we have the following theorem:
\begin{theorem}[W1 Equivalence]
\label{co:rince}

Let ${\bf K}_\theta$ denote the augmentation kernel as in Definition~\eqref{def:gibbs} with cosine similarity. Set target plan \(\PP_{\text tgt}=\bf I\), \(d_\Gamma\) equal to the KL-divergence, $d_M$ equal to the 1-Wasserstein distance $(W_1)$ in Definition~\eqref{def:wdm}, and the constraint set as \(C_1^\mu\) defined in Equation~\eqref{eq:Birkhoff}. The RINCE object in Equation~\eqref{eq:RINCE} with $q=1$ can be re-expressed as a GCA problem as follows:
\begin{equation}\label{eq:w1eqrince}
\min_{\theta} W_1 \big( {\PP_{tgt}} ||\text{Prox}_{C_1^\mu}^{KL}(\K_\theta)).\end{equation}
\end{theorem}
\vspace{-3mm}
See Appendix~\ref{app:wdm} for the proof. 
This connection to RINCE suggests an extended iterative formulation to calculate the coupling plan as the projection point \(\PP^{(\infty)}=\text{Prox}_{C_1^\mu\cap C_2^\nu}^{\text{KL}}(\K_\theta)\) of \(\K_\theta\) on the constraint set \(C_1^\mu\cap C_2^\nu\). In this case, we can write an iterative algorithm for robust alignment called GCA-RINCE as follows:
\begin{equation}\label{eq:gcarince}
    L_{\text{GCA-RINCE}}^{\lambda,q} =\min_\theta -q^{-1}( \operatorname{diag}(\PP_\theta^{(2t-1)} )/{\uu^{(t)}})^q+ q^{-1}( \lambda \PP_{tgt}/\uu^{(t)})^q, 
\end{equation}
where $\lambda$ and $q$ are hyperparameters, $\PP^{(1)} \coloneqq \operatorname{diag}(\uu^{(1)})\K_\theta\operatorname{diag}(\vv^{(0)})$, and $t$ is the number of iterations.

\vspace{-2mm}
\subsection{Connection to BYOL}
\vspace{-1mm}
Our framework also allows us to make connections to BYOL \cite{grill2020bootstrap}.
BYOL learns by encouraging similarity between positive image pairs, without explicitly conditioning on negative examples. To build this connection, recall that BYOL has the online network parameterized by \(\theta\) and target network parameterized by \(\xi\), where \(\z'_\theta=\widetilde{f}_{\theta}(\x')\) and \(\z''_\xi=\widetilde{f}_{\xi}(\x'')\) are the normalized outputs of the online and target networks, respectively. A simplified version of the BYOL loss can be written as: $\label{eq:byol}L_\text{BYOL}=  \| \widetilde{q}_{\theta}(\z'_{\theta}) - \z''_{\xi} \|_2^2,
$
where \(\widetilde{q}_{\theta} (\z'_\theta)\) is the normalized output after online network and \(q_\theta\) is the predictor.\footnote{In practice, BYOL also switches the order of views to symmetrize the loss. For ease of discussion, we consider just one pair of views but the same could be argued for the full symmetric version.} In this case, we can provide the following connection between \ours and BYOL as follows.

\begin{theorem}[BYOL Equivalence]\label{thm:byol}
Let \(\bS_{\theta}(\x'_i, \x''_j) = \exp ( - \| \widetilde{q}_{\theta}({\bf z}_i') -  {\bf z}_j'' \|) \) denote the augmentation kernel. Set the target plan \(\PP_{\text tgt}=\bf I\), \(d_\Gamma\) equal to the L2-distance, $d_M$ equal to the KL-divergence, and constraint set as \(R^{B\times B}\). The BYOL objective can be re-expressed as a GCA problem as follows:
\begin{equation}\label{eq:eqbyol}
\min_\theta \text{KL} \big(  {\bf I} || \bS_\theta),~~\text{with}~~\bS_\theta=\text{Prox}_{R^{B\times B}}^{\|\cdot\|}(\bS_\theta).\end{equation}
\end{theorem}
See the proof in Appendix~\ref{app:byol}. 

\section{Theoretical Analysis}\label{sec:theoyanaylsis}

In this section, we aim to show how the GCA-methods can improve alignment and uniformity in the latent space~\cite{wang2020understanding}. 
Here, {\em alignment} means that the features of the positive samples are as close as possible, while {\em uniformity} means that the features of negative samples are uniformly distributed on latent space (see  Appendix~\ref{app:alignment} for formal definitions). These quantities have been studied in a number of related works~\cite{wang2020understanding, pu2022alignment}, where one can show that improved alignment and uniformity can lead to different  benefits in representation learning.

\subsection{Improved alignment with GCA}
\label{subsec:betterERM}
\vspace{-1mm}
Contrastive learning minimizes the deviation between the target alignment plan with the transport plan in Definition~\ref{def:gibbs} through empirical risk minimization (ERM). Therefore, a tighter bound on the empirical risk corresponds to a smaller difference between the ideal alignment with the coupling matrix. We show that this in turn leads to better alignment of the positive views.

\vspace{-2mm}
\paragraph{Analysis of INCE vs GCA-INCE.}GCA-INCE ensures that the final transport plan \(\PP^{(\infty)}\) is closer to the ideal identity matrix compared to the INCE, as we show in the following theorem.

\begin{theorem}[Improved Alignment with INCE]
\label{thm:lowest}

Let ${\bf K}_\theta$ denote the augmentation kernel as in Definition~\eqref{def:gibbs}. Set $d_M$ and $d_\Gamma$  to the KL-divergence, and  \({\PP}_{\text{tgt}}=\bf I\). The GCA-INCE loss with converged plan \(\PP_\theta^{(\infty)}\) is lower than the GCA-INCE loss with $\PP_\theta^{(t)}$ in Equation~\eqref{eq:sinkp} for all $t$. 
\end{theorem}
The full proof is provided in Appendix~\ref{app:klfg}. The above theorem tells us that solving Equation~\eqref{eq:mainobj} with iterative projection will converge to a transport plans $\PP_\theta^{(\infty)}$ with lower KL divergence than the one-step solution provided by INCE. We can establish the convergence of the \(\PP^{(t)} \rightarrow \PP^{(\infty)}\), based on the convergence of Bregman projection.

\vspace{-2mm}
\paragraph{Analysis of RINCE vs GCA-RINCE.}~

GCA also benefits from other Bregman divergences, like the WDM in RINCE, which provides robustness against distribution shift compared to the KL-divergence in INCE. GCA-RINCE provides a lower bound on the RINCE loss in Equation~\eqref{eq:RINCE}, which allows us to develop a tighter bound with \(\PP^{(\infty)}\) obtained by several proximal steps with GCA.

\begin{theorem}[Improved Alignment with RINCE]\label{thm:gcarince}
GCA-RINCE loss with $\PP_\theta^{(t)}$ in Equation~\eqref{eq:gcarince}  is lower than the loss in the Theorem~\eqref{thm:rince} as
 \( L_{\text{GCA-RINCE}}^{\lambda,q=1}(\PP_\theta^{(t)}) \leq  L_{\text{RINCE}}^{\lambda,q=1}(\PP_\theta^{(1)})\).
\end{theorem}

See Appendix~\ref{app:klfg} for the full proof and an analysis of GCA methods  for different choices of \(d_M\).

\vspace{-2mm}
\subsection{Improved Uniformity of Representations Through GCA}
\vspace{-1mm}
\label{subsec:align_uni}

The improved alignment of GCA-methods comes from maximization of the uniformity under the constraint of intersection \(C_1^\mu \cap C_2^\nu\) in Equation~\eqref{eq:Birkhoff}, rather than the constraint set \(C_1^\mu\) in INCE (see Table~\ref{tab:diff}). Finding the projection of \(\K_\theta\) on set of \(C_1^\mu \cap C_2^\nu\) through proximal steps is equivalent to solving the dual problem of EOT, which can be summarized through the following theorem.
\begin{theorem}[Improved Uniformity]\label{thm:uniformity}
Given the constraint sets in Equation~\eqref{eq:Birkhoff}, the optimal transport coupling upon convergence of Equation~\eqref{eq:sinkp}, denoted as \(\PP^{(\infty)}\), achieves a higher uniformity loss compared to the single-step transport plan \(\PP^{(1)}\) obtained by INCE.  
\vspace{-1mm}
\end{theorem}
The proof is provided in the Appendix~\ref{app:uniformity}. Through loss propagation, we show that the alignment plan offered by \(\PP^{(\infty)}\) will guide the subsequent iterations towards more uniform representations.

\vspace{-2mm}
\subsection{Impacts of GCA on a downstream classification task}
\vspace{-1mm}
We take this one step further and examine the impact of GCA on a downstream classification task.
For a classification task, using a labeled dataset $\mathcal{D} = \{(\bar{\x}_i, \y_i)\} \in \bar{\mathcal{X}} \times \mathcal{Y}$ where $\mathcal{Y} = [1,\dots,M]$ with $M$ classes, we consider a fixed, pre-trained encoder $f_\theta \in \mathcal{F}: \mathcal{X} \to \mathcal{S}$. Assume that positive and negative views of $n$ original samples $(\bar{\x}_i)_{i \in [1..n]} \subset \bar{\mathcal{X}}$ are sampled from the data distribution $p(\bar{\x})$.

In this case, the uniformity loss is equivalent to optimizing the downstream supervised classification tasks with cross-entropy (CE) loss when  the following two assumptions are satisfied~\cite{dufumier2023integrating}.
\begin{assumption} [Expressivity of the Encoder]\label{asm:express_encoder}
Let us define $\mathcal{H}_{\bar{\mathcal{X}}}$ is the RKHS associated with the kernel $\K_{\bar{\mathcal{X}}}$ defined on \(\bar{\mathcal{X}}\), and \( (\mathcal{H}_{f_\theta}, \K_\theta) \) defined on $\mathcal{X}$ with augmentation kernel $\K_\theta = \langle f_\theta(\cdot), f_\theta(\cdot) \rangle_{\mathbb{R}^d}$ in Definition~\ref{def:gibbs}. And we assume that $\forall g \in \mathcal{H}_{f_\theta}, \ \mathbb{E}_{\mathcal{A}(x|\cdot)} g(x) \in \mathcal{H}_{\bar{\mathcal{X}}}$.
\end{assumption}

\begin{assumption} [Small Intra-Class Variance]\label{asm:small_intra_variance}
For \(y \neq y'\), the intra-class variance \(\delta_i,\delta_j\) are negligible compared to the distance among different class centroids, \(\mu_y, \mu_{y'}\) as \(\|\mu_y - \mu_{y'}\| \gg \| \delta_i-\delta_j\| \).
\end{assumption}
\begin{claim}\label{cl:ce_eq_uniform}
If Assumption~\ref{asm:express_encoder} and Assumption~\ref{asm:small_intra_variance} hold, then maximizing the uniformity is equivalent to minimizing the downstream CE loss.
\end{claim}
The proof is provided in Appendix~\ref{app:uniformity}. Optimizing the self-supervised loss under ideal conditions improves downstream CE tasks and helps to explain why maximizing uniformity aids classification.

{\bf Remark.}. Maximizing uniformity can enhance downstream classification but risks ``feature suppression'' by encouraging shortcut features that harm generalization~\cite{robinson2021can}. 
In GCA-UOT, adding penalties modifies the transport plan from that of a pure uniformity loss, helping to avoid feature suppression. We find empirical evidence that UOT provides a more robust transport plan which appears to circumvent some of these shortcut features from being learned (Figure~\ref{fig:diagnolline} in Appendix~\ref{app:gcauot-feature}).
\section{Experiments}
\label{sec:exp}

In this section, we conduct empirical evaluations to study the performance of our approach in both handling noisy and corrupted views and in domain generalization tasks. 

\vspace{-1mm}
\subsection{Comparison with CL Baselines}
\vspace{-1mm}

\paragraph{Experiment setup.}~To examine the robustness of our framework, we trained INCE and RINCE as baselines, and developed their GCA-based alternatives (+\ours). In addition, we also compared with our novel GCA-UOT method, two variants of IOT established in~\cite{shi2023understanding},  and other CL baselines, including BYOL and SimCLR. For experiments with SVHN \cite{netzer2011reading} and ImageNet100 \cite{deng2009imagenet} we use the ResNet-50 encoder as the backbone and use a ResNet-18 encoder as the backbone for CIFAR-10, CIFAR-100~\cite{krizhevsky2009learning} and a corrupted version of CIFAR called CIFAR-10C \cite{hendrycks2019benchmarking}.

\begin{table}[t!]
  \begin{center}
    \resizebox{\textwidth}{!}{
    \begin{tabular}{|r|c|c|c|c||c|c|c|c|}
    \hline
    & \multicolumn{4}{c||}{\footnotesize{\bf Standard Setting}}
    & \multicolumn{4}{c|}{\footnotesize{\bf Noisy Setting}} \\
    \hline
    \footnotesize{Method}
    & \footnotesize{CIFAR-10}
    & \footnotesize{CIFAR-100}
    & \footnotesize{SVHN}
    & \footnotesize{ImageNet100}
    & \footnotesize{CIFAR-10 (Ex)}
    & \footnotesize{CIFAR-100 (Ex)}
    & \footnotesize{CIFAR-10C}
    & \footnotesize{CIFAR-10C (Ex)} \\
    \hline

    \footnotesize{INCE}
    & 92.01 {\footnotesize ± 0.40}        
    & 71.09 {\footnotesize ± 0.31}        
    & 92.42 {\footnotesize ± 0.24}        
    & 73.01 {\footnotesize ± 0.61}        
    & 82.03 {\footnotesize ± 0.32}        
    & 62.54 {\footnotesize ± 0.20}        
    & 81.52 {\footnotesize ± 1.04}        
    & 83.28 {\footnotesize ± 0.25}        
    \\
    
    \footnotesize{GCA-INCE}
    & \underline{93.02 {\footnotesize ± 0.19}}  
    & \underline{71.55 {\footnotesize ± 0.12}}        
    & \underline{92.64 {\footnotesize ± 0.26}}
    & \underline{73.04 {\footnotesize ± 0.76}}
    & \underline{82.18 {\footnotesize ± 0.69}}
    & \underline{62.65 {\footnotesize ± 0.17}}
    & \underline{82.63 {\footnotesize ± 0.28}}
    & \underline{82.72 {\footnotesize ± 0.27}}
    \\
    \footnotesize{$\Delta$}
    & +1.01
    & +0.46            
    & +0.22
    & +0.03
    & +0.15
    & +0.11
    & +1.11
    & -0.56
    \\
    \hline
     
    \footnotesize{RINCE}
    & 93.27 {\footnotesize ± 0.20}
    & 71.63 {\footnotesize ± 0.36}        
    & 93.26 {\footnotesize ± 0.15}
    & 71.91 {\footnotesize ± 0.43}
    & 82.60 {\footnotesize ± 0.63}
    & 63.55 {\footnotesize ± 0.14}
    & 82.86 {\footnotesize ± 0.21}
    & 83.64 {\footnotesize ± 0.26}
    \\
    
    \footnotesize{GCA-RINCE}
    & \underline{93.47 {\footnotesize ± 0.30}}
    & \underline{71.95 {\footnotesize ± 0.48}}  
    & \underline{93.57 {\footnotesize ± 0.26}}
    & \underline{73.44 {\footnotesize ± 0.55}}
    & \underline{82.76 {\footnotesize ± 0.49}}
    & \underline{63.14 {\footnotesize ± 0.41}}
    & \underline{82.87 {\footnotesize ± 0.11}}
    & \underline{\textbf{83.69 {\footnotesize ± 0.16} }}
    \\
    \footnotesize{$\Delta$}
    & +0.20
    & +0.32            
    & +0.31
    & +1.53
    & +0.16
    & -0.47
    & +0.01
    & +0.05
    \\
    \hline
    
    \footnotesize{SimCLR}
    & 92.07 {\footnotesize ± 0.90}
    & 70.85 {\footnotesize ± 0.50}        
    & 92.13 {\footnotesize ± 0.34}
    & 72.20 {\footnotesize ± 0.78}
    & 81.87 {\footnotesize ± 0.53}
    & 62.94 {\footnotesize ± 0.13}
    & 81.74 {\footnotesize ± 1.54}
    & 83.25 {\footnotesize ± 0.18}
    \\
      
    \footnotesize{BYOL}
    & 90.56 {\footnotesize ± 0.59}
    & 69.75 {\footnotesize ± 0.37}        
    & 89.50 {\footnotesize ± 0.46}
    & 69.75 {\footnotesize ± 0.83}
    & 81.55 {\footnotesize ± 0.50}
    & 62.11 {\footnotesize ± 0.25}
    & 82.43 {\footnotesize ± 0.06}
    & 70.09 {\footnotesize ± 0.34}
    \\
    
    \footnotesize{IOT \cite{shi2023understanding}}
    & 92.09 {\footnotesize ± 0.22}
    & 68.37 {\footnotesize ± 0.42}        
    & 92.25 {\footnotesize ± 0.19}
    & 72.27 {\footnotesize ± 0.53}
    & 80.59 {\footnotesize ± 0.64}
    & 62.69 {\footnotesize ± 0.34}
    & 82.01 {\footnotesize ± 0.80}
    & 81.57 {\footnotesize ± 0.83}
    \\
    
    \footnotesize{IOT-uni \cite{shi2023understanding}}
    & 91.49 {\footnotesize ± 0.11}
    & 68.62 {\footnotesize ± 0.35}        
    & 92.33 {\footnotesize ± 0.11}
    & 72.88 {\footnotesize ± 0.71}
    & 80.79 {\footnotesize ± 0.24}
    & 62.56 {\footnotesize ± 0.22}
    & 81.19 {\footnotesize ± 1.12}
    & 81.82 {\footnotesize ± 0.51}
    \\
    
    \hline
    \footnotesize{GCA-UOT}
    & \textbf{93.50 {\footnotesize ± 0.31}}
    & \textbf{72.16 {\footnotesize ± 0.38}}  
    & \textbf{93.82 {\footnotesize ± 0.17}}
    & \textbf{74.09 {\footnotesize ± 0.40}}
    & \textbf{83.18 {\footnotesize ± 0.44}}
    & \textbf{63.62 {\footnotesize ± 0.27}}
    & \textbf{82.90 {\footnotesize ± 0.50}}
    & 83.64 {\footnotesize ± 0.19}
    \\
    \hline

    \end{tabular}
    }
  \caption{\footnotesize{{\em Test accuracy (\%) on a downstream classification task after pretraining.} 
    Results are shown for CIFAR-10 (ResNet18), CIFAR-100 (ResNet18), 
    SVHN (ResNet18), and ImageNet100 (ResNet50) under standard 
    and extreme (Ex) augmentation conditions (averaged over 5 seeds). 
    The {\bf CIFAR-100} column is now updated with the latest values from
    Table~\ref{tab:cifar1004c}. The top model is in bold and the second-place 
    model is underlined. For INCE and RINCE, we also provide the 
    improvement $\Delta$ by adding GCA to each method.}}
    \label{tab:new_combine}
  \end{center}
  \vspace{-8mm}
\end{table}

In all of these cases, we follow the standard self-supervised learning evaluation protocol \cite{chen2020simple}, where we train the encoder on the training set in an unsupervised manner and then train a linear layer on top of the frozen representations to obtain the final accuracy on the test set.
In addition to standard data augmentation policies commonly used \cite{chuang2022robust}, we also apply three different extreme augmentation policies to examine the robustness of \ours~towards noisy views (details in Appendix~\ref{app:corruptset}). Learning rates and other training details for CIFAR-10, CIFAR-100, SVHN, and ImageNet100 are provided in Appendix~\ref{app:hpdetails}, while specific training details for CIFAR-10C are included in Appendix~\ref{app:corruptset}.

\vspace{-3mm}
\paragraph{Results on Standard Augmentations.}~ 
First, we performed experiments on CIFAR-10, CIFAR-100, SVHN, and ImageNet100 using standard sets of augmentations that are applied to achieve state-of-the-art performance (Table~\ref{tab:new_combine}, Standard Setting). 
We found the +GCA versions of INCE and RINCE exhibit performance gains in almost all settings except for SVHN, with bigger gains observed when adding GCA to RINCE. Additionally, we find that our unbalanced OT method, GCA-UOT, achieves the top performance across the board, on all four datasets tested. The transport plans obtained by each methods are provided in Figure~\ref{fig:diagnolline} along with a study of the sensitivity of the methods to hyperparameters (Appendix~\ref{fig:ablation_study}).

\vspace{-3mm}
\paragraph{Results on Corrupted Data and Extreme Augmentations.}~
Next, we tested the methods in two noisy settings. In the first set of experiments, we apply extreme augmentations to CIFAR-10 (Ex) and CIFAR-100 (Ex) (see Appendix~\ref{app:corruptset}) to introduce noisy views during training. In the second set of experiments, we used the CIFAR-10C to further test the ability of our method to work in noisy settings.

Our experimental results demonstrate that the GCA-based strategy effectively enhances the model's generalization ability and adaptability to aggressive data augmentations. 
In addition to improving classification accuracy, the GCA-based methods also improve the representational alignment and uniformity, as shown in Appendix~\ref{app:representation}. This observation is in line with our theoretical analysis in Section~\ref{subsec:align_uni}, where we show that the obtained representations provide better overall alignment of positive views and better spread in terms of uniformity \cite{wang2020understanding}.

\vspace{-1mm}
\subsection{Block Diagonal Transport in Domain Generalization}
\vspace{-1mm}
\label{sec:domaingen}

\begin{wrapfigure}{r}{0.38\textwidth} 
\vspace{-8mm}
    \centering{
       \includegraphics[width=0.54\linewidth, clip, trim=0 1.3cm 25cm 0cm]{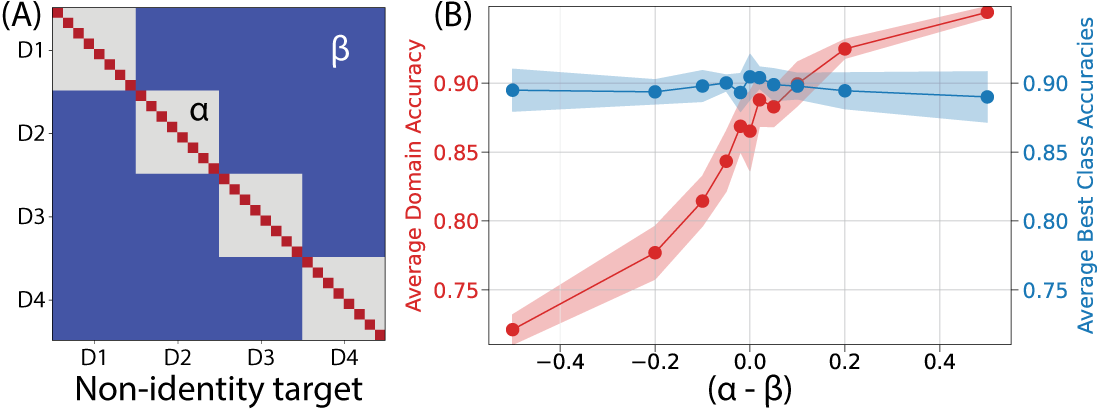}

 \includegraphics[width=\linewidth, clip, trim=13.7cm 0 0 0]{files/imgs/figure_v3.png}}
        \caption{\footnotesize{ {\em Incorporating different priors into learning across multiple domains.}} (A) Example target alignment plan \(\PP_{\text tgt}\), where the target over all samples from the same domain are set to \( \alpha \), the diagonal values are set to 1, and across-domain samples are set to \(\beta\). (B) The domain classification accuracy (red) and overall class accuracy (blue) with (\(\alpha-\beta\)) increases.}
        \vspace{-7mm}
        \label{fig:Ptgt}

\end{wrapfigure}

In a final experiment, we aimed to demonstrate the flexibility and robustness of our framework by applying it to a domain generalization task, where samples originate from different domains (e.g., Photo, Cartoon, Sketch, Art). We explored the effects of introducing domain-specific alignment constraints in our transport plan, hypothesizing that this could enhance the latent space organization to capture more nuanced domain similarities.

Our approach enables additional contextual information to be seamlessly integrated into the transport process. In this case, domain information was incorporated to distinguish the alignment of samples from the same versus different domains. To achieve this, we adjusted the target transport plan \({\bf P}_{tgt}\), selectively modifying parameters \((\alpha, \beta)\) to vary the influence of domain-based alignment constraints as shown in Figure~\ref{fig:Ptgt}(A). Specifically, we set \(\{\alpha=0, \beta>0\}\) to prioritize cross-domain alignment and \(\{\alpha>0, \beta=0\}\) to focus on intra-domain alignment.

The training was conducted on the PACS dataset \cite{li2017deeper} using a ResNet-18 encoder with the \ours-INCE objective. After training the encoder in an unsupervised manner, we freeze the encoder and then train a linear readout layer to predict either the sample’s class or the domain it belonged to. This setup allowed us to isolate the effect of our transport adjustments on the latent space’s capacity to encode both class and domain information.

The results, displayed in Figure~\ref{fig:Ptgt}(B), revealed that increasing the domain alignment weight enhances the accuracy of domain classification (from 72.11\% to 95.16\%) without diminishing classification performance. This outcome suggests that \ours~can effectively encode both domain and class information in a single latent representation. The ability to adjust alignment constraints provides a powerful tool for domain generalization tasks, enabling multiple types of similarity to be jointly encoded. This flexibility can potentially alleviate issues related to information loss from data augmentation, especially in fine-grained classification settings, by retaining essential domain-specific characteristics across transformations.

\section{Conclusion}
In this work, we introduced \emph{generalized contrastive alignment} (GCA), a flexible framework that redefines contrastive learning as a distributional alignment problem using optimal transport to control alignment.
By allowing targeted control over alignment objectives, GCA demonstrates strong performance across both standard and challenging settings, such as noisy views and domain generalization tasks. This work opens up broader possibilities for learning robust representations in real-world scenarios, where data is often diverse, noisy, or comes from multiple domains.

Future work includes applications of GCA to graphs  and time series data, as well as multi-modal settings where our approach can integrate various forms of similarity. As alignment strategies become integral to contrastive learning, GCA offers a promising foundation for more adaptive and expressive self-supervised models.

\section*{Acknowledgements}
We would like to thank Mehdi Azabou, Divyansha, Vinam Arora, Shivashriganesh Mahato, and Ian Knight for their valuable feedback on the work. This work was funded through NSF IIS-2212182, NSF IIS-2039741, and the support from the Canadian Institute for Advanced Research (CIFAR). We would also like to acknowledge the use of ChatGPT for providing useful feedback and suggestions on the writing of the paper.

\bibliographystyle{plain}
\bibliography{main}

\begin{thebibliography}{10}

\bibitem{an2022efficient}
Dongsheng An, Na~Lei, Xiaoyin Xu, and Xianfeng Gu.
\newblock Efficient optimal transport algorithm by accelerated gradient descent.
\newblock In {\em Proceedings of the AAAI Conference on Artificial Intelligence}, volume~36, pages 10119--10128, 2022.

\bibitem{arjovsky2017wasserstein}
Martin Arjovsky, Soumith Chintala, and L{\'e}on Bottou.
\newblock Wasserstein generative adversarial networks.
\newblock In {\em International conference on machine learning}, pages 214--223. PMLR, 2017.

\bibitem{benamou2015iterative}
Jean-David Benamou, Guillaume Carlier, Marco Cuturi, Luca Nenna, and Gabriel Peyr{\'e}.
\newblock Iterative bregman projections for regularized transportation problems.
\newblock {\em SIAM Journal on Scientific Computing}, 37(2):A1111--A1138, 2015.

\bibitem{berman2020sinkhorn}
Robert~J Berman.
\newblock The sinkhorn algorithm, parabolic optimal transport and geometric monge--amp{\`e}re equations.
\newblock {\em Numerische Mathematik}, 145(4):771--836, 2020.

\bibitem{bregman1967relaxation}
Lev~M Bregman.
\newblock The relaxation method of finding the common point of convex sets and its application to the solution of problems in convex programming.
\newblock {\em USSR computational mathematics and mathematical physics}, 7(3):200--217, 1967.

\bibitem{cai2022developments}
Xing-Ju Cai, Ke~Guo, Fan Jiang, Kai Wang, Zhong-Ming Wu, and De-Ren Han.
\newblock The developments of proximal point algorithms.
\newblock {\em Journal of the Operations Research Society of China}, 10(2):197--239, 2022.

\bibitem{chen2021wasserstein}
Liqun Chen, Dong Wang, Zhe Gan, Jingjing Liu, Ricardo Henao, and Lawrence Carin.
\newblock Wasserstein contrastive representation distillation.
\newblock In {\em Proceedings of the IEEE/CVF conference on computer vision and pattern recognition}, pages 16296--16305, 2021.

\bibitem{chen2020simple}
Ting Chen, Simon Kornblith, Mohammad Norouzi, and Geoffrey Hinton.
\newblock A simple framework for contrastive learning of visual representations.
\newblock In {\em International conference on machine learning}, pages 1597--1607. PMLR, 2020.

\bibitem{chen2021exploring}
Xinlei Chen and Kaiming He.
\newblock Exploring simple siamese representation learning.
\newblock In {\em Proceedings of the IEEE/CVF conference on computer vision and pattern recognition}, pages 15750--15758, 2021.

\bibitem{chen2023instance}
Zining Chen, Weiqiu Wang, Zhicheng Zhao, Fei Su, Aidong Men, and Yuan Dong.
\newblock Instance paradigm contrastive learning for domain generalization.
\newblock {\em IEEE Transactions on Circuits and Systems for Video Technology}, 34(2):1032--1042, 2023.

\bibitem{chizat2018scaling}
Lenaic Chizat, Gabriel Peyr{\'e}, Bernhard Schmitzer, and Fran{\c{c}}ois-Xavier Vialard.
\newblock Scaling algorithms for unbalanced optimal transport problems.
\newblock {\em Mathematics of Computation}, 87(314):2563--2609, 2018.

\bibitem{chuang2022robust}
Ching-Yao Chuang, R~Devon Hjelm, Xin Wang, Vibhav Vineet, Neel Joshi, Antonio Torralba, Stefanie Jegelka, and Yale Song.
\newblock Robust contrastive learning against noisy views.
\newblock In {\em Proceedings of the IEEE/CVF Conference on Computer Vision and Pattern Recognition}, pages 16670--16681, 2022.

\bibitem{chuang2020debiased}
Ching-Yao Chuang, Joshua Robinson, Yen-Chen Lin, Antonio Torralba, and Stefanie Jegelka.
\newblock Debiased contrastive learning.
\newblock {\em Advances in neural information processing systems}, 33:8765--8775, 2020.

\bibitem{courty2014domain}
Nicolas Courty, R{\'e}mi Flamary, and Devis Tuia.
\newblock Domain adaptation with regularized optimal transport.
\newblock In {\em Machine Learning and Knowledge Discovery in Databases: European Conference, ECML PKDD 2014, Nancy, France, September 15-19, 2014. Proceedings, Part I 14}, pages 274--289. Springer, 2014.

\bibitem{deng2009imagenet}
Jia Deng, Wei Dong, Richard Socher, Li-Jia Li, Kai Li, and Li~Fei-Fei.
\newblock Imagenet: A large-scale hierarchical image database.
\newblock In {\em 2009 IEEE Conference on Computer Vision and Pattern Recognition}, pages 248--255. Ieee, 2009.

\bibitem{dufumier2023integrating}
Benoit Dufumier, Carlo~Alberto Barbano, Robin Louiset, Edouard Duchesnay, and Pietro Gori.
\newblock Integrating prior knowledge in contrastive learning with kernel.
\newblock In {\em International Conference on Machine Learning}, pages 8851--8878. PMLR, 2023.

\bibitem{eisenberger2022unified}
Marvin Eisenberger, Aysim Toker, Laura Leal-Taix{\'e}, Florian Bernard, and Daniel Cremers.
\newblock A unified framework for implicit sinkhorn differentiation.
\newblock In {\em Proceedings of the IEEE/CVF Conference on Computer Vision and Pattern Recognition}, pages 509--518, 2022.

\bibitem{fang2023s}
Zhenghan Fang, Sam Buchanan, and Jeremias Sulam.
\newblock What's in a prior? learned proximal networks for inverse problems.
\newblock {\em arXiv preprint arXiv:2310.14344}, 2023.

\bibitem{ghosal2022convergence}
Promit Ghosal and Marcel Nutz.
\newblock On the convergence rate of sinkhorn's algorithm.
\newblock {\em arXiv preprint arXiv:2212.06000}, 2022.

\bibitem{ghosh2015making}
Aritra Ghosh, Naresh Manwani, and PS~Sastry.
\newblock Making risk minimization tolerant to label noise.
\newblock {\em Neurocomputing}, 160:93--107, 2015.

\bibitem{grathwohl2019your}
Will Grathwohl, Kuan-Chieh Wang, J{\"o}rn-Henrik Jacobsen, David Duvenaud, Mohammad Norouzi, and Kevin Swersky.
\newblock Your classifier is secretly an energy based model and you should treat it like one.
\newblock {\em arXiv preprint arXiv:1912.03263}, 2019.

\bibitem{grill2020bootstrap}
Jean-Bastien Grill, Florian Strub, Florent Altch{\'e}, Corentin Tallec, Pierre Richemond, Elena Buchatskaya, Carl Doersch, Bernardo Avila~Pires, Zhaohan Guo, Mohammad Gheshlaghi~Azar, et~al.
\newblock Bootstrap your own latent-a new approach to self-supervised learning.
\newblock {\em Advances in neural information processing systems}, 33:21271--21284, 2020.

\bibitem{gulrajani2020search}
Ishaan Gulrajani and David Lopez-Paz.
\newblock In search of lost domain generalization.
\newblock {\em arXiv preprint arXiv:2007.01434}, 2020.

\bibitem{he2020momentum}
Kaiming He, Haoqi Fan, Yuxin Wu, Saining Xie, and Ross Girshick.
\newblock Momentum contrast for unsupervised visual representation learning.
\newblock In {\em Proceedings of the IEEE/CVF conference on computer vision and pattern recognition}, pages 9729--9738, 2020.

\bibitem{hendrycks2019benchmarking}
Dan Hendrycks and Thomas Dietterich.
\newblock Benchmarking neural network robustness to common corruptions and perturbations.
\newblock {\em arXiv preprint arXiv:1903.12261}, 2019.

\bibitem{jaiswal2020survey}
Ashish Jaiswal, Ashwin~Ramesh Babu, Mohammad~Zaki Zadeh, Debapriya Banerjee, and Fillia Makedon.
\newblock A survey on contrastive self-supervised learning.
\newblock {\em Technologies}, 9(1):2, 2020.

\bibitem{jing2020self}
Longlong Jing and Yingli Tian.
\newblock Self-supervised visual feature learning with deep neural networks: A survey.
\newblock {\em IEEE transactions on pattern analysis and machine intelligence}, 43(11):4037--4058, 2020.

\bibitem{kim2021selfreg}
Daehee Kim, Youngjun Yoo, Seunghyun Park, Jinkyu Kim, and Jaekoo Lee.
\newblock Selfreg: Self-supervised contrastive regularization for domain generalization.
\newblock In {\em Proceedings of the IEEE/CVF International Conference on Computer Vision}, pages 9619--9628, 2021.

\bibitem{krizhevsky2009learning}
Alex Krizhevsky and Geoffrey Hinton.
\newblock Learning multiple layers of features from tiny images.
\newblock Technical report, University of Toronto, 2009.

\bibitem{lee2019hierarchical}
John Lee, Max Dabagia, Eva Dyer, and Christopher Rozell.
\newblock Hierarchical optimal transport for multimodal distribution alignment.
\newblock {\em Advances in neural information processing systems}, 32, 2019.

\bibitem{li2017deeper}
Da~Li, Yongxin Yang, Yi-Zhe Song, and Timothy~M Hospedales.
\newblock Deeper, broader and artier domain generalization.
\newblock In {\em Proceedings of the IEEE international conference on computer vision}, pages 5542--5550, 2017.

\bibitem{li2019learning}
Ruilin Li, Xiaojing Ye, Haomin Zhou, and Hongyuan Zha.
\newblock Learning to match via inverse optimal transport.
\newblock {\em Journal of machine learning research}, 20(80):1--37, 2019.

\bibitem{lin2021making}
Chi-Heng Lin, Mehdi Azabou, and Eva~L Dyer.
\newblock Making transport more robust and interpretable by moving data through a small number of anchor points.
\newblock {\em Proceedings of machine learning research}, 139:6631, 2021.

\bibitem{liu2021drop}
Ran Liu, Mehdi Azabou, Max Dabagia, Chi-Heng Lin, Mohammad Gheshlaghi~Azar, Keith Hengen, Michal Valko, and Eva Dyer.
\newblock Drop, swap, and generate: A self-supervised approach for generating neural activity.
\newblock {\em Advances in neural information processing systems}, 34:10587--10599, 2021.

\bibitem{montesuma2023recent}
Eduardo~Fernandes Montesuma, Fred~Ngole Mboula, and Antoine Souloumiac.
\newblock Recent advances in optimal transport for machine learning.
\newblock {\em arXiv preprint arXiv:2306.16156}, 2023.

\bibitem{netzer2011reading}
Yuval Netzer, Tao Wang, Adam Coates, Alessandro Bissacco, Bo~Wu, and Andrew~Y Ng.
\newblock Reading digits in natural images with unsupervised feature learning.
\newblock In {\em NIPS Workshop on Deep Learning and Unsupervised Feature Learning}, volume 2011, 2011.

\bibitem{nguyen2010estimating}
XuanLong Nguyen, Martin~J Wainwright, and Michael~I Jordan.
\newblock Estimating divergence functionals and the likelihood ratio by convex risk minimization.
\newblock {\em IEEE Transactions on Information Theory}, 56(11):5847--5861, 2010.

\bibitem{oord2018representation}
Aaron van~den Oord, Yazhe Li, and Oriol Vinyals.
\newblock Representation learning with contrastive predictive coding.
\newblock {\em arXiv preprint arXiv:1807.03748}, 2018.

\bibitem{ozair2019wasserstein}
Sherjil Ozair, Corey Lynch, Yoshua Bengio, Aaron Van~den Oord, Sergey Levine, and Pierre Sermanet.
\newblock Wasserstein dependency measure for representation learning.
\newblock {\em Advances in Neural Information Processing Systems}, 32, 2019.

\bibitem{parikh2014proximal}
Neal Parikh, Stephen Boyd, et~al.
\newblock Proximal algorithms.
\newblock {\em Foundations and trends{\textregistered} in Optimization}, 1(3):127--239, 2014.

\bibitem{pass2015multi}
Brendan Pass.
\newblock Multi-marginal optimal transport: theory and applications.
\newblock {\em ESAIM: Mathematical Modelling and Numerical Analysis}, 49(6):1771--1790, 2015.

\bibitem{peyre2015entropic}
Gabriel Peyr{\'e}.
\newblock Entropic approximation of wasserstein gradient flows.
\newblock {\em SIAM Journal on Imaging Sciences}, 8(4):2323--2351, 2015.

\bibitem{peyre2019computational}
Gabriel Peyr{\'e}, Marco Cuturi, et~al.
\newblock Computational optimal transport: With applications to data science.
\newblock {\em Foundations and Trends{\textregistered} in Machine Learning}, 11(5-6):355--607, 2019.

\bibitem{pham2020unbalanced}
Khiem Pham, Khang Le, Nhat Ho, Tung Pham, and Hung Bui.
\newblock On unbalanced optimal transport: An analysis of sinkhorn algorithm.
\newblock In {\em International Conference on Machine Learning}, pages 7673--7682. PMLR, 2020.

\bibitem{pu2022alignment}
Shi Pu, Kaili Zhao, and Mao Zheng.
\newblock Alignment-uniformity aware representation learning for zero-shot video classification.
\newblock In {\em Proceedings of the IEEE/CVF Conference on Computer Vision and Pattern Recognition}, pages 19968--19977, 2022.

\bibitem{radford2021learning}
Alec Radford, Jong~Wook Kim, Chris Hallacy, Aditya Ramesh, Gabriel Goh, Sandhini Agarwal, Girish Sastry, Amanda Askell, Pamela Mishkin, Jack Clark, et~al.
\newblock Learning transferable visual models from natural language supervision.
\newblock In {\em International conference on machine learning}, pages 8748--8763. PMLR, 2021.

\bibitem{robinson2020contrastive}
Joshua Robinson, Ching-Yao Chuang, Suvrit Sra, and Stefanie Jegelka.
\newblock Contrastive learning with hard negative samples.
\newblock {\em arXiv preprint arXiv:2010.04592}, 2020.

\bibitem{robinson2021can}
Joshua Robinson, Li~Sun, Ke~Yu, Kayhan Batmanghelich, Stefanie Jegelka, and Suvrit Sra.
\newblock Can contrastive learning avoid shortcut solutions?
\newblock {\em Advances in neural information processing systems}, 34:4974--4986, 2021.

\bibitem{rout2021generative}
Litu Rout, Alexander Korotin, and Evgeny Burnaev.
\newblock Generative modeling with optimal transport maps.
\newblock {\em arXiv preprint arXiv:2110.02999}, 2021.

\bibitem{saunshi2022understanding}
Nikunj Saunshi, Jordan Ash, Surbhi Goel, Dipendra Misra, Cyril Zhang, Sanjeev Arora, Sham Kakade, and Akshay Krishnamurthy.
\newblock Understanding contrastive learning requires incorporating inductive biases.
\newblock In {\em International Conference on Machine Learning}, pages 19250--19286. PMLR, 2022.

\bibitem{shi2023understanding}
Liangliang Shi, Gu~Zhang, Haoyu Zhen, Jintao Fan, and Junchi Yan.
\newblock Understanding and generalizing contrastive learning from the inverse optimal transport perspective.
\newblock In {\em International Conference on Machine Learning}, pages 31408--31421. PMLR, 2023.

\bibitem{shurrab2022self}
Saeed Shurrab and Rehab Duwairi.
\newblock Self-supervised learning methods and applications in medical imaging analysis: A survey.
\newblock {\em PeerJ Computer Science}, 8:e1045, 2022.

\bibitem{stuart2020inverse}
Andrew~M Stuart and Marie-Therese Wolfram.
\newblock Inverse optimal transport.
\newblock {\em SIAM Journal on Applied Mathematics}, 80(1):599--619, 2020.

\bibitem{taherkhani2021self}
Fariborz Taherkhani, Ali Dabouei, Sobhan Soleymani, Jeremy Dawson, and Nasser~M Nasrabadi.
\newblock Self-supervised wasserstein pseudo-labeling for semi-supervised image classification.
\newblock In {\em Proceedings of the IEEE/CVF conference on computer vision and pattern recognition}, pages 12267--12277, 2021.

\bibitem{tolstikhin2017wasserstein}
Ilya Tolstikhin, Olivier Bousquet, Sylvain Gelly, and Bernhard Schoelkopf.
\newblock Wasserstein auto-encoders.
\newblock {\em arXiv preprint arXiv:1711.01558}, 2017.

\bibitem{vishnubhotla2024towards}
Ankit Vishnubhotla, Charlotte Loh, Akash Srivastava, Liam Paninski, and Cole Hurwitz.
\newblock Towards robust and generalizable representations of extracellular data using contrastive learning.
\newblock {\em Advances in Neural Information Processing Systems}, 36, 2024.

\bibitem{wang2020understanding}
Tongzhou Wang and Phillip Isola.
\newblock Understanding contrastive representation learning through alignment and uniformity on the hypersphere.
\newblock In {\em International Conference on Machine Learning}, pages 9929--9939. PMLR, 2020.

\bibitem{wang2024exploring}
Yule Wang, Chengrui Li, Weihan Li, and Anqi Wu.
\newblock Exploring behavior-relevant and disentangled neural dynamics with generative diffusion models.
\newblock {\em arXiv preprint arXiv:2410.09614}, 2024.

\bibitem{wang2024extraction}
Yule Wang, Zijing Wu, Chengrui Li, and Anqi Wu.
\newblock Extraction and recovery of spatio-temporal structure in latent dynamics alignment with diffusion model.
\newblock {\em Advances in Neural Information Processing Systems}, 36, 2024.

\end{thebibliography}

\newpage
\appendix
\onecolumn

\section*{Appendix}\label{app:outline}

\setcounter{figure}{0}
\renewcommand{\thefigure}{A\arabic{figure}}

\setcounter{table}{0}
\renewcommand{\thetable}{A\arabic{table}}

\setcounter{appendixalgorithm}{1}
\renewcommand{\thealgorithm}{A\arabic{appendixalgorithm}}

\addcontentsline{toc}{section}{Appendix} 

\section{Background and Notation}

\subsection{Notation}
\label{app:notations}
\paragraph{Datasets and contrastive pairs:} Let $\mathbf{x}$ denotes a vector and $\mathbf{X}$ denotes a matrix, with right subscript \({\bf X}_b\) denote the batch of the input samples, ${\bf X}_b:=[\x_1,\x_2,\dots,\x_B]$, here \(B\) is equal to batch size. For each sample $\x_i$ in the input batch matrix $\mathbf{X}_b$, $\x'_i$ means augmented view 1 of $\x'_i$, $\x''_i$ means augmented view 2 of $\x_i$, the positive pairs in input data denoted as ($\x_i$, $\x''_i$), negative pairs in input data denoted as ($\x'_i$, $\x''_j$), $i\neq j$. Give a weights (\(\theta\)) parametrized representation function (artificial neural network) $f_\theta$ with adjustable adjustable temperature $\varepsilon$, which project the the positive pairs in latent space denoted as  \(s^+=\langle \varepsilon^{-1} \widetilde{f}_\theta(\x'_i),\widetilde{f}_\theta(\x''_i) \rangle\), and negative pairs in latent space denoted as \(s^-=\langle \varepsilon^{-1} \widetilde{f}_\theta(\x'_i),\widetilde{f}_\theta(\x''_j) \rangle, i\neq j\). Here, \(\langle \cdot,\cdot \rangle\) is the inner product, which means $\langle \widetilde{f}_\theta(\x'_i),\widetilde{f}_\theta(\x''_i) \rangle = f_{\theta}({\x_i'})^\top f_{\theta}(\x_j'')/\|f_{\theta}({\x_i'})\|\|f_{\theta}({\x_j''})\|$ is the normalized form.
     
\paragraph{Continuous settings for optimal transport}
     \(\mathcal{X}\) and \(\mathcal{Y}\) are topological spaces, \(\mathcal{X} \times \mathcal{Y}\) is the product space, or Torus. \(C(\mathcal{X})\) is the compact topological space which contains all of continuous functions on \(\mathcal{X}\) endowed with the sup-norm. On Torus \(\mathcal{X}\) and \(\mathcal{Y}\) we define $M$ as a compact n-dimensional manifold in product space \(\mathcal{X}\times\mathcal{Y}\) ($\mathcal{X}=\mathcal{Y}:= \mathbb{R}^n / \mathbb{Z}^n$) endowed with a cost function \(c(x, y) := d_{M}(x, y)^2/2\) (Euclidean dsitance function) on $\mathbb{R}^n$. Transport plan $\pi(x,y):\mathcal{X} \times \mathcal{Y}\rightarrow \mathbb{R}$ is an element in  \(\mathcal{P}(\mathcal{X}\times\mathcal{Y})\). \(\mathcal{P}(\mathcal{X}\times\mathcal{Y})\) means the collections of the joint distributions of the two marginal distributions $\mu\in\mathcal{P}(\mathcal{X})$ and $\nu\in\mathcal{P}(\mathcal{Y})$.  $\mathcal{P}(\mathcal{X})$ the space of all (Borel) probability measures on \(\mathcal{X}\), $\mathcal{P}(\mathcal{Y})$ means the same to $\mathcal{Y}$.  To find a joint distribution (or plan) \( \pi(x,y) \) in collections \(U(\mu,\nu)\) with marginals \( \mu \) and \( \nu \) in the product space \(\mathcal{X} \times \mathcal{Y}\), we can formulate as:
\begin{equation*}
 \label{ot}
\min_{\pi\in U(\mu, \nu)}\sum_{x\in\mathcal{X},y\in\mathcal{Y}}\pi(x, y)c(x,y) \quad \text{s.t.} ~\sum_{y\in\mathcal{Y}}\pi(x,y) =\mu(x),~\sum_{x\in\mathcal{X}}\pi(x,y) =\nu(y)
\end{equation*} 

\paragraph{Discrete settings of optimal transport}: \( \mu \) and \( \nu \) can be discrete probability measures whose supports are finite sets \( X := \{ f_\theta(\x'_i) \}_{i=1}^{N}, Y := \{ f_\theta(\x''_i)\}_{i=1}^{B} \). And we define \( \mu = \sum_{i=1}^{B} \delta_{f_\theta(\x'_i)} p_i, \nu = \sum_{i=1}^{B} \delta_{f_\theta(\x''_i)} q_i, \) with vectors \( p \) and \( q \) in a simplex \( \Delta_{B} \) in \( \mathbb{R}^{B} \) defined by \( \Delta_B := \left\{ v \in \mathbb{R}^B : v_i \geq 0, \sum_{i=1}^B v_i = 1 \right\}\), which we identify with \( \mathcal{P}(\{1, \ldots, B\}) \).
      \(\C\) is a \(B\times B\) cost matrix calculated by \(c(x,y)\), whose sampled from finite sets \(X\) and \(Y\) defined previously, and N is the batch size. \(\uu\) and \(\vv\) are \(B\times 1\) scale factors matrix, \(\uu^{(t)}\) and \(\vv^{(t)}\) mean the scale factor matrix after t iterations of sinkhorn algorithms. \(\PP\) is a \(B\times B\) joint distribution matrix of \(\mu\) and \(\nu\), which represents the transport plan \(\pi(x,y)\) that corresponds to minimize the cost. \(\PP^{(2t-1)}\) means we use t iterations \(\uu^{(t)}\) and \(\vv^{(t-1)}\) to calculate \(\PP^{(t)}\). When \(t=1\) means we use \(\uu^{(1)}\) and \(\vv^{(0)}\) to calculate \(\PP^{(1)}\), which is called half-step OT or one step Bregman projection. When \(t=2\) means we use \(\uu^{(1)}\) and \(\vv^{(1)}\) to calculate \(\PP^{(2)}\), which is called half-step OT or one step Bregman projection.

\subsection{Proximal operator setup}
\label{app:prox}
In this section, we are going to provide the detailed illustration about the proximal operators. How the proximal operator would convert to the projection. And how to solve the Bregman projection with KL divergence.

\subsubsection{Explanation of Definition 1 }

Let \( h: \mathcal{X} \rightarrow [-\infty, +\infty]\) be a proper, lower semi-continuous convex function on Hilbert space $\mathcal{X}$. The \textbf{proximal operator} of \(h\) at point \(\vv \in \mathcal{X}\) is defined as a unique minimizer of the function \(\x\mapsto h(\x)+\frac{1}{2}\|\x-\vv \|^2_2\)~\cite{parikh2014proximal}. For an instance, \( h \) is convex function relative to the constraint set \( \mathcal{B} \), like the indicator function \(h_\mathcal{B}\). Given the Euclidean norm on $\mathcal{X}$, we can write the proximal operator \(\text{Prox}_{h,\mathcal{B}}^{\|\cdot\|^2}(\mathbf{v}) \) as the most common way:
\[ \text{Prox}_{h,\mathcal{B}}^{\|\cdot\|^2}(\mathbf{v}) = \arg\min_{\mathbf{x} \in \mathcal{B}} \left\{ h(\mathbf{x}) + \frac{1}{2} \|\mathbf{x} - \mathbf{v}\|^2_2 \right\} \]

The solution to the proximal operator exists and is unique due to the strong convexity of the above function.

\subsubsection{Connection to the projection}
Here, if we treat \(h(x)\) as the indicator functions of the constraint set \(\mathcal{B}\) as:
\[
h(x) = 
\begin{cases} 
0, & \text{if } x \in \mathcal{B} \\
\infty, & \text{if } x \notin \mathcal{B}.
\end{cases}
\]
Then the proximal operators problem could be understood intuitively as finding the "shortest distance" between the point \(v\) with the constraint set \(\mathcal{B}\), which means the projection. And the following lemma holds:
\begin{lemma} \label{lm:unique}
    If the constraint set \(\mathcal{B}\) is closed and convex, then the projection of point \(v\) is unique on \(\mathcal{B}\).
\end{lemma}

\textbf{Proof of the Lemma~\ref{lm:unique}}: This lemma can be proved by the strict convexity of the proximal operator.

\begin{figure}[ht]
\centering
    \includegraphics[width=0.8\linewidth]{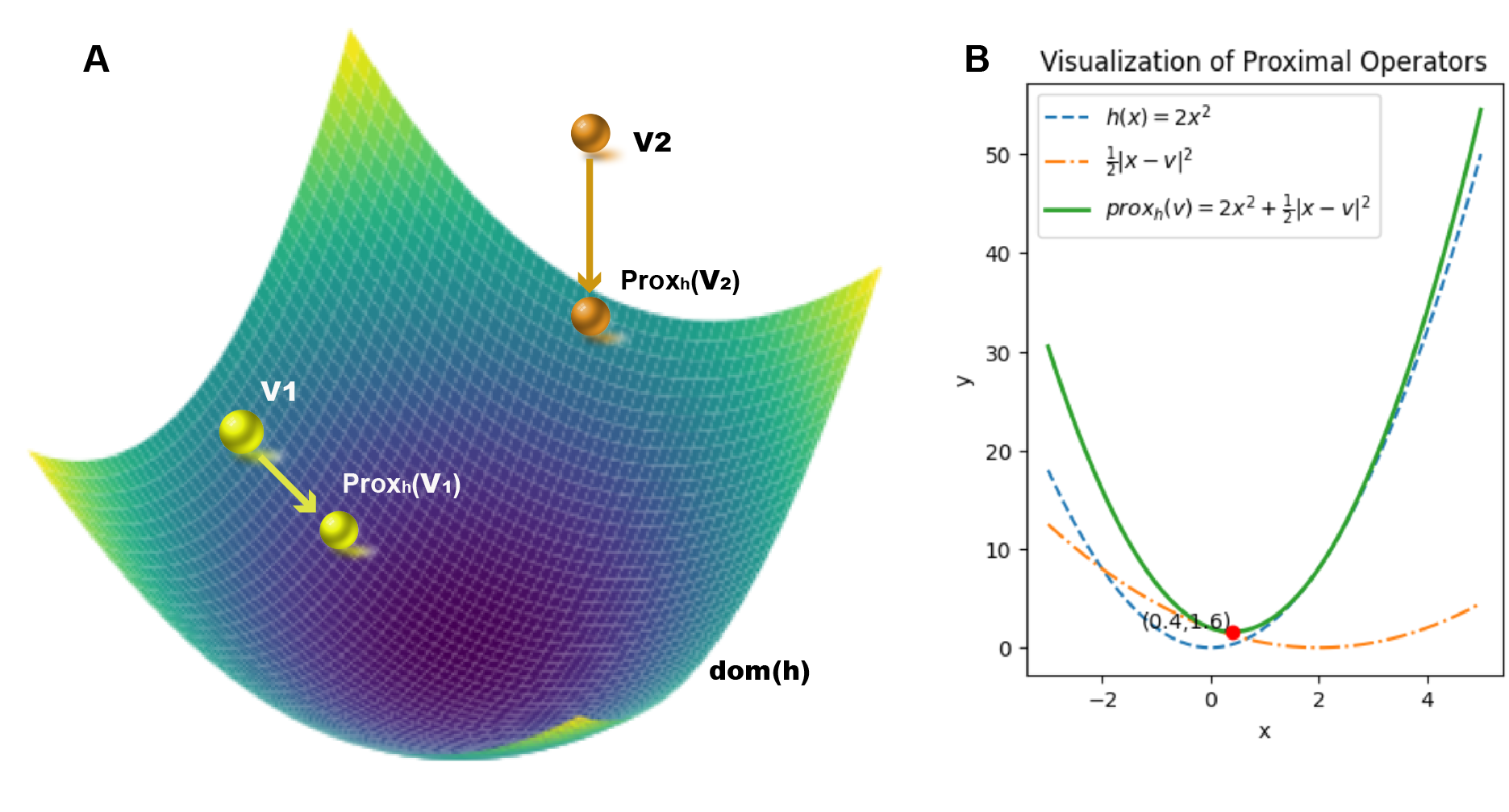}
\caption{ \footnotesize{{\em Illustration of the proximal operators}  A. Visualization of proximal operators in $\mathbb{R}^3$. On the surface defined by $h(x, y) = x^2 + y^2$ within the domain constraints $-1.2 < x < 1.2$ and $-1.2 < y < 1.2$. If $v=v_1=(0.76, 0.76, 1.16)$, it lies within the domain of $h$, represented on the surface at the exact location matching its third coordinate with $h(x, y)$. If $v=v_2=(1.5, 1.5, 6)$, which is outside the feasible region defined by $h$, the proximal operator projects it to the closest point within the domain, resulting in $v_2$'s projection to approximately $(0.85, 0.85, 1.45)$.  B. Visualization of proximal operators in $\mathbb{R}^2$.} The blue dashed line represents the function $h(x) = x^2$. The orange dash-dotted line illustrates the penalty term $\frac{1}{2} \|x - v\|^2$ with $v = (2, 0)$, indicating the squared distance from any $x$ to $v$. The green solid line is the proximal operator $2x^2 + \frac{1}{2} \|x - v\|^2$, which gets close to the minimization point of h(x) from $v$. The red point marks the $\text{Prox}_h(v)$ in this space.}
    \label{fig:proximal}
\end{figure}

\subsubsection{Connection to the Bregman divergence}
First, we define $d_\Gamma$ as a generic Bregman divergence on some convex set $\mathcal{B}$, and the proximal map of a convex function $d_\phi$ according to this divergence is:
\begin{equation}
\text{Prox}^{d_\Gamma}_{d_\phi,\mathcal{B}}(\K) := \arg\min_{\PP \in \mathcal{B}} d_\Gamma( \PP \|\K) + d_\phi( \PP).
\end{equation}
$\Gamma$ is a strictly convex function smooth on $\text{int}(\mathcal{B})$, and $\text{Prox}^{d_\Gamma} _{d_\phi}(\K) \in \text{int}(\mathcal{B})$ is always uniquely defined by strict convexity. (Note that this theory is general and does not need to parametrize the \(\K\) and \(\PP\) as models with \(\theta\)). As $\mathcal{B}= \text{dom}(\Gamma)$, $$\forall (\PP,\K) \in \mathcal{B} \times \text{int}(\mathcal{B}), d_\Gamma(\PP\|\K) = \Gamma(\PP) - \Gamma(\K) - \langle \nabla \Gamma(\K), \PP - \K \rangle,$$
which has its Legendre transform is also smooth and strictly convex: $$\Gamma^*(\rho) = \max_{\PP \in \mathcal{B}} \langle \PP, \rho \rangle - \Gamma(\PP)$$
 The Bregman divergence for a convex function \( \Gamma \) between points \( \x \) and \( \y \) is defined as:
\[
d_\Gamma(\x, \y) = \Gamma(\x) - \Gamma(\y) - \langle \nabla \Gamma(\y), \x - \y \rangle
\]
where  \( \nabla \Gamma(\y) \) is the gradient of \( \Gamma \) at \( \y \). Giving the squared L2 distance can be viewed as a Bregman divergence derived from the convex function \( \Gamma(\x) = \|\x\|^2 \). For this function, the Bregman divergence between two points \( \x \) and \( \y \) becomes:
\[
d_\Gamma(\x, \y) = \|\x\|^2 - \|\y\|^2 - 2\y^\top(\x - \y)=\|\x - \y\|^2
\]

\begin{table}[h]
\centering
\caption{\footnotesize {\em Examples of functions $\Gamma$ and their corresponding divergences $d_\Gamma$.}}
\begin{tabular}{c|c|c}
\hline
\textbf{$\Gamma$} & \textbf{$d_\Gamma$} & \textbf{Description} \\ \hline
$\|\x\|^2$ & $\|\x - \y\|^2$ & squared Euclidean distance \\
$\x \ln \x$ & $\y \ln \frac{\y}{\x} - (\y - \x)$ & Kullback–Leibler (KL) divergence \\
$-H(p) = \sum_j p_j \ln p_j$ & $KL(q\|p) = \sum_j q_j \ln \frac{q_j}{p_j}$ & KL divergence between distributions $p, q$ \\
 & $\sum p_j = \sum q_j = 1$ &  \\ \hline
\end{tabular}
\end{table}

\subsubsection{Connection to the Bregman projection}
Bregman projections solve the alignment problem onto the two constraints sets that encode the marginals along the rows and columns ~\cite{benamou2015iterative, bregman1967relaxation, peyre2019computational}.
 \begin{equation}\label{eq:birkhoffapp}
    C_1^\mu \coloneqq \{\PP : \PP\mathbbm{1}_B = \mu\}, C_2^\nu \coloneqq \{\PP : \PP^\top \mathbbm{1}_B = \nu\}
\end{equation}

If we specify the constraint set \(\mathcal{B}\) as some set \(C_1^\mu \coloneqq \{\PP : \PP\mathbbm{1}_m = \mu\}\), and select the \(h_\mathcal{F}(x)\) as some indicator function of \(C_1^\mu\), which satisfies:
\begin{equation}~\label{eq:indf}
h_\mathcal{F}(\x ) = 
\begin{cases} 
0, & \text{if } \x\in \C_1^\mu \\
\infty, & \text{if } \x \notin \C_1^\mu.
\end{cases}
\end{equation}

For the first step Bregman projection \(\text{Prox}_{C_1^\mu}^{\text KL}(\K)\) onto the set $\mathcal{C}_1^{\mu}$ with indicator function in Equation~\eqref{eq:indf}:
\begin{equation}~\label{eq:breg1}
\PP^{(1)}=\text{Prox}_{C_1^\mu}^{\text{KL}}(\K_\theta)  = \arg\min_{\PP\in C_1^\mu} \{ h_\mathcal{F}(\PP)+\text{KL}(\PP\|\mathbf{\K})\} = \arg\min \{\text{KL}(\PP\|\K): \PP\mathbbm{1}_B=\mu \}.
\end{equation}
We can minimize the function in Equation~\eqref{eq:breg1} with Lagrange multiplier \(f\) on \(C_1^\mu\):
\begin{equation}\label{eq:oriprob}
\varepsilon\text{KL}(\PP\|\K)- f(\PP\mathbbm{1}_B -\mu),
\end{equation}
Then we get \(\PP^{(1)}\) as the minimizer through the derivatives with respect to \(\PP\), we have 
\begin{equation}
\varepsilon\log\left(\PP^{(1)}/ \K\right) - f\mathbbm{1} = 0 \Rightarrow \PP^{(1)} = \uu\K,~\text{as}~\uu = e^{f/\varepsilon} > 0 , 
\end{equation}
and we can use these relationship into the constraints sets with $\PP^{(0)} = \diag(\mathbbm{1}) {\bf K} \diag(\mathbbm{1})$.
\begin{equation}\label{eq:scaleapp}
 \langle \PP^{(1)}, \mathbbm{1}\rangle = \mu \Rightarrow \langle \uu^{(1)}\K, \mathbbm{1}\rangle = \mu, \uu^{(1)} = \frac{\mu}{\sum_i \K_{ij}},  \PP^{(1)} =  \text{diag}\left(\frac{\mu}{\PP^{(0)}\mathbbm{1}_B}\right)\PP^{(0)}, 
\end{equation}

If we repeat this progress for the set \(C_2^\nu \coloneqq \{\PP : \PP^T\mathbbm{1}_n = \nu\}\), we will get the \(\text{Prox}_{C_2^\nu}^{\text{KL}}(\PP^{(t+1)})\).
 And in the second step, we project onto the second constraint set \(C_2^\nu\) with indicator function \(h_\mathcal{G}(x)\) defined on \(C_2^\nu\) and get:
\begin{equation}\label{eq:Bregmaniterations}
\PP^{(2)} \coloneqq \text{Prox}^\text{KL}_{C_2^\nu}(\PP^{(1)}) = \PP^{(1)} \text{diag}\left(\frac{\nu}{\PP^{(1)\top}\mathbbm{1}_B}\right).
\end{equation}
Iterating over these two sets of projections \(\PP^{(t+1)} \coloneqq \text{Prox}^\text{KL}_{C_1^\mu}(\PP^{(t)})\) and \(\PP^{(t+2)} \coloneqq \text{Prox}^\text{KL}_{C_2^\nu}(\PP^{(t+1)})\) until convergence could be summarized as Sinkhorn algorithm with \(t\) via recursive form:
\begin{equation}\label{eq:uvupdatesapp}  
\uu^{(t+1)} \stackrel{\text { def }}{=} \frac{{\mu}}{\mathbf{K} \vv^{(t)}}, \quad \vv^{(t+1)} \stackrel{\text { def }}{=} \frac{{\nu}}{\mathbf{K}^{\mathrm{T}} \uu^{(t+1)}}, \quad \PP^{(2t+2)}=\operatorname{diag}(\uu^{(t+1)})\mathbf{K}\operatorname{diag}(\vv^{(t+1)}). 
\end{equation}

The Sinkhorn algorithm is composed with two steps Bregman projection, 
Similarly, we can write out this recursive relationship as: \(\PP^{t+1}\) can be updated with dual variables \(f\), \(g\) and \(\uu^{(t)}=e^{f^{(t)}/\varepsilon}, \vv^{(t)}=e^{g^{(t)}/\varepsilon}\). The set $ U(\mu, \nu) = C_1^\mu \cap C_2^\nu$, representing the feasible transport plans with given marginals. It could be any random sets, i. e. $\mathcal{B} = C_1^\mathbbm{1} \cap C_2^\mathbbm{1}$ denote the Birkhoff polytope of doubly stochastic matrices where $\mu = \mathbbm{1}$ and $\nu = \mathbbm{1}$ are the uniform distributions with all one element.

\subsection{Background on OT}
\label{app:ot}
This section defines discrete and continuous optimal transport. Since the section~\ref{sec:otppm} lacks a discrete OT definition, we discuss it here and show the equivalence between solving Bregman projection and the entropy-regularized OT (EOT) problem.

To support convergence proofs later, we introduce definitions of continuous measures. Symbols \( \mu \) and \( \nu \) may represent both discrete and continuous measures for intuitive consistency, with precise definitions at the start of each subsection.

\subsubsection{Background on discrete OT}
\label{app:descrete_ot}
In section~\ref{sec:otppm}, we provide a general definition of the discrete optimal transport. Here, we specifically define the optimal transport on the representations space after an encoder \(f_\theta\) with two augmented views \((\x',\x'')\).  As we mainly discussed the distribution on the representation space, so here we suppose there is an encoder \(f_\theta\) will project the augmented views into the latent. 
Here we define \( \mu = \sum_{i=1}^{N} \delta_{f_\theta(\x'_i)} p_i, \nu = \sum_{i=1}^{N} \delta_{f_\theta(\x''_i)} q_i, \) with vectors \( p \) and \( q \) in a simplex \( \Delta_{B} \) in \( \mathbb{R}^{B} \) defined by \( \Delta_B := \left\{ v \in \mathbb{R}^B : v_i \geq 0, \sum_{i=1}^B v_i = 1 \right\}\). Here, \(\PP\) is a \(B\times B\) joint coupling matrix of the marginal distributions \(\mu\) and \(\nu\), which describes how much mass is needed to convert one distribution to match another. \(\C\) is a \(B\times B\) cost matrix calculated by the cost function \(c(x,y)\) i.e. cosine dissimilarity, and we can write the OT problem as the constrained linear programming problem: 
\begin{align}\label{eq:dot}
   \min_{\bf{P}}\langle\bf{P},\C \rangle ~~\text{s.t. } \bf{P}\mathbbm{1} = \mu,~\bf{P}^\top\mathbbm{1} = \nu.
\end{align}

Even though directly solving Equation~\eqref{eq:dot} is high computational complexity $O(n^3)$, we introduce a common relaxation called entropic regularization to smooth the transport plan.

\subsubsection{Entropy regularized OT and the Sinkhorn algorithm.} Solving the exact OT problem above can be very computationally intensive.  In this case, we can add the Shannon entropy $H(\PP)= -(\PP_{ij} (\log(\PP_{ij}))$ to our objective in Equation~\eqref{eq:dot} and obtain an approximation of entropy-regularized optimal transport (EOT) plan as: 
\begin{equation}\label{eq:eotapp}
\min_{\PP \in \mathcal{B}} ~ \langle\PP,\C\rangle -\varepsilon H(\PP), \quad \text{where}~H(\PP)= -\sum \PP_{ij} \log(\PP_{ij}),
\end{equation}
where $\varepsilon$ is a user specified parameter that controls the amount of smoothing in the transport plan. The cost matrix \(\C\) could be transformed into the Gibbs kernel matrix \(\K\) on a Hilbert space with the given formula, 
\begin{equation}\label{eq:gibbs}  
\K_{ij} =\exp\left(-\varepsilon^{-1}\C_{ij}\right) 
\end{equation}

To solve (\ref{eq:eotapp}) under the kernel space induced by \(\K\), we can use the  iterative Sinkhorn algorithm with the initialization of $\uu^{(0)}$ and $\vv^{(0)}$ as all one vector divided by the batch size, and the update rules:
\begin{align}
\uu^{(t+1)} \stackrel{\text { def }}{=} \frac{{\mu}}{\mathbf{K} \vv^{(t)}} \text { and } \vv^{(t+1)} \stackrel{\text { def }}{=} \frac{{\nu}}{\mathbf{K}^{\mathrm{T}} \uu^{(t+1)}},
\end{align}
Then, the output of plan after $t$ iterations is 
\begin{equation}\label{Pt}
    \PP^{(t)}=\operatorname{diag}(\uu^{(t)})\mathbf{K}\operatorname{diag}(\vv^{(t)}).
\end{equation}
It also could be interpreted with dual variables \(f\) and \(g\):
\begin{equation}\label{eq:dual}
\PP^{(t)}_{i,j} = e^{f^{(t)}_i/\varepsilon} e^{-\C_{i,j}/\varepsilon} e^{g^{(t)}_j/\varepsilon}, \quad \uu^{(t)}=e^{f^{(t)}/\varepsilon}, \vv^{(t)}=e^{g^{(t)}/\varepsilon}
\end{equation}
After convergence, the resulting $\PP$ will be the optimal solution to Equation~\eqref{eq:eotapp}. The convergence and dynamics of OT and the dual formulation have been studied extensively in~\cite{berman2020sinkhorn, peyre2019computational, ghosal2022convergence, an2022efficient}. Here, iterations converge to a stable transport plan \(\PP^{(\infty)}\)as the optimal solution of Equation~\eqref{eq:eot}, which provides the minimum cost matching between two distributions. 
The convergence and dynamics of OT and its dual formulation have been studied extensively in~\cite{berman2020sinkhorn, peyre2019computational, ghosal2022convergence, an2022efficient}. Thus, these results guarantee that the iterates will converge to the optimal solution of the EOT objective, or that \(\PP^{(t)} \rightarrow \PP^{(\infty)}\) with \(t \rightarrow \infty\).  

This allows us to state the following lemma: 
\begin{lemma}\label{lm:eoteqbreg}
Solving the entropy optimal transport in Equation~\eqref{eq:eot} is consistent with iterative solving the Bregman projection.
\end{lemma}
\textbf{Proof of the Lemma~\ref{lm:eoteqbreg}}: 
Giving that some points \(\K\) and \(\PP\), their distance could be measured by KL divergence:
\[\text{KL}(\PP\|\K) = \sum_{ij} \PP_{ij} \log\left(\frac{\PP_{ij}}{\K_{ij}}\right) - \PP_{ij} + \K_{ij}\]
As \(\C_{ij} =-\varepsilon \log\K_{ij}\) in Sinkhorn, we can see find \(\PP\)  to minimize the Equation~\eqref{eq:eot} can be transformed into some formula about \(\K_{ij}\):
\begin{align}
 \min_\PP\langle\PP,\C\rangle -\varepsilon H(\PP) & = \min_\PP \sum_{i,j}  \C_{ij} \PP_{ij} +\varepsilon \sum_{i,j} \PP_{ij} (\log(\PP_{ij}) ) \\
&= \min_\PP \varepsilon \sum_{i,j} (- \PP_{ij}\log\K_{ij}+\PP_{ij} \log(\PP_{ij}))\\
&= \min_\PP \varepsilon \text{KL}(\PP\|\K) ~~\text{s.t. } \PP\mathbbm{1}= \mu,~\PP^\top\mathbbm{1} = \nu,
\end{align}
Consider the \(\K\) is a point in Hilbert kernel space, and \(\varepsilon \) is the constant, we set the \(\mu\) and \(\nu\) form the \(\mathcal{B}\), so here can have:
\begin{equation}~\label{eq:eoteqbreg}
\PP = \text{Prox}^{\text{KL}}_{\mathcal{B}}(\K)  = \arg\min_{\PP \in \mathcal{B}} \text{KL}(\PP||\K) =\arg\min_\PP \{\langle\PP,\C\rangle +\varepsilon H(\PP): \PP\mathbbm{1}= \mu,~\PP^\top\mathbbm{1} = \nu\}
\end{equation}

\subsubsection{Background on continuous optimal transport}
\label{app:continous}
To show the convergence of the Bregman projection, here we define the optimal transport problem with the continuous measure. 
Inherit the definition of \(\mathcal{X}\) and \(\mathcal{Y}\) in the Appendix~\ref{app:notations}, finding the optimal transport between two continuous measure \(\mu\) and \(\nu\) could be transformed into some problems with the minimization of Kantorovich functional. 

\begin{definition}[\textit{Continuous optimal transport}] 
We redefine $\mu$ and $\nu$ be two probability measures on latent manifold $\mathcal{M}$ with Hölder continuous and strictly positive densities $e^{f}$ and $e^{g}$, respectively: \(\mu = e^{f} dM, \nu = e^{g}dM \), where $dM$ is the Riemannian normalized volume form on $\mathcal{X}$. For each \(x \in \mathcal{X}\) and \(y \in \mathcal{Y}\):
\begin{equation}
W(\mu, \nu) = \inf_{\pi \in \Pi(\mu, \nu)} \int_{\mathcal{X} \times \mathcal{Y}} c(x, y) \, d\pi(x, y).
\label{eq:wdist}
\end{equation}

\end{definition}

\begin{definition}[\textit{Dual of continuous OT}] The dual of standard OT reads:
\begin{equation}
W(\mu, \nu) = \sup_{f,g \in \mathcal{U}(c)} \int_{\mathcal{X}} fd\mu(x)+\int_{\mathcal{Y}} d\nu(y)
\label{eq:continousotdual}
\end{equation}
where the constraint set \(\mathcal{U}(c)\)  is defined by \(\mathcal{U}(c):=\{(\mu,\nu)\in\mathcal{C}(\mathcal{X})\times\mathcal{C}(\mathcal{Y})\}| f(x)+g(y) \leq c(x,y)\} \).
\end{definition}

Here, \(C(\mathcal{X})\) is the space of all continuous functions on \(\mathcal{X}\), the functions which measured using the supreme norm \(||f||_\infty\), with the Legendre transform:

 \begin{definition}[\textit{Legendre c-transforms}]
For the dual variables, or so called potentials, there exists the Legendre c-transforms:
\begin{equation}
\label{eq:legendre}
f^c(y) := \sup_{x \in \mathcal{X}} (-c(x, y) + f(x)), \quad g^c(x) := \sup_{y \in \mathcal{Y}} (-c(x, y) + g(y)).
\end{equation}
In which \( g^c(x)  \) and \( f^c(y) \) are Legendre c-transforms of \(g(y) \in \mathcal{C}(\mathcal{Y})\) and  \(f(x)\in \mathcal{C}(\mathcal{X})\) with cost function \(c(x,y)\).  
\end{definition}

\begin{definition}[\textit{Pushforward measure}]
The pushforward measure of \( \mu \) under the map \( T \), denoted as \( T_{\mu} \), is a measure on \( \mathcal{X} \) defined by \( T_\mu(B) = \mu(T^{-1}(B)) \) for any Borel set \( B \) in \( \mathcal{X} \). \( T_\mu = \nu \) when \( T \) is an optimal transport map. Following the similar way we can define the push-forward measure \(T_\mu\) and \(T_\nu\) as:
\begin{equation}
T_\mu : C(\mathcal{X}) \rightarrow C(\mathcal{Y}): = \log \int{e^{-c(x,\cdot)+f(x)}}{\mu(x)},~~T_\nu : C(\mathcal{Y}) \rightarrow C(\mathcal{X}): = \log \int{e^{-c(\cdot,y)+g(y)}}{\nu(y)} 
\end{equation}

\end{definition}

\begin{definition}[\textit{$\varphi$-divergence regularized OT in continuous}]\label{def:phidivergence}
Given two dual variables (also called potentials) \( f \in \mathbb{R}^n \) and \( g \in \mathbb{R}^m \) for each marginal constraint, the entropy regularized optimal transport in Equation~\eqref{eq:eot} could be transformed into some problems with the Kantorovich functional:
\begin{equation}\label{eq:dualphiot}
\begin{aligned}
W^\varphi_{\varepsilon,c}(\mu, \nu) = \inf_{\pi in \Pi(\mu,\nu) }  ( \int_{\mathcal{X}\times\mathcal{Y}} c(x,y) d\pi(x,y) + \varepsilon\int_{\mathcal{X}\times\mathcal{Y}} \varphi  \left( \frac{d\pi(x,y)}{d\mu(x)d\nu(y)} \right) d\mu(x) d\nu(y) \Bigg)
\end{aligned}
\end{equation}
\end{definition}

\begin{proposition}[Dual of EOT]\label{eq:eotdual}
Consider OT between two probability measures \(\mu\) and \(\nu\) with a convex
regularizer \(\phi\) on \(\mathbbm{R}^+\) in Equation~\eqref{eq:dualphiot}
\begin{equation}
W^\varphi_{c,\varepsilon}(\mu, \nu) = \sup_{f,g \in \mathcal{C}(\mathcal{X})\times\mathcal{C}(\mathcal{Y})} \int_{\mathcal{X}} fd\mu(x)+\int_{\mathcal{Y}} d\nu(y)-\varepsilon\int_{\mathcal{X}\times\mathcal{Y}}\varphi^*(\frac{f(x)+g(y)-c(x,y)}{\varepsilon})d\mu(x)d\nu(y)
\label{eq:wdual}
\end{equation}
where \(\varphi^*\) is the Legendre transform of \(\varphi\) defined by \(\varphi^*({\bf v}):=\sup_\x \x{\bf v}-\phi(\x)\)
\end{proposition}

A good choice for \(\varphi^*\) is that the \(\varphi^*({\bf v})=e^{\bf v}\).
The entropy regularization term ensures the problem is solvable, especially for computational schemes. If the \( \varepsilon \rightarrow \infty \), the optimal primal plan \( \pi^* \) can be retrieved using, which corresponds to the mutual information formula:
\[
\frac{d \pi^*}{d\mu d\nu}(x, y) = \exp \left( \frac{f^*(x) + g^*(y) - c(x,y)}{\varepsilon} \right)
\]

In the discrete version in Equation~\eqref{eq:eot}, the optimal transport plan \( \PP \) can often be expressed in terms of the optimal transport map \( T^* \) when it exists,  one can define the so-called barycentric projection map
\[ T^* : \x_i \in \mathcal{X} \mapsto \frac{1}{\mu_i} \sum_{j} \PP_{i,j} \y_j \in \mathbb{R}^d, \]
This link provides the connection between the mutual information with the optimal mapping:

\[ T^* : x \in X \mapsto \int_{Y} y \frac{d\pi(x, y)}{d\mu(x)d\nu(y)} d\nu(y). \]

Note that the joint distribution $\pi$ always has a density $\frac{d\pi(x,y)}{d\mu(x)d\nu(y)}$ with respect to $\mu \otimes \nu$, and the mutual information method will lead us to the optimal solution.

\section{Analysis of GCA}

\subsection{Convergence of GCA}
\label{app:gca_converged}

In this section, we provide a proof of  convergence in the forward pass for our GCA algorithm.
To do this, we show the general form in Algorithms~\ref{alg:gca} for all Bregman divergence (\(d_\Gamma\)) in forward pass in GCA algorithms could be converged through Djkstra’s projection algorithms. Finally, we show the uniformly convergence of the transport plan \(\PP\), and the convergence of its dual variables \(f^{(t)}\) in each iteration.

\begin{algorithm}[t!]
    \caption{Generalized Contrastive Alignment (GCA)\label{alg:example}}
\begin{algorithmic}
        \State \textbf{Input:} Encoder $f_\theta$, projector $g_\theta$, data $\{\x_k\}_{k=1}^N$, batch size $B$, cost function $c(x,y)$, entropy parameter $\varepsilon$, constant $\tau$, total iterations $T$, marginal constraints $\mu$ and $\nu$, relax items $d_1, d_2$ and constant $\delta_{\text{eps}}$, some divergence $d_M$ and $d_\Gamma$ (could be KL or WDM),
        \For{sampled minibatch $\{\x_k\}_{k=1}^{B}$}
            \State Generate two views $(\z'_{k}, \z''_{k})$ using $f_\theta$, $g_\theta$ with randomly sampled augmentations.
        \EndFor
        \State $\mathbf{u}^{(0)} = \mathbbm{1}$, $\mathbf{v}^{(0)}=\mathbbm{1}$, $f=0$, $g=0$, $\C_{ij} = c(\z'_i , \z_j'')$
        \State $d_1 \gets d_1/ (d_1 + \varepsilon)$, $d_2 \gets d_2/ (d_2 + \varepsilon)$, $\K=\exp(C_{ij}/\varepsilon^{-1})$  
        \For{$i=1$ {\bfseries to} $T$}
            \State $\delta f \leftarrow \exp{ -f/(\varepsilon + d_1)}$, $\delta g \leftarrow \exp{ -g/(\varepsilon + d_2)}$ 
            \State $\mathbf{u} \leftarrow \delta f \cdot \text{Prox}_\mathcal{F}(K\mathbf{v}+\delta_{\text{eps}})^{fi}$
            \State $\mathbf{v} \leftarrow \delta g \cdot \text{Prox}_\mathcal{G}(K^T\mathbf{u}+\delta_{\text{eps}})^{fi}$
            \If {$\mathbf{u} > \tau$ \Or $\mathbf{v}>\tau$}
                \State $f\leftarrow f+\varepsilon \cdot \log(\max(\mathbf{u}))$, $g\leftarrow g+\varepsilon \cdot \log(\max(\mathbf{v}))$
                \State $K \leftarrow \exp{(f+g-\C)}/\varepsilon$, $\mathbf{v}=\mathbbm{1}$
            \EndIf
        \EndFor
        \State $\log \mathbf{u} \gets f /  (\varepsilon+\mathbf{u})$, $\log \mathbf{v} \gets g / (\varepsilon+\mathbf{v})$
        \State Compute transport plan as:
        \State $\PP \gets \exp (\log \mathbf{u}+\log \mathbf{v} - C/\varepsilon)$
        \State Normalize $\PP_{u}^{(T)}$ by its column sums. 
        \State Loss: $\mathcal{L}_{GCA} = d_M ( {\bf \mu\otimes\nu}, \PP_{u}^{(T)} )$
        \State Update networks $f_\theta$ and $g_\theta$ to minimize $\mathcal{L}_{GCA}$
    \end{algorithmic}
\end{algorithm}

\subsubsection{Convergence of GCA-INCE}

\begin{corollary}  (Convergence of GCA-INCE~\cite{peyre2019computational}) \label{Convergence}
Given \(\uu^*=e^{f^{\infty}}\), \(\vv^*=e^{g^{(\infty)}}\) and a kernel space \(H\) with the Hilbert-Birkhoff metric \(d_H(\uu, \uu^*) :=\log \max_{i, j}  \frac{\uu_i \uu^*_j}{\uu_j \uu^*_i},\) for all positive pairs \((\uu, \uu^*)\), with \(\uu^{(t)}\rightarrow \uu^*\) and \(\vv^{(t)}\rightarrow \vv^*\), we can prove that:
\begin{equation}
\|\log\PP^{(t)} - \log\PP^* \|_{\infty} \leq d_H(\uu^{(t)}, \uu^*) + d_H(\vv^{(t)}, \vv^*),
\end{equation}
\end{corollary}

\textbf{Proof for Corollary~\ref{Convergence}:}
First, let's define the following Hilbert space:
 \(\forall (\uu, \uu') \in (\mathbb{R}^n_{+, *})^2, \quad d_H(\uu, \uu') :=\log \max_{i, j}  \frac{\uu_i \uu'_j}{\uu_j \uu'_i}.\) 
For any pairs of vectors that $(\vv, \vv') \in (\mathbb{R}^m_{+, *})^2$ holds:
\begin{equation}
d_H(\vv, \vv') = d_H\left(\frac{\vv}{\vv'}, \mathbbm{1}_m\right) = d_H\left(\mathbbm{1}_m/\mathbf{v}, \mathbbm{1}_m/\mathbf{v}'\right).
\end{equation}
Let $\K \in \mathbb{R}^{n \times m}_{+,*}$, then for $(\vv, \vv') \in (\mathbb{R}^m_{+, *})^2$ we have
\[
d_H(\uu^{(t+1)}, \uu^{*}) = d_H\left(\frac{\mathbbm{1}_n}{\K\mathbf{\vv}^{(t)}}, \frac{\mathbbm{1}_n}{\K\mathbf{\vv}^{*}}\right) = d_H(\K\mathbf{\vv}^{(t)}, \K\mathbf{\vv}^{*}) \leq \lambda(\K)d_H(\mathbf{\vv}^{(t)}, \mathbf{\vv}^{*}),
\]

where
\begin{equation}
\lambda(\K) := \frac{\sqrt{\eta(\K) - 1}}{\sqrt{\eta(\K) + 1}} < 1, \quad \eta(\K) := \max_{i,j,k,\ell} \frac{\K_{i,k} \K_{j,\ell}}{\K_{j,k} \K_{i,\ell}}.
\end{equation}
Based on the contraction mapping theory, one has $(\uu^{(\ell)}, \vv^{(\ell)}) \rightarrow (\uu^*, \vv^*)$ and
\begin{align}
d_H(\uu^{(t)}, \uu^{*})  & \leq d_H(\uu^{(t+1)}, \uu^{(t)}) + d_H(\uu^{(t+1)}, \uu^{*}) \\ & \leq d_H\left(\frac{\mu}{\K\mathbf{\vv}^{(t)}}, \uu^{(t)}\right) + \lambda(\K)^2d_H(\uu^{(t)}, \uu^{*}) \\ & = d_H\left(\mu, \uu^{(t)} \odot (\K\mathbf{\vv}^{(t)})\right) + \lambda(\K)^2d_H(\uu^{(t)}, \uu^{*}),
\end{align}
\begin{equation}
d_H(\uu^{(t)}, \uu^*) \leq \frac{d_H(\PP^{(t)} \mathbbm{1}_m, \mu)}{1 - \lambda(\K)^2}, \quad d_H(\vv^{(t)}, \vv^*) \leq \frac{d_H(\PP^{(t)\top} \mathbbm{1}_n, \nu)}{1 - \lambda(\K)^2},
\end{equation}
where we denoted $\PP^{(t)} := \diag(\uu^{(t)}) \K \diag(\vv^{(t)})$. Last, one has
\begin{equation}
\| \log(\PP^{(t)}) - \log(\PP^*) \|_{\infty} \leq d_H(\uu^{(t)}, \uu^*) + d_H(\vv^{(t)}, \vv^*),
\end{equation}
where $\PP^*$ is the unique solution of Equation~\eqref{eq:eot}. The above formula also shows that the t-step solution gives a better lower bound than the 1-step solution. \qed

\subsubsection{Convergence of the Djkstra's projection algorithms}
The previous subsection proved the convergence of GCA-INCE. Here, we extend this to show the convergence of all generalized proximal operators in Algorithm~\ref{alg:gca}. Additionally, we demonstrate that these operators can iteratively solve alignment problems in the forward pass, following Dykstra’s projection algorithm.

We present a general convergence proof for Dykstra’s projection algorithm, sharing the form in Definition~\eqref{eq:prox}. 
First, we define $d_\Gamma$ as a generic Bregman divergence on some convex set $\mathcal{B}$, and the proximal map of a convex function $d_\phi$ according to this divergence is:
\begin{equation}
\text{Prox}^{d_\Gamma}_{d_\phi,\mathcal{B}}(\K) := \arg\min_{\tilde{\PP} \in \mathcal{B}} d_\Gamma( \tilde{\PP} \|\K) + d_\phi( \tilde{\PP}).
\end{equation}
$\Gamma$ is a strictly convex function smooth on $\text{int}(\mathcal{B})$, and $\text{Prox}^{d_\Gamma} _{d_\phi}(\K) \in \text{int}(\mathcal{B})$ is always uniquely defined by strict convexity. As $\mathcal{B}= \text{dom}(\Gamma)$, $$\forall (\PP,\K) \in \mathcal{B} \times \text{int}(\mathcal{B}), d_\Gamma(\PP\|\K) = \Gamma(\PP) - \Gamma(\K) - \langle \nabla \Gamma(\K), \PP - \K \rangle,$$
which has its Legendre transform is also smooth and strictly convex: $$\Gamma^*(\rho) = \max_{\PP \in \mathcal{B}} \langle \PP, \rho \rangle - \Gamma(\PP)$$
In particular, one has that $\nabla\Gamma$ and $\nabla\Gamma^*$ are bijective maps between $\text{int}(\mathcal{B})$ and $\text{int}(\text{dom}(\Gamma^*))$ such that $\nabla\Gamma^* = (\nabla\Gamma)^{-1}$. For $\Gamma = ||\cdot||^2$, one recovers the squared Euclidean norm $d_\Gamma = ||\cdot||^2$. One has KL = $d_\Gamma$ for $\Gamma(\PP) = h(\PP)=-\sum_{i,j=1}^B(\PP_{ij} (\log\PP_{ij}-1))$. Dykstra’s algorithm starts by initializing $\PP^{(0)} := \K$ and $\U^{(0)} = \U^{(-1)} := 0$. One then iterative defines, for $k > 0$, 
\begin{align}
\PP^{(k)} &:= \text{Prox}^{d_\Gamma}_{d_{\phi_{[k]_2}}}(\nabla\Gamma^*(\nabla\Gamma(\PP^{(k-1)})+\U^{(k-2)})),\\
\U^{(k)} &:= \U^{(k-2)} + \nabla\Gamma(\PP^{(k-1)}) - \nabla\Gamma(\PP^{(k)}),
\end{align}

\begin{proposition}~\label{prop:pi}
Giving $d_{\phi_1},d_{\phi_2}$ are two proper, lower-semicontinuous convex functions defined on $\mathcal{B}$. We also assume that the following qualification constraint holds:
\begin{equation}~\label{eq:qual}
\text{ri}(\text{dom}(d_{\phi_1})) \cap \text{ri}(\text{dom}(d_{\phi_2})) \cap \text{ri}(\text{dom}(d_\Gamma)) = \emptyset,
\end{equation}
where ri is the relative interior and $\text{dom}(\phi) = \{\pi;\phi(\pi) = +\infty\}$. Then the \(\PP^{t}\) converges to the solution of the following equation: 
\begin{equation}\label{eq:pi}
\text{prox}_{h,\mathcal{B}}^{d_\Gamma}(\K)  = \arg\min_{\PP \in \mathcal{B}}\{  d_\Gamma ( \PP \| \K)  + \lambda_1 d_{\phi_1}(\PP) + \lambda_2 d_{\phi_2}(\PP)\}
\end{equation}
\end{proposition}

\textbf{Proof of the Proposition~\ref{prop:pi}}: Proof in \cite{peyre2015entropic} section 3.2.

\subsubsection{Convergence of Bregman projection}
\label{app:monotonicity}
This section aims to show for the continuous measure, the convergence of Bregman projection holds~\cite{berman2020sinkhorn}. Finding the optimal transport map \( T \) could be derived in minimizes some functionals derived from the potential function \( f \) defined on \(\mathcal{X}\), with Legendre transform in Equation~\eqref{eq:legendre} defined on \(\mathcal{Y}\):
\begin{equation}~\label{eq:funtional}
J(f) := \int_{\mathcal{X}} f \mu(x) + \int_{\mathcal{Y}} f^\mathsf{c} \nu(y)= I_\mu  -L  
\end{equation}

\begin{lemma} [\textit{Uniformly convergence}]\cite{berman2020sinkhorn} \label{lm:monotonicity} When \(t_1\rightarrow \infty \), \(f^{(t_1)}\) converges uniformly to a fixed point \( f^{(\infty)} \), with \(f^{(t_1)}\leq f^{(\infty)}\).
\end{lemma}

\textbf{Proof for the Lemma~\ref{lm:monotonicity}} (\textit{Uniformly convergence}): We follow the procedures of methods in ~\cite{berman2020sinkhorn}.

Giving push-forward measure \(T_\mu\) and \(T_\nu\) and a composed operator \(S= T_\nu \circ T_\mu \), which yields an iteration on \( C(\mathcal{X}) \) as \( S: C(\mathcal{X}) \rightarrow C(\mathcal{X}), f\rightarrow f\circ g\circ f\), \(f^{(m+1)}=S(f^{(m)})\), and \(e^{S(f)-f} \mu\) is the probability measure on \(\mathcal{X}\). 

\begin{lemma}[\textit{Existence and uniqueness}]
The following conditions are equivalent for a function \( f \) in the space \( C(X) \), where \( C(X) \) denotes the space of continuous functions on a set \( X \):
\begin{itemize}
    \item \( f \) is a critical point for the functional \( F \) on \( C(X) \).
    \item The function \(\exp(S(f)-f)=0\) hold almost everywhere (a.e.) with respect to (w.r.t.) $\mu$.
\end{itemize}
Moreover, if \( f \) is a critical point, then \( f^* := S(f) \) is a fixed point for the operator \( S \) on \( C(X) \).
\end{lemma}

\textbf{Proof of the Lemma~\ref{lm:unique}}
Consider the functional \( L \) defined in Equation~\eqref{eq:funtional}, the differential of \( L \) at an element \( f \in C(X) \) is represented by the probability measure \(\exp({S(f)-f})  \mu \). For some iterations \(f^{(m+1)}-f^{(m)}=S(f^{(m)})-f^{(m)}\), when \(f\) is a critical point (derivative is zero or undefined) for the functional $J$ on $C(X)$ , and $f^* := S(f)$ is a fixed point for the operator $S$ on $C(X)$, proved by realizing for any $\dot{f} \in C(X)$:

\begin{equation}
    \left. \frac{d}{dt} L(f + t\dot{f}) \right|_{t=0} = \int_X \dot{f} e^{(S(f)-f)} d\mu.
\end{equation}
This follows readily from the definitions by differentiating $t \mapsto g[(f + t\dot{f})]$ to get an integral over $(X, \mu)$ and then switching the order of integration. As a consequence, \( f \) is a critical point of the functional \( F \) on \( C^0(X) \) if and only if \( e^{(S(f)-f)} \mu = \mu \), i.e., if and only if \( e^{(S(f)-f)} = 1 \) almost everywhere with respect to \( \mu \). Finally, if this is the case, then \( S(f) = f \) almost everywhere with respect to \( \mu \) and hence \( S(S(f)) = S(f) \) (since \( S(f) \) only depends on \( f \) viewed as an element in \( L^1(X, \mu) \)).

\begin{lemma}
Given a point \( x_0 \in X \), the subset \( K_{x_0} \) of \( C(\mathcal{X}) \) defined as all elements \( f \) in the image of \( S \) satisfying \( f(x_0) = 0 \) is compact in \( C(\mathcal{X}) \).
\label{lm:compact}
\end{lemma}
\textbf{Proof of the Lemma~\ref{lm:compact}:} Based on the compactness of the product space \(\mathcal{X}\times\mathcal{Y}\), the continuous function \( c \) is uniformly continuous on \( \mathcal{X} \). So \( S(C(\mathcal{X})) \) is an equicontinuous family of continuous functions on \( X \). By Arzelà-Ascoli theorem, it follows that the set \( K_{x_0} \) is compact in \( C(\mathcal{X}) \).

\begin{proposition}
The operator \( S \) has a fixed point \(f^* \) in \( C(\mathcal{X}) \). Moreover, \( f^* \) is uniquely determined a.e. wrt \( \mu \) up to an additive constant, and \( f^* \) minimizes the functional \( F \). More precisely, there exists a unique fixed point in \( S(C(\mathcal{X}))/\mathbb{R} \).
\label{pp1}
\end{proposition}
\textbf{Proof of the Proposition~\eqref{pp1}:} Then based on the Jensen’s inequality, we have 
\begin{align}
& I_\mu(f^{(m+1)}) - I_\mu(f^{(m)}) = \int\log {\exp{(S(f^{(m)})-f^{(m)}}})d\mu \leq \log \int{\exp{(S(f^{(m)})-f^{(m)})}}{d\mu}  = 0, \\
&L(f^{(m)}) - L(f^{(m+1)}) = \int \log {\exp{(S(g^{(m)})-g^{(m)})}}{d\nu} \leq \log \int{\exp{(S(g^{(m)})-g^{(m)})}}{d\nu} = 0.
\end{align}
So we know the functionals are strictly decreasing at \( f^{(m)} \) unless \( S(f^*) = f^* \) for \( f^* := S(f^{(m)}) \). Then based on the Lemma~\ref{lm:compact}, we know for each initial data \( f_0 \), the closure of its images denoted as \( K_{f_0} \) in \( C(\mathcal{X})/\mathbb{R} \) is compact, under the operator \(S\). Hence, \( f^{(m)} \rightarrow f^{(\infty)} \) in \( C(\mathcal{X})/\mathbb{R} \). And \( J \) is decreasing along the orbit but has lower bound:
\[ J(f^{(\infty)}) = \inf_{K_{f^{(0)}}} J. \]
By the condition for strict monotonicity, it must be that \( S (f^{(\infty)}) = f^{(\infty)} \) a.e. wrt \( \mu \). It then follows from the Proposition~\eqref{pp1} that \( f^{(\infty)} \) is uniquely determined in \( C(\mathcal{X})/\mathbb{R} \) (by the initial data \( f^{(0)} \)), i.e. the whole sequence converges in \( C(\mathcal{X})/\mathbb{R} \). We first show that there exists a number \( \lambda \in \mathbb{R} \) such that
\( \lim_{m \rightarrow \infty} I_\mu(f^{(m)}) = \lambda. \) \( I_\mu \) is decreasing and hence it is enough to show that \( I_\mu(f^{(m)}) \) is bounded from below. By \( I_\mu = J + L \), and \( J \) is bounded from below (by \( F(f^{(\infty)}) \)). Moreover, by the first step \( L(f^{(m)}) \geq L(f^{(0)}) \). Next, decompose
\[ f^{(m)} = \tilde{f}^{(m)} + f^{(m)}(x_0), \]
By the Lemma~\ref{lm:compact} the sequence \( (\tilde{f}^{(m)} \) is relatively compact in \( C(\mathcal{X}) \) and we claim that \( |f^{(m)}(x_0)| \leq C \) for some constant \( C \). Indeed, if this is not the case then there is a subsequence \( f^{(m_j)} \) such that \( |f^{(m_j)}| \rightarrow \infty \) uniformly on \( X \). But this contradicts that \( I_\mu(f^{(m)}) \) is uniformly bounded. It follows that the sequence \( (f^{(m)}) \) is also relatively compact. Hence, by the previous step the whole sequence \( f^{(m)} \) converges to the unique minimizer \( f^* \) of \( F \) in \( S(C(\mathcal{X})) \) satisfying \( I_\mu(f^*) = \lambda \).

\subsection{GCA version of unbalanced optimal transport (GCA-UOT)}
\label{app:gcauot}

In this section, we are going to introduce the relaxation of the EOT plan as Unbalanced optimal transport plan (UOT). And its relationship with the dual formula of EOT. Here we need to emphasize that the GCA-UOT not just add constraint to the proximal operators which computes the coupling matrix \(\PP_\theta\), but also add the penalty (i.e. KL-divergence) to the loss function \(d_M\). For the specific function we used in the method of GCA-UOT in Table~\ref{tab:new_combine}, we employed a version with the loss in Equation~\eqref{eq:hsrince} plus the loss in Equation~\eqref{eq:kleqince} with a weight control parameter.

\subsubsection{Explanation of the unbalanced OT} Unbalanced optimal transport (UOT) in Equation~\eqref{eq:uot} seeks to generalize the OT problem in Equation~\eqref{eq:dot} by allowing for the relaxation of these constraints~\cite{chizat2018scaling}, as penalization by certain divergence measures \( d_{\phi_1} \) and \( d_{\phi_2} \) (e.g., Kullback-Leibler divergence). Here we provide the unbalanced OT for the entropic regularization optimal transport in Equation~\eqref{eq:eot}, which ensure that the transported mass respects the given source \(\mu\) and target distributions \(\nu\):

\begin{equation}\label{eq:ueot}
\U_{OT}(\mu, \nu) = \min_\PP \left\langle \PP,\C \right\rangle + \lambda_1 d_{\phi_1}(\PP\mathbbm{1} || \mu) + \lambda_2 d_{\phi_2}(\PP^\top\mathbbm{1} || \nu) + \varepsilon H(\PP)
\end{equation}

Here \( \left\langle P,C \right\rangle \) represents the total transport cost. \( \lambda_1 \) and \( \lambda_2 \) are regularization parameters that control the trade-off between the transport cost and the divergence penalties. 

\subsubsection{Connection to dual formula of EOT} 
\begin{lemma}\label{lm:uotdual}
The entropy regularized OT problem is a special case of a structured convex optimization problem of Equation~\eqref{eq:ueot} the by giving functions \(h_\mathcal{F}\) and \(h_\mathcal{G}\), $h_\mathcal{F} = \iota_{\{C_1^\mu\}}$ and $h_\mathcal{G} = \iota_{\{C_2^\nu\}}$, as the indicator function of a closed convex set     \(C_1^\mu \coloneqq \{\PP : \PP\mathbbm{1}_m = \mu\}, C_2^\nu \coloneqq \{\PP : \PP^\top \mathbbm{1}_n = \nu\}.\) 
\begin{equation}
\min_{\PP} \langle\PP,\C\rangle + \varepsilon H(\PP) + h_\mathcal{F}(\PP \mathbbm{1}_m) + h_\mathcal{G}(\PP^\top \mathbbm{1}_n).\quad \iota_C(x) = 
\begin{cases}
0 & \text{if } x \in C, \\
+\infty & \text{otherwise},
\end{cases} 
\end{equation}
\end{lemma}

\textbf{Proof of the Lemma~\ref{lm:uotdual}:} Let's start with the dual formula of the Equation~\eqref{eq:eot} with \(\mathcal{B}=C_1^\mu \cap C_2^\nu\), we can introduce the Lagrangian \(\mathcal{E}(\PP,f, g)\) of Equation~\eqref{eq:eot} reads:
\begin{align}
 \text{Prox}^{\text{KL}}_{\mathcal{B}}(\K) & := \min_{\PP\in\mathcal{B}}\langle\PP,\C\rangle -\varepsilon H(\PP) = \mathcal{E}(\PP,f, g) \\ &   = \min_{P} \max_{f \in \mathbb{R}^n, g \in \mathbb{R}^m} \langle \PP, \C \rangle - \varepsilon H(\PP) - \langle f, \PP \mathbbm{1}_m - \mu \rangle - \langle g, \PP^T \mathbbm{1}_n - \nu \rangle. \qed
\label{eq:LpC}
\end{align}
To solve this problem, we can use the first order condition:
\begin{equation}
\frac{\partial \mathcal{E}(\PP,f, g)}{\partial \PP_{ij}} = \C_{ij} + \varepsilon \log(\PP_{ij}) - f_i - g_j = 0 \quad \Rightarrow \quad log \PP =\frac{1}{\varepsilon }(f \mathbbm{1}_m^T + \mathbbm{1}_n g^T - \C) 
\end{equation}

The solution to the Equation~\eqref{eq:eot} is unique with scaling variabl \( (\uu, \vv) \in \mathbb{R}^n_+ \times \mathbb{R}^m_+ \) in Equation~\eqref{eq:uvupdatesapp}. And each items in the optimal transport matrix \( \PP \) is, and optimal \( (f, g) \) are linked to non-negative vectors \( (\uu, \vv) \) through $(\uu, \vv) = (e^{f/\varepsilon}, e^{g/\varepsilon})$.
\begin{equation}
\PP_{ij} = e^{f_i/\varepsilon} e^{-\C_{ij}/\varepsilon} e^{g_j/\varepsilon} = \uu_i \K_{ij} \vv_j, \quad (f^{(t)}, g^{(t)}) = \varepsilon(\log(\uu^{(t)}), \log(\vv^{(t)})),
\end{equation}

\subsection{Equivalence of INCE objective with single step Bregman projection}

In this section, we are going to discuss how to build the equivalence between minimizing the KL-divergence \(d_M\) between the \(\PP^{(1)}\) and the \(\PP_{\text tgt}\) with respect to \(\theta\) in GCA objective:
$$\min_{\theta}\text{KL} \big( {\bf I}  ||\text{Prox}_{C_1^\mu}^{KL}(\K_\theta)),$$ 
with the INCE loss minimization in Equation~\eqref{eq:INCE}.
Here \(\PP^{(1)}\) is the nearest point of \(\K_\theta\) on constraint set \(C_1^\mu\) measured by the KL-divergence $d_\Gamma$ defined in Equation~\eqref{eq:breg1}, through one step of proximal operator (Bregman projection).
And ${\bf K}_\theta$ denote the augmentation kernel as in Definition~\eqref{def:gibbs} with cosine similarity.

\label{app:proof_cleqot}
\subsubsection{Proof of the Theorem~\ref{thm:cl_eq_ot}}  Suppose we had a encoder \(f_\theta\) with parameter \(\theta\) in INCE, with $\widetilde{f}_\theta$ to represent its normalized form,then we can use the following proposition to assist our proof:

\begin{proposition}~\label{prop:pij}
Given the cost matrix as $\C_{i,j}=1-\widetilde{f}_\theta(\x_i')^\top \widetilde{f}_{\theta}(\x_j'')$, and  Gibbs kernel \({\bf K}_\theta=\exp(-\C_{i,j}/\varepsilon)\), based on the cosine dissimilarity scores of the inner products \( \langle \z_{\theta i}, \z_{\theta j} \rangle \), with \( \z_i=\frac{f_\theta(\x'_i)}{\|f_\theta(\x'_i)\|} \) and \( \z_j = \frac{f_\theta(\x''_j)}{\|f_\theta(\x''_j)\|} \). Set \(d_M\) and \(d_\Gamma\) to KL-divergence, and the target  transport plan \({\PP}_{\text{tgt}}=\bf I\). The probability matrix \(\PP\) after one-step Bregman iteration of entropy optimal transport problem could be represented as: 
\begin{equation}
\PP_{ij} = \frac{\K_{\theta ij}}{\sum_{j=1}^B \K_{\theta ij} } = \frac{\exp\left(\varepsilon^{-1}\langle \z_i, \z_j \rangle\right)}{\sum_{j=1}^B \exp\left(\varepsilon^{-1}\langle \z_i, \z_k \rangle\right)}
\end{equation}
\end{proposition} 
\textbf{Proof of the Proposition~\eqref{prop:pij}}: 
We assume that gibbs kernel \( \K_\theta \) is a matrix  which can be expressed as:
\[
\K_{\theta ij}=\exp\left(-\varepsilon^{-1}\C_{i,j}\right) = \exp\left(-\varepsilon^{-1} |1 -\langle \z_i, \z_j \rangle |\right),
\]
with a temperature parameter \(\varepsilon\). \( \mu \), \( \nu \), \(\uu^{(0)}\) and \(\vv^{(0)}\)  can be initialized as a vector of ones with the same size as B, the batch size, \[\mu = \mathbbm{1}, \nu = \mathbbm{1}\quad \uu^{(0)}=\mathbbm{1}, \vv^{(0)}=\mathbbm{1}.\]

For \( t \) iterations of the Sinkhorn algorithm, \( \uu^{(t)}\) is updated as:
\[
\mathbf{\uu}^{(t+1)} \stackrel{\text { def }}{=} \frac{\mu}{\K_\theta \mathbf{\vv}^{(t)}}, \quad
\mathbf{\vv}^{(t+1)} \stackrel{\text { def }}{=} \frac{\nu}{\K_\theta^T \mathbf{\uu}^{(t)}}.
\]
So we know that:
\[
\uu^{(1)}=\frac{1}{\sum_{j=1}^b \mathbf{K}_{\theta ij}}.
\]
Thus, half-step sinkhorn iteration or one-step Bregman interation for \( \PP \) can be expressed as:
\[
\PP_{ij} = \uu^{(1)}_i \mathbf{K_\theta}_{ij} \vv^{(0)}_j= \frac{\K_{\theta ij}}{\sum_{j=1}^b \K_{\theta ij} } = \frac{\exp\left(\varepsilon^{-1}\langle \z_i, \z_j \rangle\right)}{\sum_{j=1}^b \exp\left(\varepsilon^{-1}\langle \z_i, \z_k \rangle\right)} \qed 
\]

This concludes the expressions of \(\PP\) at half-step iteration. Reminds us the formula of the KL divergence \(\text{KL}(\I\|\PP)\) and the entropy \(H(\PP)\):
\begin{equation}
\text{KL}(\I\|\PP) \ \stackrel{\text { def }}{=} \sum_{i,j} \I_{i,j} \log \frac{\I_{i,j}}{\PP_{i,j}} - \I_{i,j} + \PP_{i,j}, \quad \text{where}\quad \I_{i,j} \log \frac{\I_{i,j}}{\PP_{i,j}}=0, \quad \text{if}\quad \I_{i,j}=0.
\end{equation}
And after the batch normalization of \(\PP\), the value of \(\sum_{i,j}\PP_{i,j}\) is equal to the batch size B and exactly the same as the \(\sum_{i,j} \I_{i,j}\), we can obtain:

\[
\operatorname{KL({\bf I}\|\PP}) = \sum_i \log\left(\frac{1}{\PP_{ii}}\right)= -\sum_i \log\frac{\exp\left(\varepsilon^{-1}\langle \z_i, \z_i \rangle\right)}{\sum^b_{j=1} \exp\left(\varepsilon^{-1}\langle \z_i, \z_j \rangle\right)}
\]
j represents the elements on the diagonal of the similarity matrix, which is the same structure as the INCE loss as:

\[
\mathcal{L}_{\text{INCE}} = -  \sum_i \log\left(\frac{\exp(f_\theta(\x'_i)^\top f_\theta(\x''_i))}{\sum_{j=1}^b\exp(f_\theta(\x'_i)^\top f_\theta(\x''_j))}\right) \qed
\]

\subsection{Proximal operator version of RINCE}

In this section, we are going to discuss how to build the equivalence between minimizing the some convex function of \(d_M\) with adjustable parameters \(q\) and \(\lambda\) between the \(\PP^{(1)}\) and the \(\PP_{\text tgt}\) as:

\begin{equation}\label{eq:rincedm}
d_M(\I,\PP)=-\frac{1}{q}\left(\left(\frac{\operatorname{diag}(\PP^{(1)}_\theta)}{\uu^{(1)}}\right)^q-\left(\frac{\lambda \I}{\uu^{(1)}}\right)^q\right)
\end{equation}

with respect to \(\theta\) in GCA objective:
$$L_{\text{RINCE}}^{\lambda,q} = \min_\theta -\frac{1}{q} \bigg( \frac{\operatorname{diag}(\PP^{(1)}_\theta)}{\uu^{(1)}} \bigg)^q+  \frac{1}{q}\bigg(\frac{\lambda \I}{\uu^{(1)}} \bigg)^q,~~\text{with}~~\PP^{(1)}_\theta = \text{Prox}_{C_1^\mu}^{\text{KL}}(\K_\theta),~~\uu^{(1)}=\text{diag}\left(\frac{\mu}{\PP^{(0)}\mathbbm{1}}\right)$$
with the RINCE loss minimization in Equation~\eqref{eq:RINCE}.
Here \(\PP^{(1)}\) is the nearest point of \(\K_\theta\) on constraint set \(C_1^\mu\) measured by the KL-divergence $d_\Gamma$ defined in Equation~\eqref{eq:breg1}, through one step of proximal operator (Bregman projection).
And ${\bf K}_\theta$ denote the augmentation kernel as in Definition~\eqref{def:gibbs} with cosine similarity.

Also, we are going to discuss when the q=1, RINCE loss is the symmetry loss, which provides the robustness in the noisy view. 

\subsubsection{Proof of the Theorem~\ref{thm:rince}}
\label{app:rinceeqprox}
The loss function of RINCE looks like:
\begin{equation}\label{eq:rinceapp}
    \mathcal{L}^{\lambda,q}_\text{RINCE}=\frac{1}{q}\big(-e^{ q s_{ii}} + \lambda^q(e^{ s_{ii}}+\sideset{}{_{i\neq j}}\sum e^{s_{ij}})^q\big)
\end{equation}
For the specific parameters \(\theta\), we record the normalized latent of the \(\z^i_{\theta+} =s_{ii}\), and \(\z^i_{\theta-} =s_{ij},j\neq i\). The positive pairs are stored in the diagonal of the gibbs kernel \(\K_\theta\), and the negative pairs are stored in the off-diagonal elements, which means:
\begin{align}    
&\K_{ii}=\exp\left(-\varepsilon^{-1}\C_{i,i}\right) = \exp\left(-\varepsilon^{-1} |1-\langle \z'_{\theta i}, \z''_{\theta i} \rangle|\right)=\exp\left(\varepsilon^{-1}{\langle \z'_{\theta i}, \z''_{\theta i} \rangle}-\varepsilon^{-1}\right) \propto  e^{\z^i_{\theta+}}.\\
&\K_{ij}=\exp\left(-\varepsilon^{-1}\C_{i,j}\right) =\exp\left(\varepsilon^{-1}{\langle \z'_{\theta i}, \z''_{\theta j} \rangle}-\varepsilon^{-1}\right) \propto  e^{\z^i_{\theta-}}, j\neq i .
\end{align}

By solving the \(\langle \uu^{(1)}\K, \mathbbm{1}\rangle = \mu\) in the Equation~\eqref{eq:scaleapp}, we have the ith column elements \(\sum_{j=1}\K_{\theta ij}=\frac{\mu}{\uu^{(1)}_i}\), in which \(\uu^{(1)}\) is given in~\ref{eq:scaleapp}:
\begin{align}
& \frac{\mu}{\uu^{(1)}_i}= \sum^B_{j=1}\K_{\theta ij}=\frac{1}{e^{\varepsilon^{-1}}} ( e^{\varepsilon^{-1}\langle \z'_{\theta i}, \z''_{\theta i} \rangle} + \sum_{j=1,j\neq i}^{B} e^{\varepsilon^{-1}{\langle \z'_{\theta i}, \z''_{\theta j} \rangle}}), i\neq j, \\ & \operatorname{diag}(\K_{\theta})=\frac{e^{\z^i_{\theta+}}} {e^{\varepsilon^{-1}}}=\frac{\operatorname{diag}(\PP^{(1)})}{ \uu^{(1)}}.
\end{align}
The diagonal of K matrix contains the positive views and the marginal distribution of the u contains the negative view, we have: 
\begin{equation}
L^{\lambda,q}_\text{RINCE}(s_\theta^i) = -\frac{e^{ q s^i_{\theta+}}}{q} +  \frac{(\lambda\cdot (e^{s^i_{\theta+}} + \sum_{j=1,j\neq i}^{B} e^{s_{\theta-}^{ij}}))^q}{q} \propto  -\frac{\operatorname{diag} ({\K_\theta})_{ii}^{q}}{q} +  \frac{(\lambda\cdot (\sum_{j=1}^{B} {\K_\theta}_{ij} ))^q}{q}
\end{equation}
Furthermore, we have:
\begin{equation}\label{eq:w1}
-E(L_{\text{RINCE}}^{\lambda,q}({\bf K}_\theta)) = \frac{1}{q}\bigg( \operatorname{diag}(\K_{\theta}) \bigg)^q -  \frac{1}{q}\bigg(\frac{\lambda \I}{\uu^{(1)}}\bigg)^q.
\end{equation}
where $\PP^{(0)} = \diag(\mathbbm{1}) {\bf K}_{\theta} \diag(\mathbbm{1})$, $\PP^{(1)} = \diag({\uu^{(1)}}) {\bf K}_{\theta} \diag(\mathbbm{1})$, we have:
\begin{equation}
L_{\text{RINCE}}^{\lambda,q}(\PP_\theta^{(1)}) = -\frac{1}{q}(\frac{\operatorname{diag}(\PP^{(1)})}{\uu^{(1)}})^q+  \frac{1}{q}(\frac{\lambda \I}{\uu^{(1)}})^q.
\end{equation}

\subsubsection{Proof of the Symmetry and robustness of RINCE}
\label{app:symmetry}

Symmetry loss is said to be noise tolerant as the classifier will keep performance with the label noise in \textbf{Empirical Risk Minimization (ERM)}. In many practical machine learning scenarios, we aim to select a model or function $f_\theta$ that minimizes the expected loss across all possible inputs and outputs from a distribution $\mathcal{D}$, which is typically unknown. Instead of minimizing the true risk, which is often not feasible due to the unknown distribution $\mathcal{D}$, we minimize what is called the \textbf{empirical risk} $\hat{R}_L(\widetilde{f}_\theta)$, which is defined as the average loss over the training dataset of size $B$, which consists of independently and identically distributed (iid) data points. Mathematically, it is given by the following formula:
\begin{equation}\label{eq:emp_risk}
\hat{R}_L(f_\theta) = \frac{1}{B} \sum_{i=1}^{B} L(\widetilde{f_\theta}(\x_i), \y_i)
\end{equation}
Here, $L(\widetilde{f_\theta}(\x_i), \y_i)$ represents the loss function, which measures the discrepancy between the predicted value $\widetilde{f_\theta}(\x_i)$ and the true value $\y_i$. The function $\widetilde{f}_\theta$ that minimizes this empirical risk is chosen as the model for making predictions. This approach is based on the assumption that minimizing the empirical risk will also approximate the minimization of the true risk, especially as the size of the training set increases.

First we show the symmetry loss is robust to the noisy view with the following Lemma~\cite{ghosh2015making}, which means they will achieve the same performance in ERM with the noisy labels. Then we show RINCE satisfy the symmetry condition when \(q=1\), so the lemma is:

\begin{lemma}~\label{lm:symmetry} Give a loss function $L(\widetilde{f}_\theta(\x),\y)$ exhibits a certain symmetry for some positive constant \( K \), with respect to the labels $\y=1$ and $\y=-1$:
\begin{equation}~\label{eq:symmetry}
L(\widetilde{f}_\theta(\x), 1) + L(\widetilde{f}_\theta(\x), -1) = K, \quad \forall x, \forall f, \quad \text{(Symmetry)}
\end{equation}
Symmetry loss is noise tolerant given the label noise \( \eta < 0.5 \), which corresponds to the flipped labels:
\begin{equation}
P_D[\text{sign}(\widetilde{f}^*_\theta(x))=\y_\x] = P_D [\text{sign}(\widetilde{f}^*_{\theta\eta}(\x))=\y_\x], \quad \text{(Noisy tolerant)}
\end{equation}
\end{lemma}

\textbf{Proof of the Lemma~\ref{lm:symmetry} is in~\cite{ghosh2015making}.}

\textbf{Second we show the RINCE loss is a symmetry loss with $q\rightarrow 1$}, so we have the Equation~\eqref{eq:RINCE}:
\begin{equation}
L^{\lambda,q=1}_\text{RINCE} = -e^{\z^{ii}_{\theta +}} + \lambda \cdot (e^{\z^{ii}_{\theta +}} + \sum_{j=1,j\neq i}^{B} e^{\z_{\theta-}^{ij}})
\end{equation}

As we know that this formula has the same structure as the exponential loss function: \( L(\z_\theta, \y) = -\y e^{\z_\theta} \). To check for symmetry, we define a new binary classification loss function as:
   \[
   \tilde{L}_x(\z_\theta(\x), \y) = B + L_x(\widetilde{f}_\theta(\x), \y) = B - \y \cdot e^{\widetilde{f}_\theta(\x)} \geq 0
   \]
   where the prediction score \( \widetilde{f}_\theta(\x) \) is bounded by \( s_{\text{max}} = \log(B) \). Then we can establish that the loss satisfies the symmetry property:
   \begin{equation}
   \tilde{L}(\widetilde{f}_\theta(\x), 1) + \tilde{L}(\widetilde{f}_\theta(\x), -1) = 2B
   \end{equation}
So we prove that this loss function is symmetry.

\subsection{Proof for RINCE is the upper bound of the 1-Wasserstein distance}
\label{app:wdm}
In this section, we are trying to build the connection when change the \(d_M\) from the KL-divergence in Equation~\eqref{eq:kleqince} to the 1-Wasserstein distance in Equation~\eqref{eq:w1eqrince}, when q=1 in the RINCE loss.

\subsubsection{Proof of the Theorem~\ref{co:rince}}

WDM~\cite{chuang2022robust} is proposed as a replacement for the KL divergence by Wasserstein distance in Mutual Information estimation. The Wasserstein distance between the joint distribution \(\pi\) on \(\mathcal{X} \times \mathcal{Y}\) and the product of the marginal distributions \(\mu\) and \(\nu\) on \(\mathcal{X}\) and \(\mathcal{Y}\), respectively, is given by:
\[
W(\pi, \mu \otimes \nu) = \sup_{f \in \mathcal{C}(\mathcal{X} \times \mathcal{Y})} \left(\mathbb{E}_{\pi(x, y)}[f(x, y)] - \mathbb{E}_{\mu \otimes \nu(x, y)}[f(x, y)]\right)
\]
where \(\mathcal{C}(\mathcal{X} \times \mathcal{Y})\) denotes the set of all 1-Lipschitz functions from \(\mathcal{X} \times \mathcal{Y}\) to \(\mathbb{R}\). A function \( f : \mathcal{X} \times \mathcal{Y} \rightarrow \mathbb{R} \) is defined to be 1-Lipschitz if, for any two points \( (x_1, y_1), (x_2, y_2) \in \mathcal{X} \times \mathcal{Y} \), the following condition is satisfied: \[|f(x_1, y_1) - f(x_2, y_2)| \leq d((x_1, y_1), (x_2, y_2))\]
where \( d((x_1, y_1), (x_2, y_2)) \) denotes the metric on \( \mathcal{X} \times \mathcal{Y} \) typically defined, for example, by the Euclidean distance:
\[
d((x_1, y_1), (x_2, y_2)) = \sqrt{(x_1 - x_2)^2 + (y_1 - y_2)^2}
\]

Based on the Lipschitz continuity and inner product, it is easy to know for two given point \((x_1,y_1)\), \((x_2,y_2)\), the following properties hold with $-\frac{1}{\varepsilon} \leq s \leq \frac{1}{\varepsilon}$, which implies $\lvert \nabla_s e^s \rvert \leq e^{1/\varepsilon}$. Therefore, by the mean value theorem, we have:
\begin{equation}~\label{eq:lip}
\begin{split}
    &\lvert e^{x_1^T y_1/\varepsilon} - e^{x_2^T y_2/\varepsilon} \rvert \leq e^{1/\varepsilon} \frac{1}{\varepsilon} \lvert \langle x_1, y_1 \rangle - \langle x_2, y_2 \rangle \rvert = e^{1/\varepsilon} \frac{1}{\varepsilon} \lvert \langle x_1- x_2, y_1 \rangle + \langle x_2, y_1 - y_2 \rangle \rvert \\
    &\leq e^{1/\varepsilon} \frac{1}{\varepsilon} \left( \| x_1 - x_2\| \|y_1\|  + \| y_1-y_2\| \| x_2\|\right) = e^{1/\varepsilon} \frac{1}{\varepsilon} \left( \lVert x_1 - x_2 \rVert + \lVert y_1-y_2 \rVert \right)
\end{split}
\end{equation}
Consider two pairs of views, \((\z'_{\theta 1}, \z''_{\theta 1})\) and \((\z'_{\theta 2}, \z''_{\theta 2})\), sampled from the joint distribution \(\pi\) of \(\mu\) and \(\nu\). Thus, each pair \((\z'_{\theta i}, \z''_{\theta i})\) for \(i = 1, 2\) represents a sample from the joint distribution \(\pi\), where \(\z'_{\theta i} \sim \mu\) and \(\z''_{\theta i} \sim \nu\). The RINCE loss is a symmetry loss with $q=1$, so we have the Equation~\eqref{eq:RINCE}:
\begin{equation}
L^{\lambda,q=1}_\text{RINCE} = -e^{\z^{ii}_{\theta +}} + \lambda \cdot (e^{\z^{ii}_{\theta +}} + \sum_{j=1,j\neq i}^{B} e^{\z_{\theta-}^{ij}}),
\begin{cases} 
\z^{ii}_{\theta+} = \varepsilon^{-1}\widetilde{f_\theta}(\x'_i)^\top\widetilde{f_\theta}(\x''_i), & \text{for } i=i, \\
\z_{\theta-}^{ij} =\varepsilon^{-1} \widetilde{f_\theta}(\x'_i)^\top\widetilde{f_\theta}(\x''_j), & \text{for } i \neq j.
\end{cases}
\end{equation}

So we know that:
\begin{align*}
& -\mathbb{E}(L_{\text{RINCE}}^{\lambda ,q=1}(z_\theta)) = \mathbb{E}_{\substack{\z'_{\theta i}\sim\mu \\ \z''_{\theta i}\sim \nu | \mu=\z'_{\theta i} \\ \z''_{\theta j}\sim \nu}} \left[ (1-\lambda) e^{\varepsilon^{-1} \z'^T_{\theta i} \z''_{\theta i}} - \lambda \sum_{j=1}^{B-1} e^{\varepsilon^{-1} \z'^T_{\theta i} \z''_{\theta j}}\right] \\ 
&= \mathbb{E}_{(\z'_{\theta i},\z''_{\theta i} ) \sim \pi} \left[ (1 - \lambda) e^{\frac{ \z'^T_{\theta i} \z''_{\theta i}}{\varepsilon}} \right] - \lambda (B-1) \mathbb{E}_{\z'^T_{\theta i} \sim \mu, \z''^T_{\theta j} \sim \nu} \left[ e^{\frac{\z'^T_{\theta i} \z''_{\theta j}}{\varepsilon}} \right] \\ 
&\leq (1 - \lambda)  \left( \mathbb{E}_{(\z'_\theta,\z''_\theta) \sim \pi} \left[ e^{\frac{\z'^T_\theta \z''_\theta}{\varepsilon}} \right] - \mathbb{E}_{\z'^T_\theta \sim \mu, \z''_\theta \sim \nu} \left[ e^{\frac{\z'^T_{\theta} \z''_{\theta}}{\varepsilon}} \right] \right) (\text{Giving setting}~\lambda (B-1) > 1 - \lambda)\\
\end{align*}
If we give two couples of two views (\(\z'_{\theta 1},\z''_{\theta 1}\)) and (\(\z'_{\theta 2},\z''_{\theta 2}\)) from joint distribution \(\pi\) of \(\mu\) and \(\nu\), \(\z'_\theta\sim\mu\) and \(\z''_\theta\sim\nu\), which means to maximize:
\begin{align*}
& \lvert e^{\varepsilon^{-1} \z'^T_{\theta 1} \z''_{\theta 1}} - e^{\varepsilon^{-1} \z'^T_{\theta 2} \z''_{\theta 2}} \rvert\\
& \leq (1-\lambda) e^{\frac{1}{\varepsilon}} \frac{1}{\varepsilon} \left( \| \z'_{\theta 1} - \z'_{\theta 2}\| \|\z''_{\theta 1}\| + \| \z''_{\theta 1}-\z''_{\theta 2}\| \| \z'_{\theta 2}\| \right) \text{(Mean value theorem from Equation~\eqref{eq:lip})}\\
& = (1-\lambda) e^{\frac{1}{\varepsilon}} \frac{1}{\varepsilon} \left( \lVert \z'_{\theta 1} - \z'_{\theta 2} \rVert_2 + \lVert \z''_{\theta 1}-\z''_{\theta 2} \rVert_2 \right) \\
& = (1-\lambda) e^{\frac{1}{\varepsilon}} \frac{1}{\varepsilon} d\left((\z'_{\theta 1},\z''_{\theta 1}),(\z'_{\theta 2},\z''_{\theta 2})\right) \\
& \leq  (1 - \lambda) e^{1/\varepsilon} \frac{1}{\varepsilon} W_1(\pi, \mu\ \otimes \nu).
\end{align*}

\subsection{Proof of connection with BYOL}
\label{app:byol}

In this section, we are going to who how the change of the augmetation kernel from the \(\K_\theta\) in Definition~\eqref{def:gibbs} into the BYOL kernel \(\bS_{\theta}\) would lead to the BYOL loss.


\textbf{Proof for the Theorem~\ref{thm:byol}}

BYOL has the online network parameterized by \(\theta\) and target network parameterized by \(\xi\), where \(\z'_\theta=\widetilde{f}_{\theta}(\x')\) and \(\z''_\xi=\widetilde{f}_{\xi}(\x'')\) are the normalized outputs of the online and target networks, respectively. The kernel of BYOL looks like:
$$\bS_{\theta}(\x'_i, \x''_j) = \exp (-\langle \widetilde{q}_{\theta}(\widetilde{f}_{\theta}(\x'_i)), \widetilde{f}_{\xi}(\x''_j) \rangle),$$
The kernel here involves both the parameters \(\theta\) and \(\xi\), however, the target network has the stop gradient. Therefore, the only \(\theta\) needs to be updated, so we can rewrite the kernel as $\bS_{\theta}(\x'_i, \x''_j)$ as we show in the main text. As we give in the equation, the corresponding proximal operators evolving with \(d_\Gamma\) is equal to L2-distance has the formula, and \(h(x)=0\) for all \(\PP\in\mathcal{R}^{B\times B}\):
\[ \text{Prox}_{\mathcal{R}^{B\times B}}^{\|\cdot\|^2}(\mathbf{\bS_{\theta}}) = \arg\min_{\mathbf{P} \in \mathcal{R}^{B\times B}} \left\{ h(\mathbf{P}) + \frac{1}{2} \|\mathbf{P} - \mathbf{S}_{\theta}\|^2_2 \right\}\Rightarrow \PP=\bS_{\theta} \]

The BYOL loss can be written as normalized L2-distance between the normalized output after online network \(\widetilde{q}_{\theta}(\z'_{\theta})\) in which \(\widetilde{q}_\theta\) is predictor and the stop gradient results for the target network \(\widetilde{q}_{\theta} (\z')\), and the formula of BYOL object reads as $L_\text{BYOL}=  \| \widetilde{q}_{\theta}(\z'_{\theta}) - \z''_{\xi} \|_2^2$.

In this case, there exists equivalence between 
\begin{equation}
\text{KL}(\I\|\bS_{\theta})=-\sum_i^{B} \log \bS_{\theta ii}=\sum_i^{B} \| \widetilde{q}_{\theta}(\z'_{\theta}) - \z''_{\xi} \|_2^2
\end{equation}
which is the BYOL loss.

\subsection{Complexity Analysis for GCA}
\label{app:time_complexity}

In the forward pass, iteratively running the GCA does not involve inner optimization for gradient back-propagation.
In the Sinkhorn algorithm, the transport plan $\PP_\theta $ is computed as:
\[\PP_\theta = \exp(f + g - \C_\theta)/\epsilon,\]
where $f$ and $g$ are dual variables iteratively updated in the Sinkhorn algorithm but do not involve gradients with respect to $\theta$. The Sinkhorn optimization primarily entails scaling the rows and columns of $\PP$ to satisfy the marginal constraints, which can be viewed as element-wise operations (scaling and exponentiation) on the cost matrix $\C_\theta$.

Since $\PP_\theta$ is computed through the fixed-point iteration of $f$ and $g$ that depend only on the current values of $ \C_\theta $, the gradient back-propagation process is simplified. Specifically, the gradient of the loss with respect to the cost matrix $ \C_\theta $ is the key part that needs to be differentiated, rather than through each iterative update of $ f $ and $g$. A typical workflow of these algorithms was shown in Figure 2 of~\cite{eisenberger2022unified}, the gradient flow primarily involves differentiating through $ \C_\theta $, which is done only once, and not through each step of the Sinkhorn iterations. This approach reduces computational complexity and avoids the need for back-propagation through every iterative update within the Sinkhorn algorithm, which might otherwise be computationally expensive.

\section{Proofs that GCA methods improve the alignment and uniformity}

\subsection{Improving Alignment}
\label{app:alignment}
In this section, we are going to show the GCA methods minimize the difference between the target alignment plan with the coupling matrix on latent. 
The uniformity and alignment loss have been used to exam the quality of the representation in self-supervised learning, which is defined as the following~\cite{wang2020understanding}:

\begin{definition} [Alignment loss]
Given \(\pi\) as joint distribution of positive samples on the latent, \((\z'_{\theta i}, \z''_{\theta i})\) are the normalized positive pairs sampled from the joint distribution \(\pi\) with encoder parameterized by \(\theta\), the alignment loss is:

\begin{equation}\label{eq:alignment}
\mathcal{L}_{\text{align}}=\min_\theta~ \mathbb{E}_{(\z'_{\theta i}, \z''_{\theta i}) \sim \pi}\left[\|\z'_{\theta i}-\z''_{\theta i}\|_2^2\right] 
=\min_\theta~\sum_{i} \operatorname{diag}(\C_{ii}),
\end{equation}
where \(\C\) is the cost matrix defined in Equation~\eqref{eq:dot}.
\end{definition}

We can alter the constraint sets of proximal operators to provide the better alignment plans, i.e. GCA-INCE changes the constraint sets by considering both row and column normalization in coupling matrix Rather than just the row normalization. Such change will not affect the alignment loss in forward pass, it will benefit the alignment loss in the backward pass through a tighter bound of empirical risk minimization with the identity matrix.

\subsubsection{Proof of the tighter bound of GCA in ERM}
\label{app:klfg}
In this section, we provide the evidence for using the converged coupling plan \(\PP^{(\infty)}_\theta\) is better than the \(\PP^{(1)}_\theta\) or \(\PP^{(t)}_\theta\) in Equation~\eqref{eq:breg1} for the GCA-methods loss in table~\ref{tab:diff}. This loss function will correspond to different alignment loss on the latent. And here the ERM is the definition as we provided in Appendix~\ref{app:symmetry}.

\begin{lemma}\label{lm:klfg}
Denote \(f^{(t_1)}\) and \(g^{(t_2)}\) the two dual variables in their \(t_1\) and \(t_2\) iterations, respectively.
Then the objective loss in Equation~\eqref{thm:cl_eq_ot} could be written as \(
\operatorname{KL({\bf I}\|\PP_\theta}) =\varepsilon^{-1}(\operatorname{diag}(\C)-(f^{(t_1)}+g^{(t_2)}))\).
\end{lemma}

\textbf{Proof of the Lemma~\ref{lm:klfg}}:

The above Lemma~\ref{lm:klfg} be derived form Equation~\eqref{eq:uvupdatesapp}. 
Recall that \(\uu=\exp{(f/\varepsilon)}, \quad \vv=\exp{(g/\varepsilon)}, \quad \K=\exp{(-\C/\varepsilon)}\)
\begin{align*}
\operatorname{KL({\bf I}\|\PP_\theta}) & = -\sum_i \log\left(\PP_{ii}\right) = -\sum_i(\log \operatorname{diag}(\uu)_{ii} +\log \K_{ii}+\log\operatorname{diag}(\vv)_{ii}) \\
                                 &= -\varepsilon^{-1}\sum_i(f_i-C_{ii}+g_i)=\varepsilon^{-1}(\operatorname{diag}(C)-(f+g))
                                 \qed
\end{align*}

Here we provide the proof of the \textit{Best Alignment} in \textbf{Theorem}~\ref{thm:lowest}:

\textbf{Proof of the Theorem~\ref{thm:lowest}}

Based on the Lemma~\ref{lm:klfg}, to show \(\operatorname{KL({\bf I}\|\PP^{(\infty)}}) \leq \operatorname{KL({\bf I}\|\PP^{(1)}}) \).
We have to show: 
$$\varepsilon^{-1}(\operatorname{diag}(C)-(f^{(\infty)}+g^{(\infty)}))\leq \varepsilon^{-1}(\operatorname{diag}(C)-(f^{(1)}+g^{(1)})).$$
Then give the Lemma~\ref{lm:monotonicity}, we know the \(f^{(t_1)}\) and \(g^{(t_2)}\) increase and converge weakly to their upper bound. As the \(\operatorname{diag}(\C)\) will be unchanged in each proximal operations, we know the objective function \(\text{KL}({\bf I}\|\PP_\theta)\) have lower upper bound with \(f^{(t_1)}\) and \(g^{(t_2)}\) increase and finally converged. \qed

Based on the Lemma~\ref{lm:klfg}, We have to show: 
$$\varepsilon^{-1}(\operatorname{diag}(C)-(f^{(t)}+g^{(t)}))\geq \varepsilon^{-1}(\operatorname{diag}(C)-(f^{(\infty)}+g^{(\infty)})).$$
Then give the Lemma~\ref{lm:monotonicity}, when \(t_1\rightarrow \infty \), \(f^{(t_1)}\) converge uniformly to a fixed point \( f^{(\infty)} \) with \(f^{(t_1)} \leq f^{(\infty)}\). So we know the \(f^{(t_1)}\) and \(g^{(t_2)}\) increase and converge weakly to their upper bound. As the \(\operatorname{diag}(\C)\) will be unchanged in each proximal operations, we know the objective function finally converged to \(f^{(\infty)}\) and \(g^{(\infty)}\) (Similarly, we prove \(g^{(t_1)} \leq g^{(\infty)}\) in Appendix~\ref{app:monotonicity}).\qed

\subsubsection{Proof of the Theorem~\ref{thm:gcarince}:}
To show:
\( L_{\text{GCA-RINCE}}^{\lambda,q=1, \varepsilon}(\PP_\theta^{(t)}) \leq  L_{\text{RINCE}}^{\lambda,q=1, \varepsilon}(\PP_\theta^{(1)})\).

we know that:
\begin{equation}
L_{\text{GCA-RINCE}}^{\lambda,q=1, \varepsilon}(\PP_\theta^{(t)}) = -\frac{\operatorname{diag}(\PP^{(t)})}{\uu^{(t)}} + \frac{\lambda \I}{\uu^{(t)}},
\end{equation}

\begin{equation}
L_{\text{RINCE}}^{\lambda,q=1, \varepsilon}(\PP_\theta^{(1)}) = -\frac{\operatorname{diag}(\PP^{(1)})}{\uu^{(1)}} + \frac{\lambda \I}{\uu^{(1)}},
\end{equation}

Given that the Lemma~\ref{lm:monotonicity}, we know \(\{\uu^{(t)}\}\) and \(\{\vv^{(t)}\}\) are a monotonically increasing sequence where 
\begin{align}
 \uu^{(1)}\leq \uu^{(t)}&\Rightarrow \frac{\lambda \I}{\uu^{(1)}}\geq \frac{\lambda \I}{\uu^{(t)}} \\
-\operatorname{diag}(\K_\theta)\vv^{(0)}\geq -\operatorname{diag}(\K_\theta)\vv^{(t)} & \Rightarrow - \frac{\operatorname{diag}(\PP^{(1)})}{\uu^{(1)}}\geq  -\frac{\operatorname{diag}(\PP^{(t)})}{\uu^{(t)}}
\end{align}
Combine the above two items, we have the equation like
\( L_{\text{GCA-RINCE}}^{\lambda,q=1, \varepsilon}(\PP_\theta^{(t)}) \leq  L_{\text{RINCE}}^{\lambda,q=1, \varepsilon}(\PP_\theta^{(1)})\).

\subsection{GCA methods improve the uniformity and benefit downstream classification tasks}
\label{app:uniformity}
In this section, we provide theoretical evidence that the GCA approaches could improve the performance of downstream task, i.e. classification tasks, by providing the maximum uniformity through solving the EOT, as \textbf{Theorem}~\eqref{thm:uniformity} stated. Here, the uniformity loss is defined as~\cite{saunshi2022understanding}:

\begin{definition} [Uniformity loss]
Let \(\z'_{\theta i}\sim\mu\) and \(\z''_{\theta j}\sim\nu\) in which \(\mu\) and \(\nu\) are two distributions on the representation space, we define the uniformity loss as the following:
\begin{equation}\label{eq:uniformity}
    L_{\text{uniform}} = \log \mathbb{E}_{\z'_{\theta i},\z''_{\theta j} \, \text{i.i.d.} \sim p_{\text{data}}} [ e^{-\varepsilon \left\|  \z'_{\theta i}-  \z''_{\theta j} \right\|^2_2} ]
\end{equation}, in which \( p_\text{data}(\cdot) \) is the sample distribution over latent space \( \mathbb{R}^n \).
\end{definition}
Here, \( p_\text{data}(\cdot) \) should be the marginal distribution of the samples. As the \(\z'_{\theta i}\) and \(\z''_\theta j\) are normalized latent variables, we have the right items of the uniformity loss \(e^{-\varepsilon \|\z'_{\theta i}-  \z''_{\theta j} \|^2_2}\) is the same as the entropy-regularized kernel \(\K_{ij}=e^{-\varepsilon\C_{ij}}\) with cost matrix items \(\C_{ij}=\|  \z'_{\theta i}-  \z''_{\theta j} \|^2_2\). 

\subsubsection{Proof of the Theorem~\ref{thm:uniformity}}

Here we are going to compare two different coupling plans, \(\PP_\theta^{(1)}\) and \(\PP_\theta^{(\infty)}\), and show the converged plan \(\PP_\theta^{(\infty)}\) will achieve higher the uniformity after the forward pass.  
The general logic is that we show the equivalence for the solving EOT with the minimizing the uniformity loss objective. Then we use the convergence of iterative Bregman projections to show it could achieve higher uniformity.

Based on the Entropy regularized OT defined in Definition~\ref{def:phidivergence}, we have:
\begin{equation}
W_{c, \varepsilon}(\mu, \nu) \coloneqq \min_{\pi \in \Pi(\mu, \nu)} \int_{X \times Y} c(x, y) d\pi(x, y) + \varepsilon H(\pi | \mu \otimes \nu)
\end{equation}
in which the entropy could be defined as:
\begin{equation}
H(\pi | \mu \otimes \nu) \coloneqq \int_{X \times Y} \left( \log \left( \frac{d\pi(x, y)}{d\mu(x)d\nu(y)} \right) - 1 \right) d\pi(x, y) + 1,
\end{equation}
is the relative entropy of the transport plan \( \pi \) with respect to the product measure \( \mu \otimes \nu \). So the corresponding dual problem of this EOT one is shown in the following formula:
\begin{align}
W_{c, \varepsilon}(\mu, \nu) &= \max_{f \in C(\mathcal{X}), g \in C(\mathcal{Y})}  \int_\mathcal{X} f(x) \, d\mu(x) + \int_\mathcal{Y} g(y) \, d\nu(y)\\ &~~~~~~~~~~~ - \varepsilon \int_{\mathcal{X}  \times \mathcal{Y}} e^{\frac{f(x) + g(y) - c(x,y)}{\varepsilon}} \, d\mu(x)d\nu(y) + \varepsilon \\
&= \max_{f \in C(\mathcal{X}), g \in C(\mathcal{Y})} \mathbb{E}_{\mu \otimes \nu}\left[ f(x) + g(y) - e^{\frac{f(x) + g(y) - c(x,y)}{\varepsilon}} \right] +\varepsilon
\end{align}

The \(\mu(x)\) and \(\nu(x)\) are defined as the uniformly distribution with Dirac delta function we have on the two latent supports \(\{\z'_{\theta i}\}_{i=1}^{B}\) and \(\{\z''_{\theta i}\}_{i=1}^{B}\), so the function \(f(x)\) and \(g(y)\) could be pull out of the expectation operators. 
Since the \( \|  \z'_{\theta i}-  \z''_{\theta j} \|^2_2 \) is the element in the cost matrix \(\C_{ij}\), which is computed through the cost function \(c(x,y)\). As the \(\z'_{\theta i}\) and \(\z''_{\theta j}\) are drawn independently from the latent distribution, so the remaining item \(\mathbb{E}_{\mu \otimes \nu}[e^{\frac{- c(x,y)}{\epsilon}}]\) is equivalent to the uniformity loss. The the above integral could be turned into the sum of the elements in matrix of dual variables of \(f^{(t_1)}\) and\(g^{(t_1)}\) in each iteration.
Meanwhile, based on the convergence provided in the Lemma~\ref{lm:monotonicity}, When \(t_1\rightarrow \infty \), \(f^{(t_1)}\) converge uniformly to a fixed point \( f^{(\infty)} \) with \(f^{(t_1)} \leq f^{(\infty)}\), which would provided the maximum value of the dual formula in the \(f^{(\infty)}\), which corresponding to the coupling plan the \(\PP^{(\infty)}\). \(\qed\)

\subsubsection{GCA benefits the downstream supervised classification task} \label{app:gca_sup}

Here, we further show how the minimizing the uniformity loss is equivalent to minimize the downstream supervised loss in classification tasks under several assumptions~\cite{dufumier2023integrating}.
Giving a labeled dataset $\mathcal{D} = \{(\bar{\x}_i, \y_i)\} \in \bar{\mathcal{X}} \times \mathcal{Y}$ where $\mathcal{Y} = [1..M]$ with $M$ classes, we consider a fixed, pre-trained encoder $f_\theta \in \mathcal{F}: \mathcal{X} \to \mathcal{S}$ with its representation $f_\theta({\mathcal{X}})$ and the input space $\mathcal{X}$ contains both positive and negative views of $n$ original samples $(\bar{\x}_i)_{i \in [1..n]} \in \bar{\mathcal{X}}$, sampled from the data distribution $p(\bar{\x})$. 
For each positive views $\bar{\x}'_i$ in $\mathcal{X}$, we sample from $\bar{\x}_i$ using $\x_i' \sim \mathcal{A}(\cdot|\bar{\x}_i)$, \(\mathcal{A}(\cdot|\bar{\x}_i)\) is augmentation distribution (e.g., by applying color jittering, flip, or crop with a given probability). For consistency, we assume $\mathcal{A}(\bar{\x}) = p(\bar{\x})$ so that the distributions $\mathcal{A}(\cdot|\bar{\x})$ and $p(\bar{\x})$ induce a marginal distribution $p(\x)$ over $\mathcal{X}$. Given an anchor $\bar{\x}_i$, all views $\x'' \sim \mathcal{A}(\cdot|\bar{\x}_j), j \neq i$ from different samples $\bar{\x}_j$ are considered as negatives. 

\textbf{Proof of claim~\ref{cl:ce_eq_uniform}:}
From assumption~\ref{asm:express_encoder} we know that the representation ability of encoders is good enough via the augmented samples in the Reproducing Kernel Hilbert Space (RKHS) \(\mathcal{H}_{\bar{\mathcal{X}}}\) of the original sample spaces $\bar{\mathcal{X}}$. And the kernel \( K_{\bar{\mathcal{X}}} \) with any function \( g \) RKHS defined by \( (\mathcal{H}_{f_\theta}, \K_\theta) \) also belongs to \( H_{\bar{\mathcal{X}}} \) when conditioned on the distribution \(\mathcal{A}(\x|\cdot)\). So based on the assumption we have, we can obtain a centroid estimator by~\cite{dufumier2023integrating}:

\begin{definition}[Kernel-based centroid estimator]
Let $(\x_i, \bar{\x}_i)_{i \in [1..n]} \sim \mathcal{A}(\x, \bar{\x})$, asssuming a consistent estimator of $\mu_{\bar{\x}}$ is. 
\[
\forall \bar{\x} \in \bar{\mathcal{X}}, \hat{\mu}_{\bar{\x}} = \sum_{i=1}^{n} \alpha_i(\bar{\x}) f(\x_i),
\]
where $\alpha_i(\bar{\x}) = \sum_{j=1}^{n} [(\K_n + n\lambda \I_n)^{-1}]_{ij} \K_{\bar{\mathcal{X}}}(\bar{\x}_j, \bar{\x})$ and $\K_n = [\K_{\bar{\mathcal{X}}}(\bar{\x}_i, \bar{\x}_j)]_{i,j \in [1..n]}$. It converges to $\mu_{\bar{\x}}$ with the $\ell_2$ norm at a rate $\mathcal{O}(n^{-1/4})$ for $\lambda = \mathcal{O}(n^{-1/2})$. 
\end{definition}

The above estimator allows us to use representations of images close to an anchor $\bar{\x}$ to estimate $\mu_{\bar{\x}}$. From the assumption~\ref{asm:small_intra_variance}, we assume that all the samples in the same class is achievable when give the ideal augmentation or at least close to the augmented points in an \(\epsilon\) region.

 Consequently, if the prior is “good enough” to connect intra-class images disconnected in the augmentation graph suggested by Assumption~\ref{asm:express_encoder}, then this estimator allows us to tightly control the classification risk of the representation of $f_\theta$ on a classification task with a linear classifier $g(\bar{\x}) = \W f_\theta(\bar{\x})$ (with $f_\theta$ fixed) that minimizes the multi-class classification loss.

 \textbf{First we show the cross-entropy could be transformed into centroid based distance (optimal supervised loss):}
\label{transform_ce_to_sup}
The cross-entropy (CE) to measure the difference between the true distribution (actual labels) and the estimated probability distribution (predicted probabilities from the model), which usually computes logits \( \z_k \) from the model, then apply the softmax function to obtain probabilities \( p_k \). 
The logits \( \z_k \) could be defined as negative distances between \( f(\bar{\x}) \) and class centroids \( \mu_k \) after the representation:
\[
\z_k = -\| f(\bar{\x}) - \mu_k \|^2, \quad \mu_k = \mathbb{E}_{p(\bar{\x}|\y=k)} \mu_{\bar{\x}}
\]
which encourages the model to reduce the distance to the correct class centroid while increasing distances to others. The probability of class \( k \) in M classes given input \( \bar{\x} \) is:
\[
p(\y = k | \bar{\x}) = \frac{e^{\z_k}}{\sum_{j=1}^M e^{\z_j}}, \quad  p(\y | \bar{\x}) \propto e^{-\| f(\bar{\x}) - \mu_\y \|^2}. 
\]
If the model predictions \( p(\y | \bar{\x}) \) are influenced by the distances between \( \bar{\x} \) and the class centroids \( \mu_\y \), then minimizing cross-entropy indirectly affects these distances.
The standard CE loss in supervised learning for classification tasks is:
\begin{align}
\mathcal{L}_{\text{CE}}(f_\theta) & = -\mathbb{E}_{(\bar{\x}, \y) \sim \mathcal{D}} \left[ \log p(\y | \bar{\x}) \right] \\ & =-\mathbb{E}_{(\bar{\x}, \y) \sim \mathcal{D}} \left[ -\| f(\bar{\x}) - \mu_\y \|^2 - \log Z \right] =-\frac{1}{N} \sum_{i=1}^N \sum_{k=1}^M \y_{i,k} \log(p_{i,k})
\end{align}

which focuses on maximizing the likelihood \( \hat{\y} = \arg\max_k p(\y = k | \bar{\x}) \) of the correct class for each individual sample \( \bar{\x}_i \), where \( y_{i,k} \) is the true label indicator for example \( i \) and class \( k \), \( p_{i,k} \) is the predicted probability for example \( i \) and class \( k \). Therefore, we can rewrite the  CE loss as optimal supervised loss in~\cite{dufumier2023integrating}, which is defined as:

\begin{lemma} [Optimal supervised loss]\label{lm:ot_sup_loss}
Let a downstream task $D$ with $M$ classes. We assume that $M \leq d + 1$ (i.e., a big enough representation space), that all classes are balanced and the realizability of an encoder $f^* = \arg\min_{f \in F} \mathcal{L}_{\text{sup}}(f_\theta)$ with
\[
\mathcal{L}_{\text{sup}}(f_\theta) = \log \mathbb{E}_{y, y' \sim p(y)p(y')} \left[ e^{-\| \mu_y - \mu_{y'} \|^2} \right],
\]
and $\mu_y = \mathbb{E}_{p(\bar{x}|y)} \mu_{\bar{x}}$. Then the optimal centroids $(\mu^*_y)_{y \in Y}$ associated to $f^*$ make a regular simplex on the hypersphere $S^{d-1}$ and they are perfectly linearly separable, i.e.,
\[
\min_{(w_y)_{y \in \mathcal{Y}} \in \mathbb{R}^d} \mathbb{E}_{(\bar{x}, y) \sim D} \mathbbm{1}(w_y \cdot \mu^*_y < 0) = 0.
\]
\end{lemma}
\textbf{Proof of the Lemma~\ref{lm:ot_sup_loss}}
All "labeled" centroids $\mu_y = \mathbb{E}_{p(\bar{x}|y)} \mu_{\bar{x}}$ are bounded by 1 ($\|\mu_y\| \leq \mathbb{E}_{p(\bar{x}|y)} \mathbb{E}_{A(x|x')} \| f(x) \| = 1$ by Jensen's inequality). Then, since all classes are balanced, we can re-write the supervised loss as:
\[
\mathcal{L}_{\text{sup}}(f_\theta) = \log \frac{1}{C^2} \sum_{y, y'=1}^{C} e^{-\| \mu_y - \mu_{y'} \|^2}.
\]
We have:
\[
\Gamma_\mathcal{Y}(\mu) := \sum_{y, y'} \|\mu_y - \mu_{y'}\|^2 = \sum_{y, y'} \|\mu_y\|^2 + \|\mu_{y'}\|^2 - 2 \mu_y \cdot \mu_{y'}
\leq \sum_{y, y'} (2 - 2 \mu_y \cdot \mu_{y'})
= 2C^2 - 2 \| \sum_{y} \mu_y \|^2 \leq 2C^2,
\]
with equality if and only if $\sum_{y=1}^{C} \mu_y = 0$ and $\forall y \in [1..C], \|\mu_y\| = 1$. By the strict convexity of $u \to e^{-u}$, we have:
\[
\sum_{y \neq y'} \exp(-\|\mu_y - \mu_{y'}\|^2) \geq C(C-1) \exp \left( - \frac{\Gamma_Y(\mu)}{C(C-1)} \right)
\geq C(C-1) \exp \left( - \frac{2C}{C-1} \right),
\]
with equality if and only if all pairwise distances $\|\mu_y - \mu_{y'}\|$ are equal (equality case in Jensen's inequality for a strict convex function), $\sum_{y=1}^{C} \mu_y = 0$, and $\|\mu_y\| = 1$. Thus, all centroids must form a regular $(C-1)$-simplex inscribed on the hypersphere $S^{d-1}$ centered at 0. Furthermore, since $\|\mu_y\| = 1$, we have equality in Jensen's inequality:
\[
\|\mu_y\| = \|\mathbb{E}_{\mathcal{A}(\x|\bar{\x}')} f_\theta(x)\| \leq \mathbb{E}_{ \mathcal{A}(\x|\bar{\x}')} \|f_\theta(x)\| = 1,
\] 
so $f$ must be perfectly aligned for all samples belonging to the same class: $\forall x, \bar{x}' \sim p(\cdot|y), f_\theta(\bar{x}) = f_\theta(\bar{x}')$. $\qed$

\textbf{Seond we show optimizing the uniformity loss is equivalent to the supervised loss:}

As we have uniformity Loss defined in Equation~\eqref{eq:uniformity}
\begin{equation}
L_{\text{uniform}}(f_\theta) = \log \mathbb{E}_{\z'_i, \z''_j \sim p_{\text{data}}} \left[ e^{ -\varepsilon \left\| \z'_i - \z''_j \right\|^2 } \right],
\end{equation}

where \( \z'_i = f(\x_i) \) and \( \z''_j = f(\x_j) \). Supervised Loss:
\[
\mathcal{L}_{\text{sup}}(f_\theta) = \log \mathbb{E}_{y, y' \sim p(y)p(y')} \left[ e^{ -\left\| \mu_y - \mu_{y'} \right\|^2 } \right],
\]
where \( \mu_y = \mathbb{E}_{p(\bar{\x}| y)} \hat{\mu}_{\bar{\x}} \). Express the expectation over all pairs in terms of class labels:
\[
\mathbb{E}_{\z'_i, \z''_j} = \mathbb{E}_{y, y'} \mathbb{E}_{\z'_i \sim p(\z | y), \z''_j \sim p(\z | y')}.
\]
So the uniformity loss could be decomposed into intra-class and inter-class components:
\[
L_{\text{uniform}}(f_\theta) = \log \left( \underbrace{\mathbb{E}_{y} \left[ \mathbb{E}_{\z'_i, \z''_j \sim p(\z | y)} \left[ e^{ -\varepsilon \left\| \z'_i - \z''_j \right\|^2 } \right] \right]}_{\text{Intra-Class Term}} + \underbrace{\mathbb{E}_{y \neq y'} \left[ \mathbb{E}_{\z'_i \sim p(\z | y), \z''_j \sim p(\z | y')} \left[ e^{ -\varepsilon \left\| \z'_i - \z''_j \right\|^2 } \right] \right]}_{\text{Inter-Class Term}} \right).
\]

Based on the assumption~\ref{asm:small_intra_variance}, we can approximate the Intra-Class term by:
\begin{align*}
\left\| \z'_i - \z''_j \right\|^2 & = \left\| (\mu_y + \delta_i) - (\mu_{y'} + \delta_j) \right\|^2 = \left\| \mu_y - \mu_{y'} + \delta_i - \delta_j \right\|^2\approx \left\| \mu_y - \mu_{y'} \right\|^2 \\
&  \implies \mathbb{E}_{\z'_i \sim p(\z | y), \z''_j \sim p(\z | y')} \left[ e^{ -\varepsilon \left\| \z'_i - \z''_j \right\|^2 } \right] \approx e^{ -\varepsilon \left\| \mu_y - \mu_{y'} \right\|^2 }
\end{align*}

for \( y = y' \), \( \z'_i \) and \( \z''_j \) are close to \( \mu_y \)
\[
\left\| \z'_i - \z''_j \right\|^2 \approx \left\| (\mu_y + \delta_i) - (\mu_y + \delta_j) \right\|^2 = \left\| \delta_i - \delta_j \right\|^2.
\]
Since \( \delta_i \) and \( \delta_j \) are small deviations:
\[
\mathbb{E}_{\z'_i, \z''_j \sim p(\z | y)} \left[ e^{ -\varepsilon \left\| \z'_i - \z''_j \right\|^2 } \right] \approx 1, \quad e^{ -\varepsilon \left\| \delta_i - \delta_j \right\|^2 } \approx 1.
\]

Then with \(M\) terms for \(y=y'\) and \(M(M-1)\) terms of \(y\neq y'\), we have:

\begin{equation}
L_{\text{uniform}} =\log (\frac{1}{M}e^{ -\varepsilon \left\| \delta_i - \delta_j \right\|^2 } +  \frac{1}{M^2} \sum_{y \neq y'} e^{ -\varepsilon \left\| \mu_y - \mu_{y'} \right\|^2 } ) 
\end{equation}

The supervised loss is:
\[
\mathcal{L}_{\text{sup}}(f_\theta) = \log ( \frac{1}{M^2} \sum_{y, y'} e^{ -\left\| \mu_y - \mu_{y'} \right\|^2 } ) = \log ( \frac{1}{M} e^{ -\left\| \mu_y - \mu_{y} \right\|^2 } + \frac{1}{M^2} \sum_{y \neq y'} e^{ -\left\| \mu_y - \mu_{y'} \right\|^2 } ) 
\]
Since \( e^{ -\left\| \mu_y - \mu_{y} \right\|^2 } = 1 \) (for \( y = y' \)), the difference will be mainly dependent on the inter-class term. Therefore, a tighter (smaller) uniformity loss leads to smaller values of the supervised loss. This supports the idea that improving uniformity in representations can benefit downstream supervised classification tasks. \(\qed\)

\subsection{Unbalanced OT assists to alleviate the feature suppression}
\label{app:gcauot-feature}

Although minimizing the uniformity loss can enhance downstream classification tasks, it may also lead the model to learn shortcut features that could impair the encoder's generalization ability. To show this, we incorporate two propositions from previous work by Robinson et al.~\cite{robinson2021can}.

\subsubsection{The uniformity loss causes feature suppression}

For an encoder $f_\theta : \mathcal{X} \rightarrow \mathbb{S}^{d-1}$ to map input data $\x$ to the surface of the unit sphere $\mathbb{S}^{d-1} = \{ u \in \mathbb{R}^d : ||u||_2 = 1 \}$. 
Suppose we have the latent feature spaces $\Z^1, \dots, \Z^n$ with a distribution $p_j$ on each latent space $\Z^j$ with $j \in [n]$ to model a distinct feature. We write $\Z$ instead of $\Z^{[n]}$ for the product as $\Z^S = \prod_{j \in S} \Z^j$, where $[n] = \{1, \dots, n\}$.
So the latent sample \( \z \) could be represented as a set of feature vectors \( \z = (\z^1, \z^2, \dots, \z^n)= (\z^j)_{j \in S} \in \Z \), where each \( \z^j \) comes from \( \Z^j \). Further, let $\lambda$ denote the measure on $\Z$ induced by $\z$ and 
$\lambda( \cdot | \z^S)$ denote the conditional measure on $\Z$ for fixed $\z^S$. For $S \subseteq [n]$ we use $\z^S$ to denote the projection of $\z$ onto $Z^S$. Finally, an injective map $g : \Z \rightarrow \mathcal{X}$ produces observations $\x = g(\z)$. The feature suppression is defined as:

\begin{definition}
Consider an encoder $f_\theta : \mathcal{X} \rightarrow \,\mathbb{S}^{d-1}$ and features $S \subseteq [n]$. For each $\z^S \in Z^S$, let $\mu(\cdot | \z^S)$ be the pushforward measure on $S^{d-1}$ by $f_\theta \circ g$ of the conditional $\lambda( \cdot | \z^S)$.

\begin{enumerate}
    \item $f_\theta$ suppresses $S$ if for any pair $\z^S, \tilde{\z}^S \in Z^S$, we have $\mu(\cdot | \z^S) = \mu(\cdot | \tilde{\z}^S)$.
    \item $f_\theta$ distinguishes $S$ if for any pair of distinct $\z^S, \tilde{\z}^S \in \Z^S$, measures $\mu(\cdot | \z^S), \mu(\cdot | \tilde{\z}^S)$ have disjoint support.
\end{enumerate}
\end{definition}

If one feature is uniformly distributed on the latent space, it might cause feature suppression
due to different features could both achieve the minimization of the uniformity loss as the following propositions~\cite{robinson2021can}:

\begin{proposition} [Feature suppression]\label{prop:feat_suppress}
For a set $S \subseteq [n]$ of features let
\[
L_S(f_\theta) = L_{\text{align}}(f_\theta) + \mathbb{E}_{\x^+} \left[ - \log \mathbb{E}_\x \left[ e^{f(\x^+ )^\top f(\x^- )} \middle| \z^S = \z^{S^-} \right] \right]
\]
denote the (limiting) InfoNCE conditioned on $\x^+, \x^-$ having the same features $S$. Suppose that $p_j$ is uniform on $Z^j = S^{d-1}$ for all $j \in [n]$. Then the infimum $\inf L_S$ is attained, and every $f_\theta \in \arg\min_f L_S(f_\theta')$ suppresses features $S$ almost surely.
\end{proposition} 

\textbf{Proof of proposition~\ref{prop:feat_suppress} is in \cite{robinson2021can}}.

\subsubsection{How the GCA methods and unbalanced OT and  alleviates the feature suppression}  

Here we extended the unbalanced OT in the Equation~\eqref{eq:uot} as the following: 
\begin{equation}
    \min_{\theta}~ d_M ( \PP_{\text tgt} \| \PP_\theta)  + \lambda_1 d_{\phi_1}(\PP_\theta) +\lambda_2 d_{\phi_2}(\PP_\theta) +\cdots+\lambda_n d_{\phi_n}(\PP_\theta)
\end{equation}

The UOT equation can be converted with finding the transport plan \( \PP_\theta \) that minimizes the transportation cost between two probability measures \( \mu \) and \( \nu \). Here we only need to show that the relaxation or adding penalties will change the optimal transport plan \(\PP_\theta\), which is empirically exhibited in the Figure~\ref{fig:diagnolline}.

Suppose we have empirical samples \( \{ \z'_i \}_{i=1}^n \) from \( \mu \) and \( \{ \z''_j \}_{j=1}^m \) from \( \nu \). We can approximate the measures using empirical distributions:
\[
\mu \approx \frac{1}{n} \sum_{i=1}^n \delta_{\z'_i}, \quad \nu \approx \frac{1}{m} \sum_{j=1}^m \delta_{\z''_j},
\]
where \( \delta_{\z} \) is the Dirac delta function at point \( \z \). The standard UOT objective can be written as:

\begin{align}
&  \min_{\PP \geq 0} \sum_{i=1}^n \sum_{j=1}^m \C(\z'_i, \z''_j) \PP_{ij} + \lambda_1 d_{\phi_1}\left( \sum_{j=1}^m \PP_{ij} \Big\| \frac{1}{n} \right) + \lambda_2 d_{\phi_2}\left( \sum_{i=1}^n \PP_{ij} \Big\| \frac{1}{m} \right)
\\ & = \min_{\PP \geq 0} \sum_{i=1}^n \sum_{j=1}^m \left[ \C_{ij} \PP_{ij} + \lambda_1 \PP_{ij} \left( \log \frac{\PP_{ij}}{r_i} - 1 \right) + \lambda_2 \PP_{ij} \left( \log \frac{\PP_{ij}}{c_j} - 1 \right) \right]
\end{align}

where \( \C \) is the cost matrix \( d_{\phi} \) could be any divergence (e.g., Kullback-Leibler divergence) with respect to a convex function \( \phi \).
\( \PP \mathbf{1}_{\mu} \) and \( \PP^\top \mathbf{1}_{\nu} \) are the marginal distributions.
\( \lambda_1, \lambda_2 \) are regularization parameters controlling the unbalancedness and \( r_i = \frac{1}{n} \) (source marginal mass for \( \z'_i \)), \( c_j = \frac{1}{m} \) (target marginal mass for \( \z''_j \)). 
Based on the UOT, here we can choose the divergence as \(\mathcal{L}\):
\[
   \mathcal{L}(\mathbf{P}) = \sum_{i,j} \left[ C_{ij} \mathbf{P}_{ij} + \lambda_1 \mathbf{P}_{ij} \left( \log \frac{\mathbf{P}_{ij}}{r_i} - 1 \right) + \lambda_2 \mathbf{P}_{ij} \left( \log \frac{\mathbf{P}_{ij}}{c_j} - 1 \right) \right]
   \]

To find the minimizer, we take the partial derivative of \(L(\mathbf{P})\) with respect to \(\mathbf{P}_{ij}\) and set it to zero:

\begin{align}
& \frac{\partial \mathcal{L}}{\partial \mathbf{P}_{ij}} = C_{ij} + \lambda_1 \left( \log \frac{\mathbf{P}_{ij}}{r_i} \right) + \lambda_2 \left( \log \frac{\mathbf{P}_{ij}}{c_j} \right) = 0
\\ & \implies \lambda_1 \left( \log \mathbf{P}_{ij} - \log r_i \right) + \lambda_2 \left( \log \mathbf{P}_{ij} - \log c_j \right) = -C_{ij}
\\ & \implies
   (\lambda_1 + \lambda_2) \log \mathbf{P}_{ij} - \lambda_1 \log r_i - \lambda_2 \log c_j = -C_{ij}
\\ & \implies    \log \mathbf{P}_{ij} = \frac{ -C_{ij} + \lambda_1 \log r_i + \lambda_2 \log c_j }{ \lambda_1 + \lambda_2 }
\\ & \implies
   \mathbf{P}_{ij} = \exp\left( \frac{ -C_{ij} + \lambda_1 \log r_i + \lambda_2 \log c_j }{ \lambda_1 + \lambda_2 } \right)
\end{align}

The minimizer \(\mathbf{P}_{ij}\) depends on \(\lambda_1\) and \(\lambda_2\) and the weights of \(r_i\) and \(c_j\), which determine the influence of the marginals \(r_i\) and \(c_j\), and through the scaling of the cost \(C_{ij}\) by \(\lambda_1 + \lambda_2\). This explicit relationship shows how \(\lambda_1\) and \(\lambda_2\) determine the minimizer.

\section{Details of Experiments}
\label{app:exp_details}

The following experiments involving with the GPU was set up on NVIDIA GeForce RTX 3090.

\subsection{Experimental details on image classification task}
\label{app:hpdetails}
In Table~\ref{tab:new_combine} standard settings, we used two different experimental setups. The first setup, referred to as the C0 or standard settings, was applied specifically to the CIFAR10 and CIFAR100 tasks. The second setup was used for the SVHN and ImageNet100 tasks, respectively. Below, we present the settings for CIFAR10 and CIFAR100, followed by the setups for SVHN and ImageNet100. 
Here is the setups for CIFAR10 and CIFAR100:
\begin{itemize}
    \item The SSL model has 512 feature dimensions with the base model (ResNet-18), which first convolutional changed as a layer with 3 input channels, 64 output channels, kernel size 3, stride 1, padding 1, and no bias. We replace the max-pooling layer as the identity.
    \item A sequential projector comprising a linear layer mapping from feature dimension to 2048, ReLU activation, and another linear layer mapping from 2048 to 128. 

\item For SSL training, an SGD optimizer is used with a learning rate of 0.6, momentum 0.9, and a weight decay of 1.0e-6. A LambdaLR scheduler is employed with linearly decay the learning rate to 1.0e-3 over total steps, which equals the length of the SSL training loader times the maximum epochs. The SSL model is trained for a maximum of 500 epochs, without loading a pre-trained model. The parameters of encoders are frozen after training. Temperature or epsilon: 0.5.

\item For supervised training, an Adam optimizer is also used with a learning rate of 0.2, momentum 0.9 and a weight decay of 0. A same LambdaLR scheduler is applied, where the learning rate is reduced by a factor of 1.0e-3.  For supervised training, the model is trained for a maximum of 200 epochs using the specified train and test loaders.

\end{itemize}

The setups for SVHN and ImageNet100 are:

\begin{itemize}
    \item The SSL model has number of feature dimensions equal to the fc layer incoming features of base model (ResNet-50). We replace the max-pooling layer as the identity.
    \item A sequential projector comprising a linear layer mapping from feature dimension to 2048, ReLU activation, and another linear layer mapping from 2048 to 128. 

\item For SSL training, an Adam optimizer is used with a learning rate of 3e-4. The SSL model is trained for a maximum of 200 epochs for ImageNet100 and 500 epochs for the SVHN, without loading a pre-trained model. The parameters of encoders are frozen after training. Temperature or epsilon: 0.5.

\item For supervised training, an Adam optimizer is also used with a learning rate of 3e-4. The model is trained for a maximum of 100 epochs using the specified train and test loaders. 

\end{itemize}
\subsection{Settings for extreme data augmentations}\label{app:corruptset}

There is the "extreme DA" (Ex DA) column in Table~\ref{tab:new_combine}, which is the average of the following three settings:
\begin{itemize}
    \item  C1: Large Erase Settings: Here, we first employed the same standard augmentation as C0 in Appendix~\ref{app:hpdetails} does, than we apply the random erase with 'p=1' (random erasing is applied every time), the 'scale=(0.10, 0.33)'. The large erase is applied before the normalization.
    
    \item C2: Strong Crop Setting: This involves a strong cropping operation followed by resizing, which applied by 'transforms.RandomCrop' and 'transforms.Resize'. The crop size varies based on the severity level, with values ranging from 96 to 224 pixels. We selected level 3 during our experiments, than Resizes the cropped image back to 32x32 pixels.
    
    \item C3: Brightness settings: This augmentation alters the brightness of the images. We have 'severity' determines the degree of brightness change, with predefined levels ranging from `.05` to `.3`, corresponding to level 1 and level 5. And we chosse the level 5 as our C3 augmentation. The brightness is adjusted in the HSV color space, specifically altering the value channel to change the brightness.

\end{itemize}

To evaluate performance on CIFAR10-C, we use a pretrained SSL model with frozen parameters. Fine-tuning is performed by training only the linear layer with \(10\%\) of CIFAR10-C data for 50 epochs. We compute the final score by averaging results across all corruption types and severity levels in CIFAR10-C.
And the details of each column are provided in Table~\ref{tab:cifar10}, Table~\ref{tab:cifar1004c} and Table~\ref{tab:cifar10c4c}.

\begin{table}[h]
\centering \caption{\footnotesize {\em Test accuracy for contrastive methods on CIFAR-10.} Test accuracy for different contrastive methods and their GCA equivalents on CIFAR-10 for ResNet-18 under extreme augmentation conditions, averaged over 5 seeds.}
\resizebox{\textwidth}{!}{%
\begin{tabular}{l|c|c|c|c|c|c|c|c|c}\label{tab:cifar10}
\textbf{Conditions} & \textbf{INCE} & \textbf{GCA-INCE} & \textbf{RINCE} & \textbf{GCA-RINCE}& \textbf{SimCLR} & \textbf{BYOL} & \textbf{IOT} & \textbf{IOT-uni} & \textbf{GCA-UOT}  \\ \hline
Standard            & 92.01 ± 0.40   & 93.02 ± 0.19  & 93.27±0.20  & 93.47±0.32  & 92.16 ±0.16  & 90.56 ± 0.59  & 92.10 ± 0.22  & 91.49  ± 0.11  & \textbf{93.50±0.31 } \\ \hline
Erase               & 88.40 ± 0.17     & 88.16 ± 0.89 & 88.80±1.01   & 89.21±0.59   &  88.44 ± 0.24  & 88.77 ± 0.58 &87.02 ± 0.43  &87.83 ± 0.30     & \textbf{89.84 ± 0.58}     \\ \hline
Crop                & 72.45 ± 0.40  & 72.79 ± 0.62   & 73.02±0.39 & 73.10±0.31    & 71.84 ± 1.02   & 70.78 ± 0.62 &70.44 ± 0.64 &70.78 ± 0.21 & \textbf{73.35 ± 0.41}  \\ \hline
Brightness          &    85.24 ± 0.41    & 85.60 ± 0.57 & 85.97±0.50  & 85.98 ± 0.58    & 85.32 ± 0.32  & 85.10 ± 0.29   &84.31 ± 0.84 &83.77 ± 0.21     & \textbf{86.36±0.34}
\end{tabular}}
\end{table}

\begin{table}[h]
\centering
\caption{\footnotesize {\em Test accuracy for contrastive methods on CIFAR-100.} Test accuracy for different contrastive methods and their GCA equivalents on CIFAR-100 using ResNet-18 under extreme augmentation conditions, averaged over 5 seeds.}
\resizebox{\textwidth}{!}{%
\begin{tabular}{l|c|c|c|c|c|c|c|c|c}\label{tab:cifar100}
\textbf{Conditions} & \textbf{INCE} & \textbf{GCA-INCE} & \textbf{RINCE} & \textbf{GCA-RINCE} & \textbf{SimCLR} & \textbf{BYOL} & \textbf{IOT} & \textbf{IOT-uni} & \textbf{GCA-UOT}  \\ \hline
Standard            & 71.09 ± 0.31  & 71.55 ± 0.12  & 71.63 ± 0.36  & 71.95 ± 0.48  & 70.85 ± 0.50  & 69.75 ± 0.37  & 68.37 ± 0.42  & 68.62 ± 0.35  & \textbf{72.16 ± 0.38} \\ \hline
Large Erase         & 62.54 ± 0.20  & 62.65 ± 0.17  & 63.55 ± 0.14  & 63.14 ± 0.41  & 62.94 ± 0.13  & 62.70 ± 0.31  & 62.69 ± 0.34  & 62.56 ± 0.22  & \textbf{63.62 ± 0.27} \\ \hline
Strong Crop         & 45.67 ± 0.31  & 46.31 ± 0.22  & 46.47 ± 0.20  & 46.50 ± 0.35  & 46.05 ± 0.34  & 43.11 ± 0.41  & 45.36 ± 0.19  & 45.29 ± 0.12  & \textbf{46.60 ± 0.34} \\ \hline
Brightness          & 59.87 ± 0.36  & 59.68 ± 0.33  & 60.38 ± 0.25  & 60.56 ± 0.25  & 60.46 ± 0.19  & 55.74 ± 0.63  & 57.52 ± 0.32  & 57.42 ± 0.18  & \textbf{61.02 ± 0.55} \\ 
\end{tabular}
}
\label{tab:cifar1004c}
\end{table}

\begin{table}[h]
\centering
\caption{\footnotesize {\em Test accuracy for contrastive methods on CIFAR-10C.} Test accuracy for different contrastive methods and their GCA equivalents on CIFAR-10C for ResNet-18 under extreme augmentation conditions, averaged over 5 seeds.}
\resizebox{\textwidth}{!}{%
\begin{tabular}{l|c|c|c|c|c|c|c|c|c}\label{tab:cifar10c}
\textbf{Conditions} & \textbf{INCE} & \textbf{GCA-INCE} & \textbf{RINCE} & \textbf{GCA-RINCE} & \textbf{SimCLR} & \textbf{BYOL} & \textbf{IOT} & \textbf{IOT-uni} & \textbf{GCA-UOT}  \\ \hline
Standard            & 81.52 ± 1.04  & 82.63 ± 0.28  & 82.86 ± 0.21  & 82.87 ± 0.11  & 81.74 ± 1.54  & 82.43 ± 0.06  & 82.01 ± 0.80  & 81.18 ± 1.12  & \textbf{82.90 ± 0.49} \\ \hline
Large Erase         & 82.67 ± 0.31  & 82.19 ± 0.35  & 83.11 ± 0.36  & \textbf{83.44 ± 0.55}  & 82.49 ± 0.24  & 82.53 ± 0.34  & 82.30 ± 0.69  & 82.30 ± 0.48  & 83.09 ± 0.62 \\ \hline
Strong Crop         & 43.55 ± 0.90  & 41.12 ± 3.28  & 42.51 ± 1.48  & 44.96 ± 1.46  & 42.74 ± 1.27  & 43.86 ± 0.91  & 42.36 ± 0.87  & 42.43 ± 1.03  & \textbf{45.20 ± 1.20} \\ \hline
Brightness          & 84.00 ± 0.47  & 84.08 ± 0.26  & 84.24 ± 0.36  & 84.54 ± 0.26  & 83.86 ± 0.46  & 83.87 ± 0.19  & 82.33 ± 0.31  & 80.86 ± 0.72  & \textbf{84.88 ± 0.44} \\ 
\end{tabular}}
\label{tab:cifar10c4c}
\end{table}

\subsection{Experimental setting for domain generalization}
\label{appendix:dgsetting}

This section is going to show the settings of experiments in Figure~\ref{fig:Ptgt}, which involves the domain generalization task. Training was executed under the DomainBed framework. Each model underwent training across multiple domains, with 5 distinct seeds (seed 71, 68, 42, 36, 15) used to ensure reproducibility:


\begin{itemize}
  \item For SSL model configuration, we employed a ResNet-18 architecture as the encoder, following with a 2048-dimensional, 3-layer projector equipped with BatchNorm1D and ReLU activations. We improved the framework of the SelfReg algorithm in Domainbed~\cite{gulrajani2020search} by a self-supervised contrastive learning phase which involves the GCA-INCE, with regularized parameters \(\varepsilon=0.2\).
  \item For SSL training hyperparameters, an Adam optimizer is used with a learning rate of 3e-4, and a weight decay of 1.5e-6. A Cosine Annealing learning rate scheduler is employed with a maximum number of 200 iterations equal to the length of the SSL training. The learning rate is scheduled to decrease to a minimum value of 0. The SSL model is trained for a maximum of 1500 epochs.

    \item In the self-supervised learning phase, we utilized 20\% of the data from each of the four datasets in the PACS dataset. The unsupervised holdout part employed contrastive learning augmentations to enhance generalization capabilities. Specifically, we implemented dual augmentation, including operations such as random resized crops, flips, color jitter, and grayscale conversion, standardized to an input shape of \(3 \times 224 \times 224\).
    \item The supervised learning rate was set at \(5 \times 10^{-5}\) using MSE loss, and the Adam optimizer with no weight decay. Training involved both domain and class labels over 3000 epochs, with checkpoints every 300 epochs to capture the model's best performance. This approach was supplemented by fine-tuning the model post-unsupervised training phase. Domain labels were categorized into four types corresponding to the PACS dataset, and class labels were divided into five categories. In domain classification, all four domains are used for training, with 70\% of the data held out for training and the remaining 30\% used for testing. Four domains are utilized for class classification tasks. We train supervised models on three domains and test on the fourth.
  \item The domain accuracy is computed as the average of the highest domain accuracies across five seeds, with each of the four test domains set sequentially as the test domain. The standard deviation for domain accuracy is calculated from the results across these five seeds.
  \item Class label accuracy is determined by averaging the accuracies of the four test environments for each domain. The average of highest performance across the domain is taken as the mean accuracy. The standard deviation for each domain is computed from the five seeds, and these values are then averaged to obtain the final class standard deviation.

\end{itemize}
Both the label classification tasks and the domain classification tasks use the Mean Squared Error (MSE) loss.

\section{Additional Experiments}\label{app:add_experiments}

\subsection{Complexity Analysis of GCA Algorithms}
\label{app:time}

\begin{figure}[ht]
    \centering
    \includegraphics[width=1\linewidth]{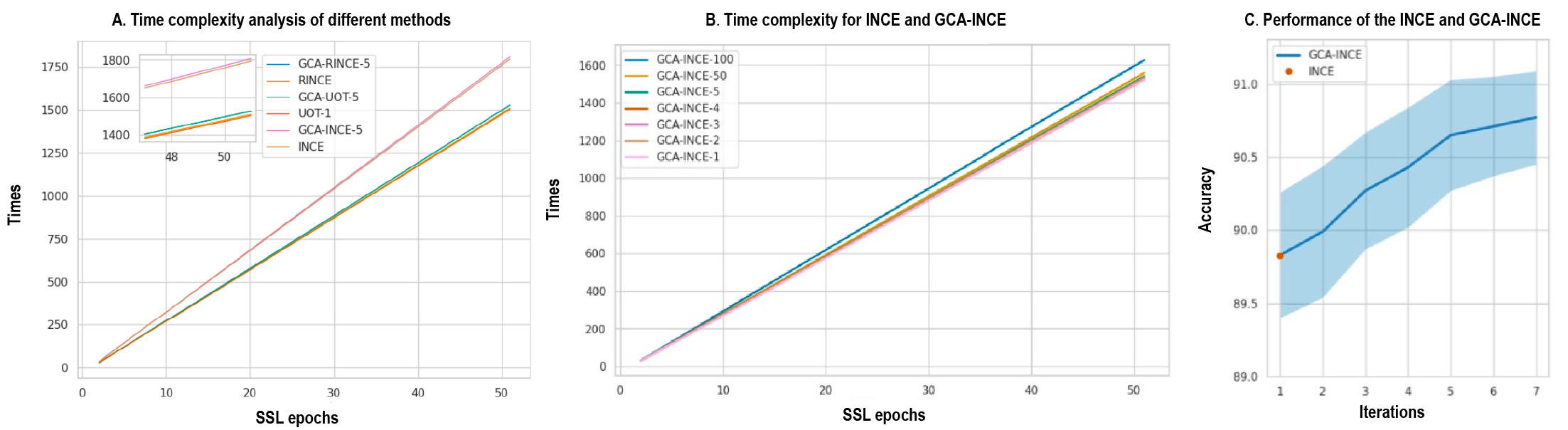}
    \caption{\footnotesize{\em Time complexity analysis}  (A) Time complexity analysis of different methods.  Here, we provide the time complexity for different contrastive methods (INCE, RINCE) and GCA-based methods (GCA-INCE, GCA-RINCE, and GCA-UOT) on CIFAR-10. (B) Time complexity for INCE (GCA-INCE-1), and GCA-INCE with different number of iterations GCA-INCE-100 denotes GCA-INCE with 100 iterations. We ran the methods on the CIFAR-10 as self-supervised learning task for 50 epochs, and compared their run time. (C) Performance of the INCE (iteration=1) and GCA-INCE (iterations>1) on the CIFAR10 with different number of iterations. The shaded blue region is the standard deviation across 5 seeds.
}
    \label{fig:time_comp}
\end{figure}

\textbf{Time complexity analysis:}
The computational complexity of GCA including the forward pass and backward propagation phases. The complexity varies in different variants. For GCA-INCE, the computational complexity of forward pass is related to the speed of Sinkhorn when solving the EOT problem as $O(n^2/\varepsilon^3)$, in which $\varepsilon$ is the regularization parameter . For GCA-UOT, the forward complexity is the Sinkhorn algorithm solving unbalanced OT, which is characterized by $$O(\tau (\alpha + \beta)^2/\varepsilon \log(n) [\log(\|C\|_\infty) + \log(\log(n)) + \log(1/\varepsilon)]),$$ where $C$ is the cost matrix, $\alpha$ and $\beta$ denote the total masses of the measures, and $\tau$ is a regularization parameter related to KL divergences in the UOT framework~\cite{pham2020unbalanced}. 
Notably, the gradient backpropagation speed is not seriously affected by scaling operations in the EOT as we explained in Section~\ref{app:time_complexity}. Moreover, the relaxations of penalties in UOT provide a even faster speed compared with the INCE and GCA-INCE (see Figure ~\ref{fig:time_comp}).

The complexity of the forward pass is affected by the choice of proximal operator, whereas the complexity of the gradient backward pass is influenced by the form of $d_M$~\cite{nguyen2010estimating}. Notably, utilizing Sinkhorn algorithms in GCA-UOT, GCA-RINCE, and GCA-INCE, only requires updating the coupling matrix \(\PP\) \((B \times B)\) without impacting the complexity of the backward pass, where \(B\) is the batch size. OT is known to have $B^2$ complexity and in many cases can converge very quickly in fewer than 10 iterations. In practice, we use a simple stopping criterion for the multiple iterations using a convergence criterion.

Upon analyzing the run time for the different methods (see Figure~\ref{fig:time_comp}) we observe that the GCA-based variants of the different base approaches (INCE, or RINCE) achieve very similar run time as their equivalent loss, but different losses (RINCE vs INCE) exhibit more significant variability. Specifically, we find that RINCE and GCA-RINCE have lower time complexity than INCE and GCA-INCE. So the runnning speed is even quicker if we utillized different \(d_M\) in Equation~\eqref{eq:mainobj}.

\subsection{Measuring the representation quality using alignment and uniformity}
\label{app:representation}

We study the uniformity and alignment of the representations learned by our GCA-INCE vs. INCE variants of GCA in Algorithms~\ref{alg:gca}.
We train the model through the corresponding settings (C0: standard provided in the , C1: erase, C2: crop, C3: brightness) provided in the Appendix~\ref{app:hpdetails} and Appendix~\ref{app:corruptset}. We find that in general, the GCA variants improve the representation quality evaluated by alignment and uniformity on both CIFAR-10 and CIFAR-10C datasets.

\begin{figure}[ht]
  \begin{minipage}[c]{0.49\textwidth}
    \centerline{\includegraphics[width=0.8\textwidth]{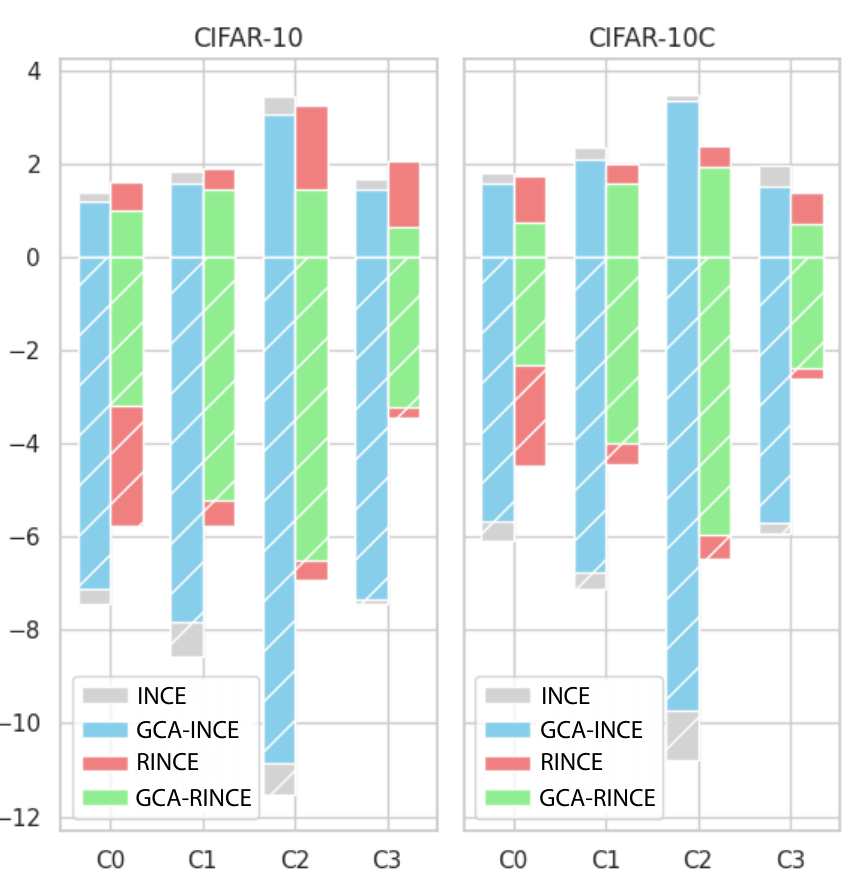}}
  \end{minipage}\hfill
  \begin{minipage}[c]{0.5\textwidth}
  \caption{ \footnotesize{\em Alignment and uniformity metrics on CIFAR-10.} To visualize the ability of uniformity and alignment with different methods under different augmentation settings (C0: standard, C1: erase, C2: crop, C3: brightness). The bar above the x axis (zero line) represents the alignment loss, while the bar under the x axis represents the uniformity loss. The shorter the color bars i.e with lower alignment loss and higher uniformity loss, correspond to the better performance of SSL models.} 
  \end{minipage}
\end{figure}

\subsection{Visualizing transport plans of different methods after training}
\label{app:vistransplan}

\begin{figure}[ht]
  \begin{minipage}[c]{0.6\textwidth}
    \centerline{\includegraphics[width=0.9\textwidth]{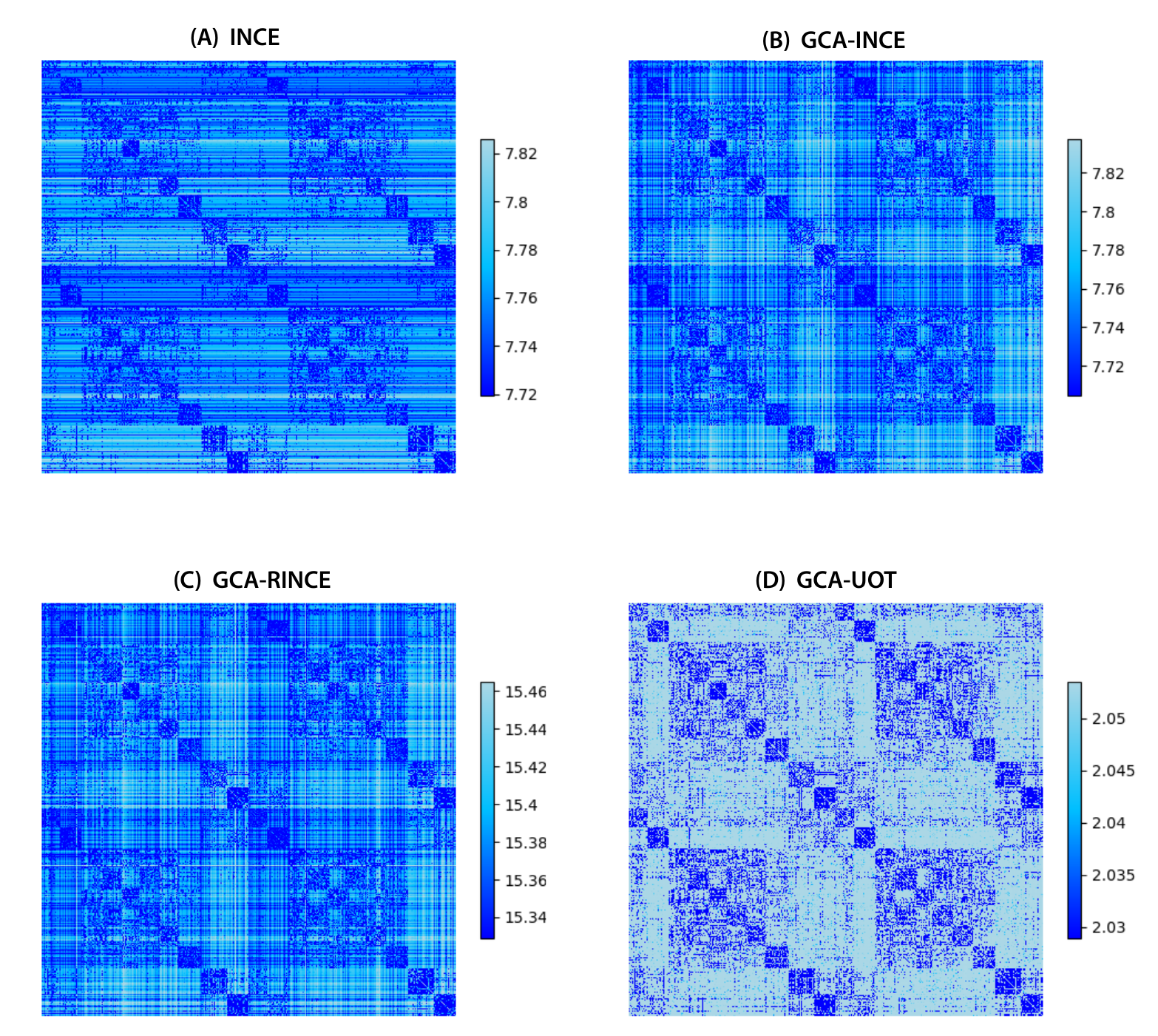}}
  \end{minipage}\hfill
  \begin{minipage}[c]{0.3\textwidth}
  \caption{\footnotesize { \em Comparison of the \( -\log(\mathbf{P}) \) matrix across different methods.} (A) The INCE matrix with row normalization. (B) The \( -\log(\mathbf{P}) \) matrix of GCA-INCE with five iterations in forward pass, both row and column normalization. (C) The \( -\log(\mathbf{P}) \) matrix of GCA-RINCE with five iterations in forward pass. (D) The \( -\log(\mathbf{P}) \) matrix of GCA-UOT with five iterations in forward pass}
 \label{fig:diagnolline}
  \end{minipage}
\end{figure}

Here we compared the optimal transport (OT) plans of different methods after training for 500 epochs under standard augmentation C0 settings in Appendix~\ref{app:hpdetails}. Specifically, we analyzed the \( -\log(\mathbf{P}) \) matrices of INCE, GCA-INCE, GCA-RINCE, and GCA-UOT, as shown in Figure~\ref{fig:diagnolline}. In these matrices, darker blue regions represent higher similarity, while lighter blue areas indicate less similarity. The matrices are rearranged based on class labels, so an effective model should display empty diagonals and block structures aligned along the main diagonal and sub-diagonals—reflecting high intra-class similarity and low inter-class similarity.

Figure~\ref{fig:diagnolline}(A) shows that INCE results in a matrix with only row normalization. In contrast, Figures~\ref{fig:diagnolline}(B) and (C) demonstrate that GCA-INCE and GCA-RINCE achieve both row and column normalization, leading to more uniform distributions. Figure~\ref{fig:diagnolline}(D) reveals that GCA-UOT produces a matrix highlighting greater differences between positive and negative pairs, underscoring its effectiveness in distinguishing them.

\subsection{Hyperparameter Tuning and Sensitivity Analysis}

In our hyperparameter modifying experiments, we investigate the influence of key parameters in transport plan regularization, iteration counts, and augmentation strengths on CIFAR-10 classification performance.

\begin{figure}[ht]
  \centering
  \includegraphics[width=0.7\textwidth]{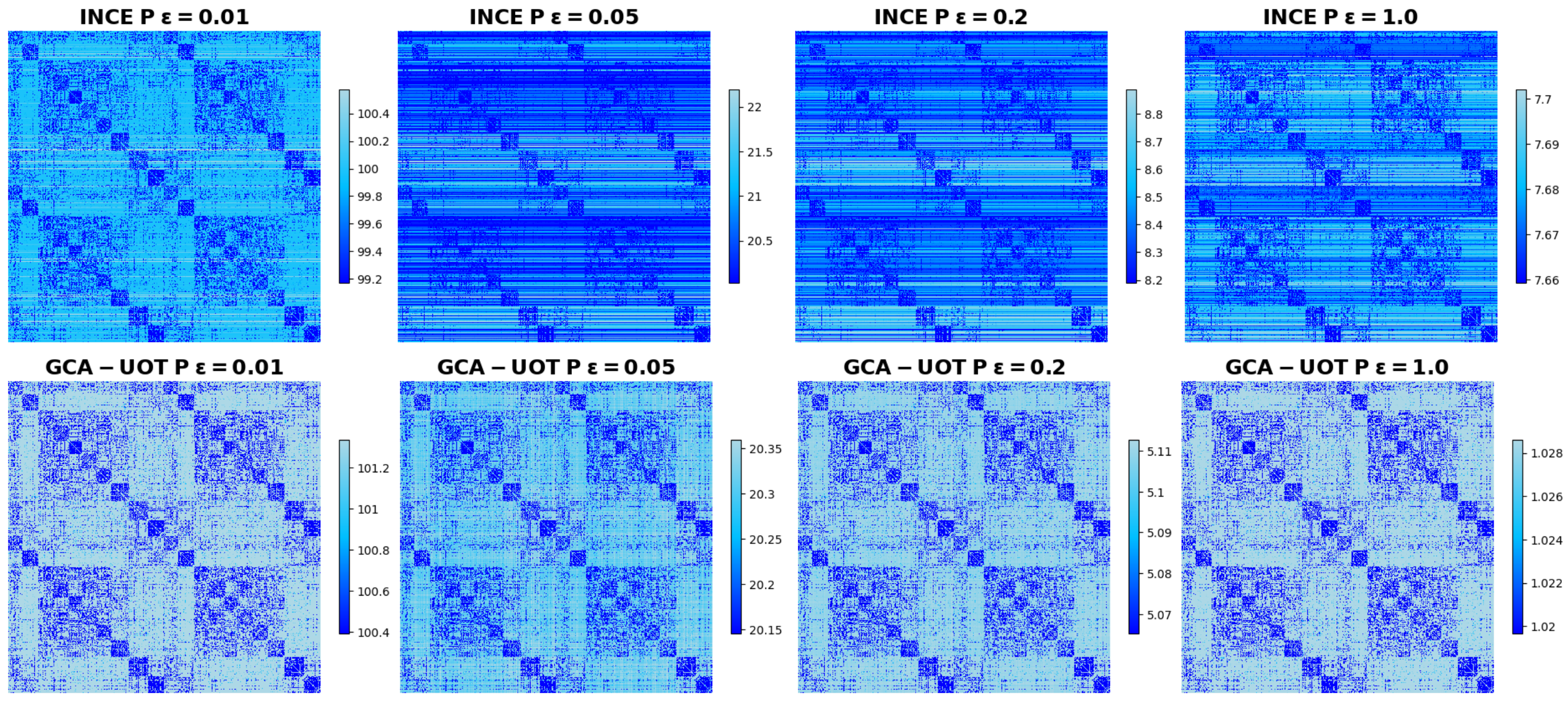}
  \caption{\footnotesize { \em Visualization transport plan P for different amounts of entropic regularization.} (Top) The transport plans for \(\epsilon\) from 0.01 to 1 for INCE and (Bottom) GCA-UOT after 5 iterations. To compute each plan, we took a mini-batch on CIFAR-10 with 1024 samples, and loaded the same weights of Resnet-18 for each subfigure.}
  \label{fig:epsilon_study}
\end{figure}

\begin{figure}[ht]
  \centering
  \begin{minipage}[c]{0.6\textwidth}
    \includegraphics[width=\textwidth]{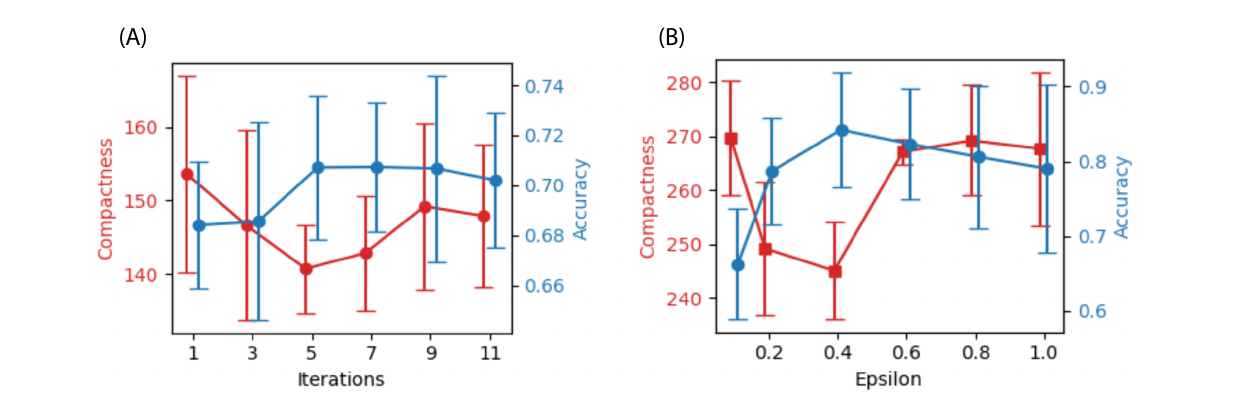}
  \end{minipage}%
  \hfill
  \begin{minipage}[c]{0.4\textwidth}
    \caption{\footnotesize {\em Hyperparameter sensitivity study.} The compactness and accuracy as a function of the (A) number of iterations and the (B) entropic regularization parameter. In our experiments, we use the same weights and perform 20 pre-training epochs for each point, then evaluate their performance by fine-tuning linear classifiers for 20 epochs. Here the compactness is the average L2 distance of each point to their corresponding class center on the representation space after the encoder.}
    \label{fig:sensitivity_study}
  \end{minipage}
\end{figure}

\begin{figure}[ht]
  \centering
  \begin{minipage}[c]{0.55\textwidth}
    \includegraphics[width=\textwidth]{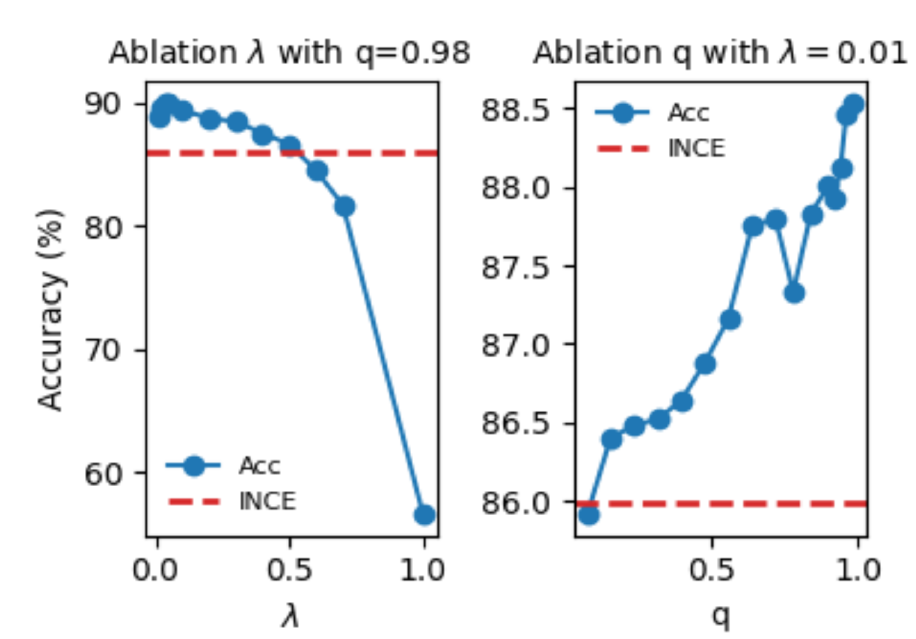}
  \end{minipage}%
  \hfill
  \begin{minipage}[c]{0.4\textwidth}
    \caption{\footnotesize {\em Hyperparameter sensitivity for \(q\) and \(\lambda\) in GCA-RINCE.} Both experiments are tested on the CIFAR-10 dataset with a ResNet-18 encoder and involve strong augmentation with large erase. (Left) Given \(q = 0.98\), we change \(\lambda\) from 0 to 1. (Right) Given \(\lambda = 0.01\), we change \(q\) from 0 to 1. The red threshold line is the INCE performance with the large erase augmentation. Each point represents the CIFAR-10 classification accuracy of the ResNet-18 model pre-trained for 400 epochs and evaluated after 300 epochs.}
    \label{fig:ablation_study}
  \end{minipage}
\end{figure}

Figure~\ref{fig:epsilon_study} visualizes transport plans under varying entropic regularization (\(\epsilon\) values from 0.01 to 1) across INCE and GCA-UOT models, illustrating adjustments after five iterations using the same ResNet-18 weights. Figure~\ref{fig:sensitivity_study} examines the impact of iteration number and entropic regularization on compactness—measured by the average L2 distance to class centers—and accuracy, with 20 pre-training epochs followed by fine-tuning. Figure~\ref{fig:ablation_study} highlights the sensitivity of GCA-RINCE to the hyperparameters \(q\) and \(\lambda\), testing classification accuracy for different settings under strong augmentation conditions; this includes a comparison against INCE with large erase augmentation after substantial pre-training and evaluation epochs.

\end{document}